\documentclass[]{worldbench}
\definecolor{w_blue}{RGB}{52,204,204}
\definecolor{w_yellow}{RGB}{255,192,0}

\definecolor{forestgreen}{rgb}{0.133, 0.545, 0.133}
\definecolor{yellowyellow}{rgb}{0.133, 0.545, 0.133}
\definecolor{lightblue}{RGB}{68, 114, 196}
\definecolor{correct}{RGB}{173, 173, 173}
\definecolor{incorrect}{RGB}{234, 59, 46}

\usepackage{amsfonts}
\usepackage{amsmath}
\usepackage{amssymb}

\usepackage{colortbl}

\usepackage{makecell}
\usepackage{multicol}
\usepackage{multirow}
\usepackage{pifont}

\usepackage{wrapfig}
\usepackage{titletoc}

\usepackage{xspace}
\usepackage{amsmath}

\newcommand*{\ie}{\emph{i.e.}\@\xspace}

\newcommand*{\Xcrop}{X_\mathrm{in}}
\newcommand*{\Ycrop}{Y_\mathrm{in}}
\newcommand*{\Xcomp}{X_\mathrm{out}}

\newcommand*{\ev}{\mathbb{E}}
\newcommand*{\evemp}{\hat{\mathbb{E}}}
\newcommand*{\Pemp}{P}
\newcommand*{\Hemp}{\hat{H}}

\usepackage{pifont}
\newcommand{\cmark}{\ding{51}}
\newcommand{\xmark}{\ding{55}}

\title{Multi-Modal Data-Efficient 3D Scene Understanding for Autonomous Driving}

\author[]{Lingdong~Kong~\raisebox{0.2em}{\includegraphics[width=0.019\linewidth]{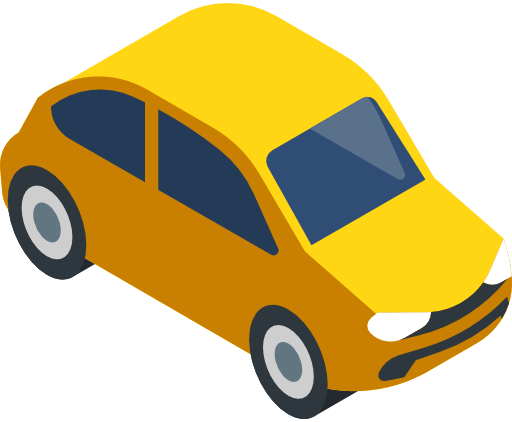}}}
\author[]{Xiang~Xu}
\author[]{Jiawei~Ren}
\author[]{Wenwei~Zhang}
\author[]{Liang~Pan}
\author[]{Kai~Chen}
\author[]{Wei~Tsang~Ooi}
\author[]{Ziwei~Liu~\raisebox{0.2em}{\includegraphics[width=0.019\linewidth]{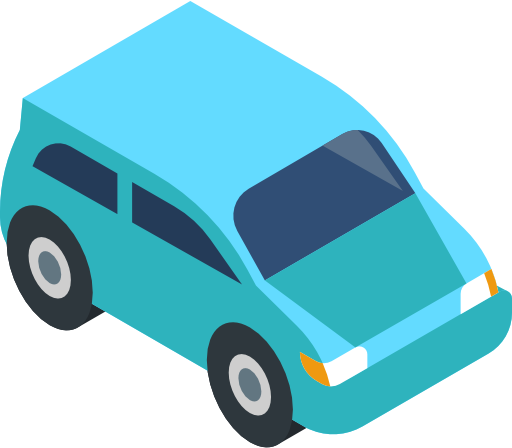}}}

\affiliation[]{
\raisebox{-0.1em}{\includegraphics[width=0.029\linewidth]{figures/icons/worldbench.png}}~WorldBench Team
\\[1.2ex]
~\raisebox{-0.2em}{\includegraphics[width=0.032\linewidth]{figures/icons/car1.png}}~{\small \textbf{Project Lead}}
\quad
\raisebox{-0.2em}{\includegraphics[width=0.031\linewidth]{figures/icons/car2.png}}~{\small \textbf{Corresponding Author}}
}

\abstract{
Efficient data utilization is crucial for advancing 3D scene understanding in autonomous driving, where reliance on heavily human-annotated LiDAR point clouds challenges fully supervised methods. Addressing this, our study extends into semi-supervised learning for LiDAR semantic segmentation, leveraging the intrinsic spatial priors of driving scenes and multi-sensor complements to augment the efficacy of unlabeled datasets. We introduce \textbf{LaserMix++}, an evolved framework that integrates laser beam manipulations from disparate LiDAR scans and incorporates LiDAR-camera correspondences to further assist data-efficient learning. Our framework is tailored to enhance 3D scene consistency regularization by incorporating multi-modality, including 1) multi-modal LaserMix operation for fine-grained cross-sensor interactions; 2) camera-to-LiDAR feature distillation that enhances LiDAR feature learning; and 3) language-driven knowledge guidance generating auxiliary supervisions using open-vocabulary models. The versatility of LaserMix++ enables applications across LiDAR representations, establishing it as a universally applicable solution. Our framework is rigorously validated through theoretical analysis and extensive experiments on popular driving perception datasets. Results demonstrate that LaserMix++ markedly outperforms fully supervised alternatives, achieving comparable accuracy with five times fewer annotations and significantly improving the supervised-only baselines. This substantial advancement underscores the potential of semi-supervised approaches in reducing the reliance on extensive labeled data in LiDAR-based 3D scene understanding systems.
}

\metadata[
\raisebox{-0.2em}{\includegraphics[width=0.025\linewidth]{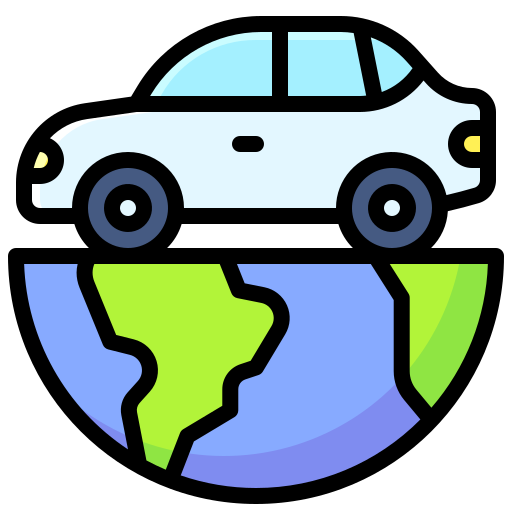}}~~Project Page]{\href{https://ldkong.com/LaserMix}{\texttt{https://ldkong.com/LaserMix}}
\\[-1.5ex]}

\metadata[
\raisebox{-0.2em}{\includegraphics[width=0.025\linewidth]{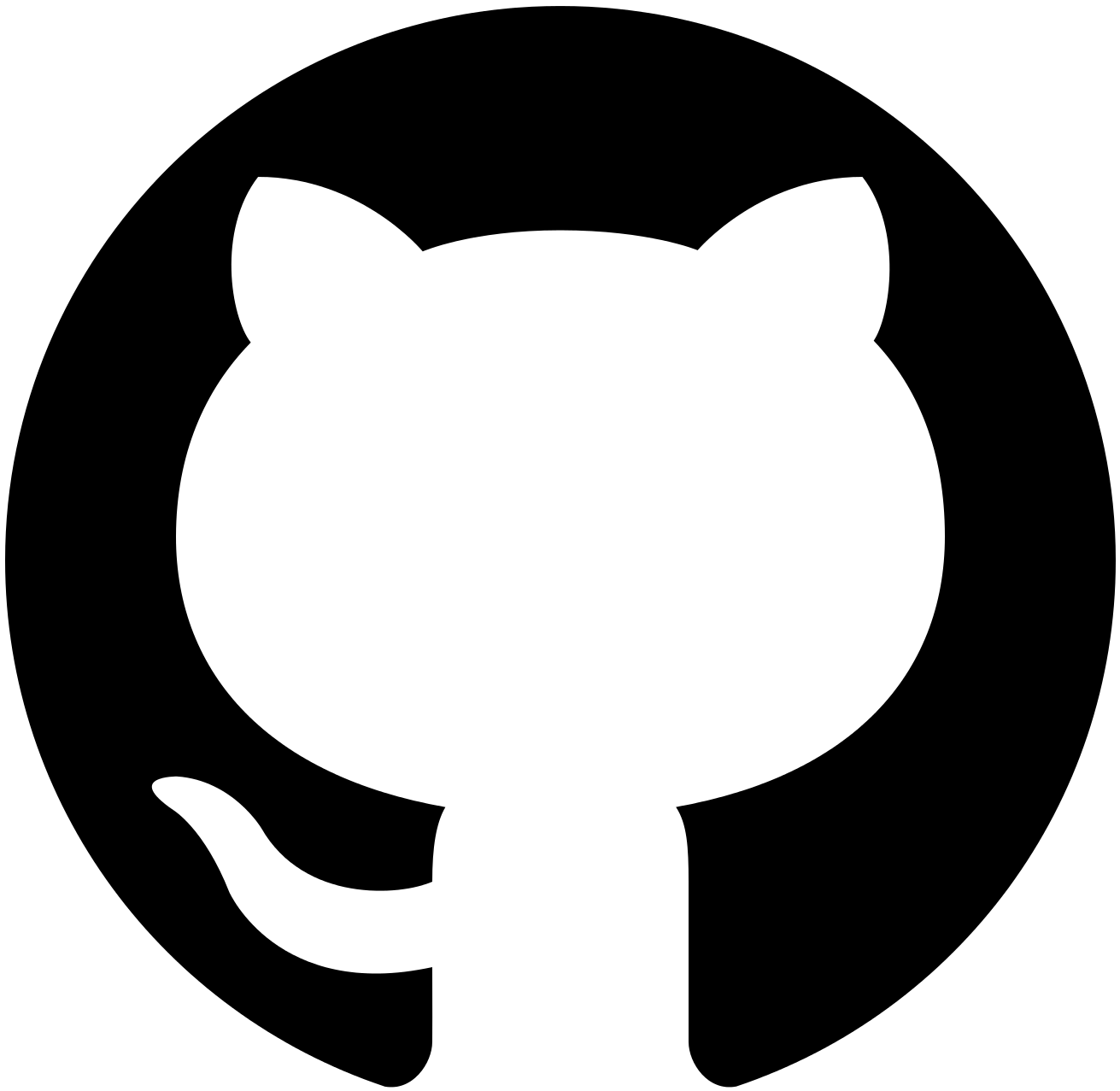}}~~GitHub Repo]{\href{https://github.com/ldkong1205/LaserMix}{\texttt{https://github.com/ldkong1205/LaserMix}}}

\begin{document}

\maketitle

\section{Introduction}
\label{sec:introduction}
LiDAR segmentation stands as a cornerstone task in the domain of autonomous driving perception, essential for vehicles to effectively understand the dense 3D structure of their surrounding environment \cite{Survey-LiDAR,Geiger2012CVPR}. This capability is fundamental for safe navigation and interaction with complex and dynamic environments \cite{nunes2022segcontrast,yan2024survey,Autonomous-Driving,kong2023robodepth}.

However, the requirement for extensive manual annotation of LiDAR point clouds imposes significant costs and logistical challenges, which severely limits the scalability of fully supervised learning methods in real-world applications \cite{ScribbleKITTI,SQN,ConDA,kong2023laserMix,li23lim3d,LESS}. Given these constraints, semi-supervised learning emerges as a promising solution that leverages the abundance of readily available unlabeled data to reduce dependence on costly human annotations while still maintaining satisfactory perception accuracy \cite{Survey-LiDAR-Data-Hungry,gebrehiwot2022teachers}.

Traditional semi-supervised learning approaches have predominantly focused on image-based tasks, whereas methods like MixMatch \cite{MixMatch}, FixMatch \cite{FixMatch}, and others \cite{ReMixMatch,MeanTeacher,CCT,GCT,CPS} have shown considerable success. However, these methods often underperform when directly applied to LiDAR data due to the inherent differences between the RGB image data and LiDAR point clouds \cite{kong2023laserMix,li23lim3d}. LiDAR data encapsulates rich geometric and topological information which presents unique challenges and opportunities for semi-supervised learning. For instance, the spatial distribution of points in a LiDAR scan directly corresponds to the physical layout of the environment, offering robust cues that are absent in traditional 2D images \cite{kong2023laserMix}.

Despite recent efforts in adapting semi-supervised learning for 3D data, most existing methodologies fail to fully exploit the synergistic potential of combining LiDAR with other sensor modalities \cite{GPC,DetMatch,Offboard-3D-OD,liu2023hssda}. This underutilization represents a missed opportunity, particularly in autonomous driving systems equipped with multiple types of sensors, including cameras and radar, alongside LiDAR \cite{nuScenes,sun2020waymoOpen,SemanticKITTI}. Each sensor type provides complementary information that can enhance the model's understanding of its environment, particularly under varying operational conditions such as low light or adverse weather \cite{yeong2021survey,chitta2023transfuser,kong2023robo3D,xie2023robobev}.

\begin{figure}[t]
    \centering
    \begin{subfigure}[b]{0.356\textwidth}
        \centering
        \includegraphics[width=\textwidth]{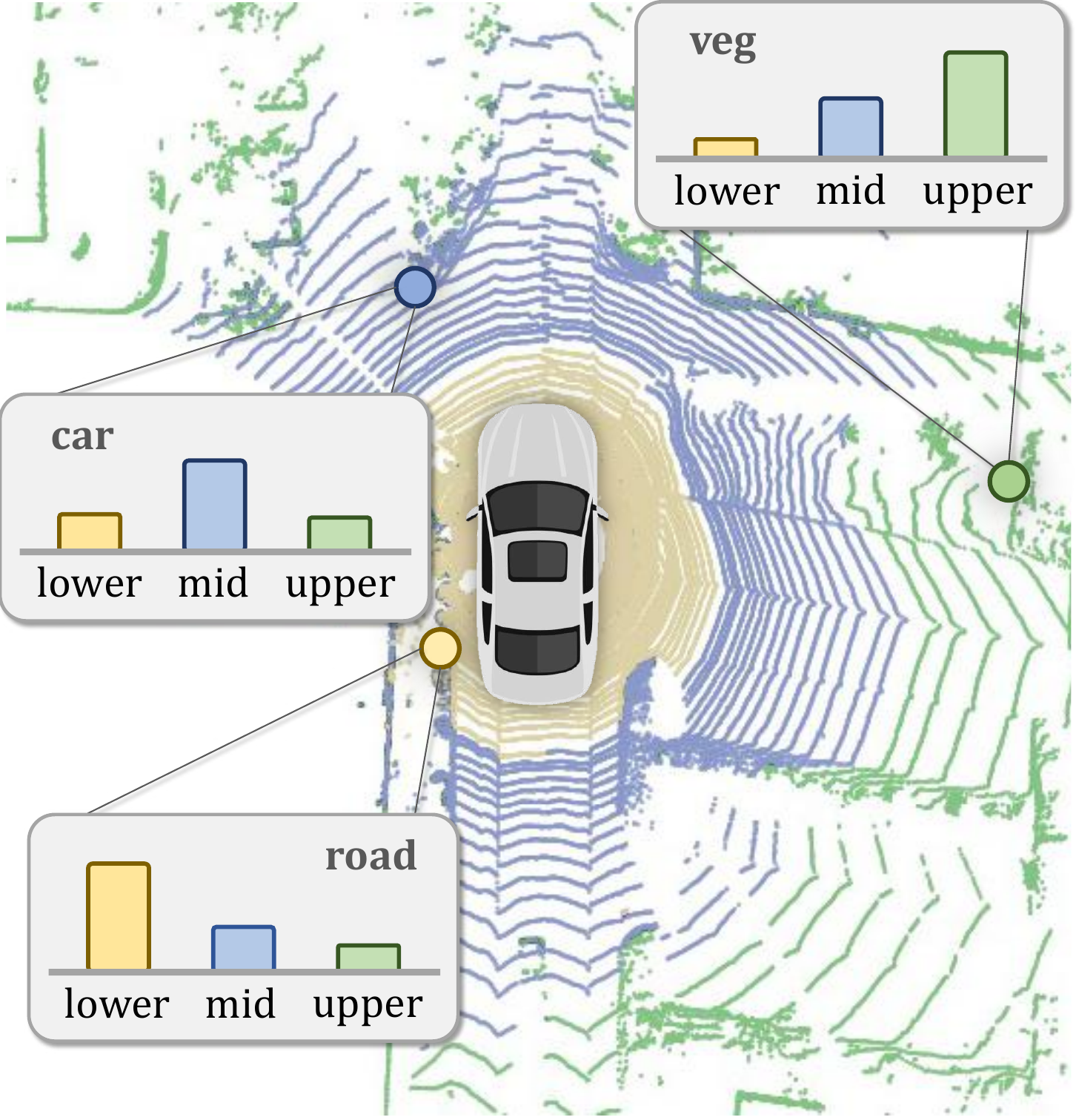}
        \caption{Spatial Prior}
        \label{fig:teaser-prior}
    \end{subfigure}~
    \begin{subfigure}[b]{0.265\textwidth}
        \centering
        \includegraphics[width=\textwidth]{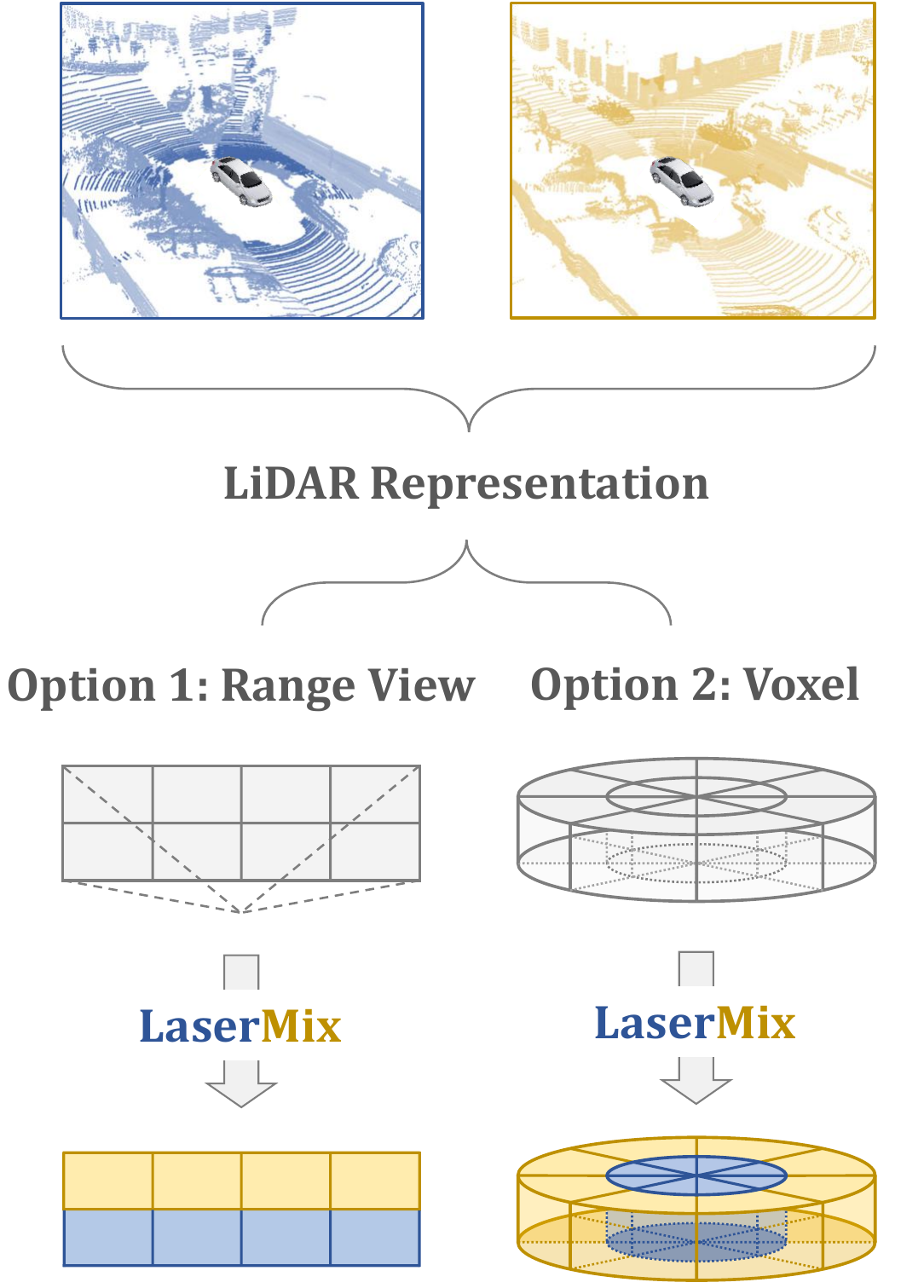}
        \caption{LiDAR Modality}
        \label{fig:teaser-modal}
    \end{subfigure}~
    \begin{subfigure}[b]{0.347\textwidth}
        \centering
        \includegraphics[width=\textwidth]{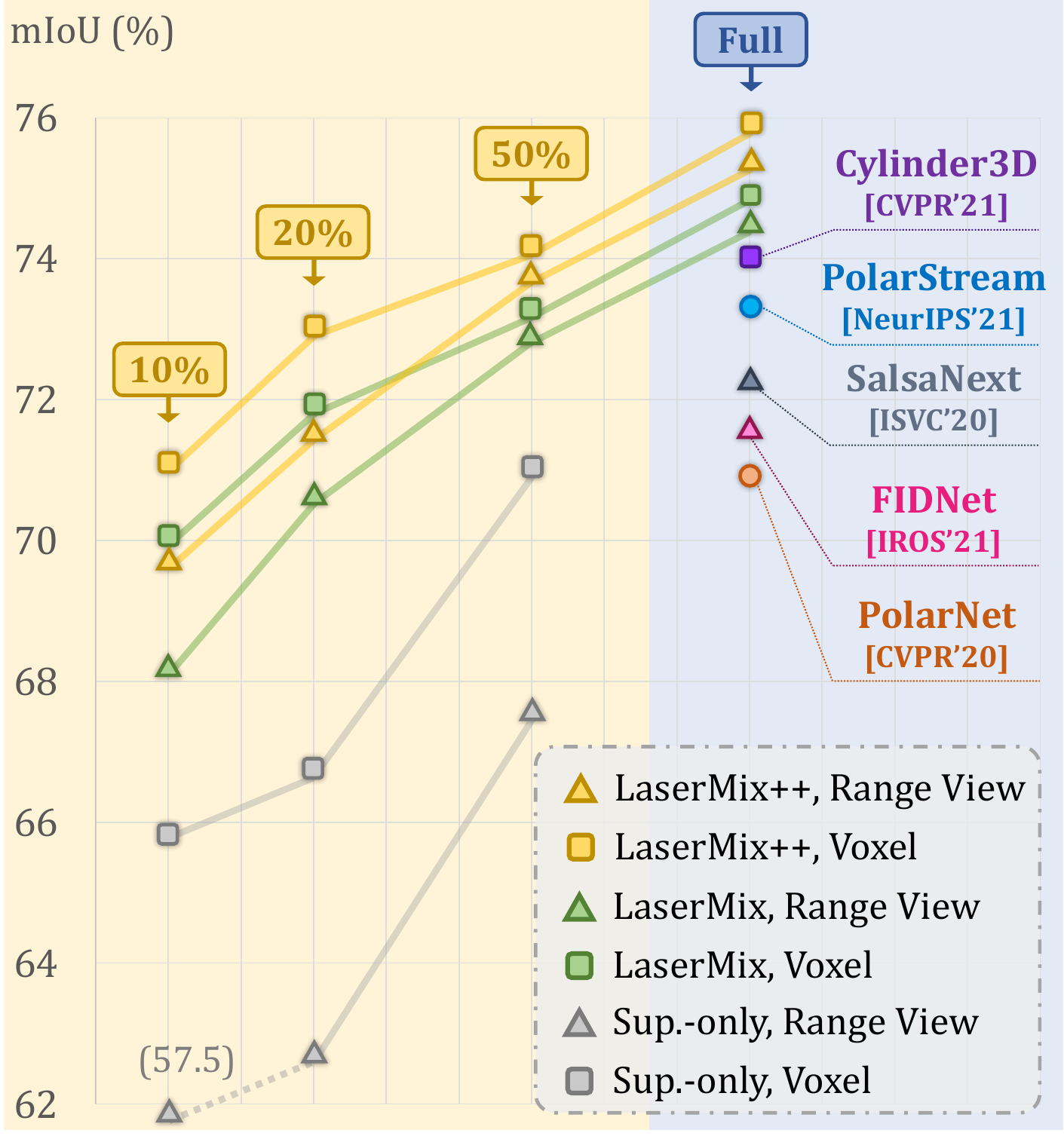}
        \caption{Performance Overview}
        \label{fig:teaser-performance}
    \end{subfigure}
    \caption{\textbf{Motivation.} \textbf{(a)} We observe a strong spatial prior from LiDAR-acquired driving scenes, where objects and backgrounds around the ego-vehicle have a patterned distribution on different (lower, middle, upper) laser beams. \textbf{(b)} The proposed laser beam mixing technique is agnostic to different LiDAR modalities and can be universally applied to existing LiDAR segmentation backbones. \textbf{(c)} Our approaches achieved superior performance than state-of-the-art methods \cite{Cylinder3D,PolarStream,SalsaNext,FIDNet,PolarNet} under low-data (10\%, 20\%, 50\% labels) and high-data (full supervision) regimes on nuScenes \cite{Panoptic-nuScenes}.}
    \label{fig:teaser}
\end{figure}

Building on this premise, this work introduces \textbf{LaserMix++}, an advanced data-efficient 3D scene understanding framework that expands the semi-supervised learning paradigm to incorporate multi-modal data integration. The baseline LaserMix \cite{kong2023laserMix} is a single-modal framework that leverages spatial priors inherent in LiDAR data (see Figure~\ref{fig:teaser-prior}) by mixing laser beams from LiDAR scans to enhance the consistency and confidence of predictions across unlabeled datasets. This approach utilized the geometric distribution of LiDAR-acquired driving scenes to infer the scene semantics with minimal supervision, setting a seminar yet strong benchmark in 3D scene understanding \cite{kong2023laserMix}.

Expanding upon this foundation, LaserMix++ integrates additional sensor modalities, specifically camera data, to address the complexities of autonomous driving environments more comprehensively. While the original LaserMix \cite{kong2023laserMix} focused on utilizing spatial continuity within LiDAR data to improve semantic understanding, LaserMix++ builds on this by incorporating textural and contextual information from camera images. By harnessing both LiDAR and camera inputs, LaserMix++ aims to exploit the complementary nature of spatial priors from LiDAR and textural details from camera images. This multi-modal approach enhances the model’s ability to generalize across different scenarios, particularly in conditions where supervision signals are not sufficiently available \cite{kong2023laserMix,li23lim3d,ScribbleKITTI}. LaserMix++ introduces three novel components to achieve better multi-modal integration:

\begin{itemize}
    \item \textbf{Multi-Modal LaserMix Operation:} We extend the original LaserMix \cite{kong2023laserMix} to include camera images, allowing the model to process, mix, and manipulate information from both LiDAR point clouds and their corresponding camera images. This fusion not only enriches the feature set but also aligns spatial and textural data, enhancing the descriptive power of the spatial priors.

    \item \textbf{Camera-to-LiDAR Feature Distillation:} Leveraging recent endeavors in image segmentation, we propose to extract semantically rich features from camera images and integrate them into the LiDAR data processing stream. This method aims to effectively bridge the gap between 2D image data and 3D point clouds, aligning with the core principle of LaserMix \cite{kong2023laserMix} by enhancing the scene consistency across different modalities. This process helps in enhancing the data-efficient learning of the LiDAR point cloud data, particularly by improving the feature representation in environments where LiDAR data annotation alone is insufficient.

    \item \textbf{Language-Driven Knowledge Guidance:} Drawing on recent advancements in vision-language models \cite{radford2021clip}, we aim to utilize open-vocabulary descriptions to provide contextual cues that assist in the data-efficient learning process. By generating auxiliary labels through these models, LaserMix++ can provide additional supervisory signals to the semi-supervised learning framework, further refining and improving the model's predictions.
\end{itemize}

By integrating these multi-modal components, LaserMix++ retains the strengths of the original LaserMix \cite{kong2023laserMix} -- such as leveraging spatial priors from LiDAR -- while adding the ability to incorporate and align additional sensor data. This combination not only addresses the limitations of using single-modality data but also boosts the robustness and accuracy of the semi-supervised LiDAR segmentation model. Each step in the expansion of our LaserMix++ framework builds logically on the last, ensuring that the enhancements contribute meaningfully to the overall effectiveness of the multi-modal data-efficient 3D scene understanding system.

Our extensive validations of LaserMix++ on prominent multi-modal driving perception datasets, such as nuScenes \cite{Panoptic-nuScenes}, SemanticKITTI \cite{SemanticKITTI}, and ScribbleKITTI \cite{ScribbleKITTI}, confirms its effectiveness and superiority. Despite the simplicity of the pipeline, LaserMix++ not only meets but often exceeds the performance of fully supervised methods while requiring significantly fewer human annotations. Moreover, LaserMix++ directly operates on LiDAR point clouds so as to be agnostic to different LiDAR representations (see Figure~\ref{fig:teaser-modal}), \emph{e.g.}, range view~\cite{RangeNet++}, bird's eye view \cite{PolarNet}, sparse voxel~\cite{Cylinder3D}, and multi-view fusion \cite{SPVNAS}. The special property marks LaserMix++ as a universally applicable solution. Besides, the substantial reduction in the need for labeled data, coupled with the ability to integrate and leverage multi-modal inputs, underscores the potential of more scalable semi-supervised approaches in the context of LiDAR-based 3D scene understanding.

By providing a robust solution to the challenges of data annotation and sensor data utilization in autonomous vehicles, LaserMix++ sets a new standard for data-efficient learning in the field. As shown in Figure~\ref{fig:teaser-performance}, our approaches exemplify how the integration of multi-modal data can lead to promising improvements in the reliability and efficiency of autonomous driving perception technologies. To sum up, this work consists of key contributions as follows:
\begin{itemize}
    \item We present LaserMix++, a novel data-efficient 3D scene understanding framework that integrates LiDAR and camera data to enhance feature learning through textural and spatial synergies, tailored to improve model interpretation under various low-data regimes.

    \item Building on cross-sensor data integration, we introduce two pivotal enhancements: camera-to-LiDAR feature distillation and language-driven knowledge guidance. These components work together to generate robust auxiliary signals that enrich the training data without the need for additional annotations.
    
    \item Our approaches are rigorously formulated to leverage spatial cues in LiDAR data effectively, facilitating semi-supervised learning and ensuring that our methodology is both practical and theoretically sound.
    
    \item  Extensively validated against state-of-the-art methods, LaserMix++ demonstrates significant performance improvements across both low- and high-data regimes, underscoring the potential to revolutionize data-efficient 3D scene understanding in a more unified manner.
\end{itemize}

\section{Related Work}
\label{sec:related_work}
This section provides a literature review of works that are closely related to data-efficient 3D scene understanding.

\subsection{3D Scene Understanding}
The task of 3D scene understanding via LiDAR data is fundamental for various applications, especially autonomous driving \cite{muhammad2022survey,kong2024calib3d,li2024place3d}, robotics \cite{jiang2021rellis3D}, and mixed reality \cite{naseer2018survey}. Several methodologies have been employed to address the challenges of LiDAR scene segmentation, categorized mainly by the type of data representation: range view \cite{RangeNet++,SqueezeSegV3,SalsaNext,FIDNet,kong2023rethinking,ando2023rangevit,xu2023frnet}, bird's eye view \cite{PolarNet,zhou2021panoptic}, sparse voxel \cite{choy2019minkowski,SPVNAS,Cylinder3D,hong20224dDSNet}, and multi-view fusion \cite{AMVNet,RPVNet,liu2023uniseg}. While these fully-supervised approaches have achieved significant milestones, their reliance on extensive and meticulously annotated datasets poses a challenge. This dependency on large-scale annotations results in diminished performance when data is scarce \cite{Survey-LiDAR-Data-Hungry}. To address this problem, innovations in weak \cite{SQN,PSD}, scribble \cite{ScribbleKITTI}, and box \cite{Box2Seg} supervisions, along with active learning techniques \cite{LiDAR-SSL-2019,LESS}, have been proposed to mitigate the high costs of LiDAR data annotation. Our approach extends these efforts by leveraging semi-supervised learning to effectively utilize unlabeled LiDAR scans, thus enhancing the robustness and reducing the reliance on extensive labeled datasets for training effective models.

\subsection{Data-Efficient Learning in 2D}
The domain of 2D image processing has seen considerable success in applying semi-supervised learning techniques to reduce the need for labeled data. Foundational algorithms such as Pi-Model \cite{Pi-Model}, Mean Teacher \cite{MeanTeacher}, and various mix-based methods like MixMatch \cite{MixMatch}, ReMixMatch \cite{ReMixMatch}, and FixMatch \cite{FixMatch} have set benchmarks in image recognition tasks. Notably, in semantic segmentation, methods like CutMix-Seg \cite{CutMix-Seg} and PseudoSeg \cite{PseudoSeg} alter input data to strategically position decision boundaries in less dense areas of the label space. Consistency-based methods such as CPS \cite{CPS} and GCT \cite{GCT} enforce model stability between modified network outputs \cite{PS-MT}. While these perturbations and consistency enforcements show promise in 2D tasks, their efficacy diminishes when directly applied to the 3D domain due to its inherent complexity and data representation challenges \cite{Simple-Baseline,ClassMix,Strong-Weak-Net}. Techniques that focus on entropy minimization, like CBST \cite{CBST} and ST++ \cite{ST++}, generate pseudo-labels to facilitate self-training, though they may introduce considerable storage demands when scaled to large LiDAR datasets \cite{nuScenes,SemanticKITTI,Panoptic-nuScenes,sun2020waymoOpen}. Our proposed LaserMix++ framework builds upon these principles, adapting and enhancing them to maintain scalability and efficiency in 3D environments without the need for significant computational resources.

\subsection{Data-Efficient Learning in 3D}
While semi-supervised learning has been extensively explored in 2D contexts, its application to 3D data, especially outdoor LiDAR point clouds, is less mature. Most existing studies focus on semi-supervised learning techniques for object-centric point clouds \cite{SRN,SSL-ShapeSeg} or indoor scenes \cite{Superpoint-SemiSeg,SSPC-Net,nunes2022segcontrast,xie2021exploring}, which typically do not encounter the scale and variability presented in outdoor environments \cite{nuScenes,SemanticKITTI,kong2024openess}. Some efforts \cite{TGNN,DetMatch,Offboard-3D-OD,liu2023hssda} have been made to apply semi-supervised strategies to 3D object detection using LiDAR data. For 3D scene understanding, GPC \cite{GPC} explores semi-supervised point cloud semantic segmentation through contrastive learning but remains focused on indoor point clouds, thus not fully addressing the unique properties of outdoor LiDAR data. LaserMix \cite{kong2023laserMix} establishes the first benchmark for semi-supervised LiDAR segmentation based on large-scale driving datasets \cite{Panoptic-nuScenes,SemanticKITTI,ScribbleKITTI}. It employs a dual-branch framework to encourage consistency in predictions from LiDAR scans before and after laser-wise mixing. The subsequent work, LiM3D \cite{li23lim3d}, proposes to reduce the spatiotemporal redundancy and split the most informative data as the labeled set, resulting in improved performance. In this work, we address the limitations of these previous single-modality methods by integrating cross-sensor data. We present LaserMix++ to enhance feature learning through fine-grained LiDAR and camera data synergies, exhibiting a stronger performance in both high- and low-data regimes.

\subsection{Multi-Modal Driving Perception}
Advanced driving perception systems are equipped with a combination of versatile sensors of different types \cite{nuScenes,sun2020waymoOpen}. Prevailing sensor configurations often include one or more LiDARs, multiple RGB cameras covering surrounding views, as well as radar, IMU, GPS, \emph{etc}. The data acquired by different sensors tend to complement each other, further enhancing the resilience of the perception system \cite{yeong2021survey,chitta2023transfuser,yan2024survey}. Recently, several works have explored the integration of LiDAR and cameras for driving perception. SLidR \cite{sautier2022slidr}, Seal \cite{liu2023segment}, and ScaLR \cite{puy2024scalr} establish pretraining objectives using the image-to-LiDAR correspondence. xMUDA \cite{jaritz2020xmuda} and the subsequent works \cite{xu2024visual,jaritz2023cross} propose to leverage image data for unsupervised domain adaptation of LiDAR segmentation models in cross-domain scenarios. CLIP2Scene \cite{chen2023clip2Scene}, OpenScene \cite{peng2023openscene}, and CNS \cite{chen2023towards} utilize CLIP \cite{radford2021clip} to generate open-vocabulary predictions on LiDAR point clouds. Most recently, M3Net \cite{liu2024m3net} introduces a unified multi-dataset training framework using the image modality to bridge heterogeneous LiDAR data acquired from different datasets. Motivated by these endeavors, in this work, we pursue the integration of LiDAR and camera data for data-efficient 3D scene understanding. The proposed LaserMix++ framework consists of camera-to-LiDAR feature distillation and language-driven knowledge guidance modules tailored to generate auxiliary supervisions for unlabeled data. These components synergize to achieve state-of-the-art performance across various benchmarks.
\section{Data-Efficient 3D Scene Understanding}
\label{sec:methods}
In this section, we first introduce the spatial prior of LiDAR-based driving scenes (Section~\ref{sec:ssl-framework}). We then present LaserMix, which strives to efficiently encourage confident and consistent LiDAR predictions (Section~\ref{sec:lasermix}). Finally, we establish a strong 3D scene consistency regularization baseline (Section~\ref{sec:consistency_framework}).

\subsection{Spatial Prior in 3D Scene Understanding}
\label{sec:ssl-framework}

\noindent\textbf{Spatial Prior Observation.}
Understanding and utilizing the spatial distribution inherent in LiDAR scenes is pivotal for enhancing semi-supervised learning. LiDAR point clouds offer unique spatial priors that are not prevalent in 2D images. As shown in Table~\ref{tab:spatial-prior}, each LiDAR-acquired semantic class poses unique distribution patterns that directly reflect real-world driving scenes. Our methodology leverages these priors by encouraging the model to maintain consistent predictions across varying LiDAR data manipulations.

\begin{table}[t]
\caption{A case study on the \textbf{strong spatial prior} of representative semantic classes from the SemanticKITTI~\cite{SemanticKITTI} dataset. For each class, we show its type (static or dynamic), proportion (valid number of points in percentage), distribution among eight areas ($A=\{a_1, a_2, ..., a_8\}$, \emph{i.e.}, eight laser beam groups), and the heatmap in range view (lighter colors correspond to areas that have a higher likelihood to appear and vice versa). Best viewed in colors.}
\vspace{-0.1cm}
\centering\scalebox{1}{
\begin{tabular}{c|c|c|c|c}
\toprule
\textbf{Class} & \textbf{Type} & \textbf{Proportion} & \textbf{Distribution} & \textbf{Heatmap}
\\\midrule\midrule
vegetation & static & $24.825\%$ & \begin{minipage}[b]{0.23\columnwidth}\centering\raisebox{-.4\height}{\includegraphics[width=\linewidth]{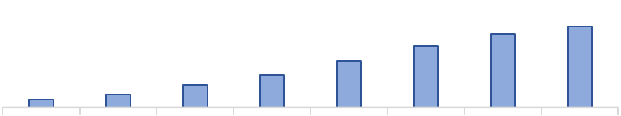}}\end{minipage} & \begin{minipage}[b]{0.32\columnwidth}\centering\raisebox{-.4\height}{\includegraphics[width=\linewidth]{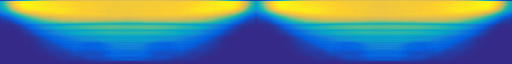}}\end{minipage}
\\\midrule
road & static & $22.545\%$ & \begin{minipage}[b]{0.23\columnwidth}\centering\raisebox{-.4\height}{\includegraphics[width=\linewidth]{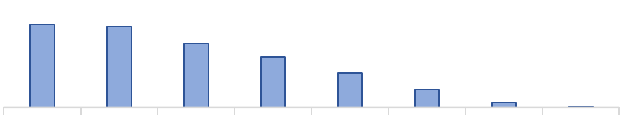}}\end{minipage} & \begin{minipage}[b]{0.32\columnwidth}\centering\raisebox{-.4\height}{\includegraphics[width=\linewidth]{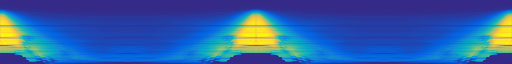}}\end{minipage}
\\\midrule
sidewalk & static & $16.353\%$ & \begin{minipage}[b]{0.23\columnwidth}\centering\raisebox{-.4\height}{\includegraphics[width=\linewidth]{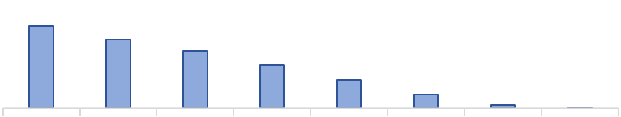}}\end{minipage} & \begin{minipage}[b]{0.32\columnwidth}\centering\raisebox{-.4\height}{\includegraphics[width=\linewidth]{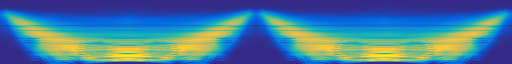}}\end{minipage}
\\\midrule
building & static & $12.118\%$ & \begin{minipage}[b]{0.23\columnwidth}\centering\raisebox{-.4\height}{\includegraphics[width=\linewidth]{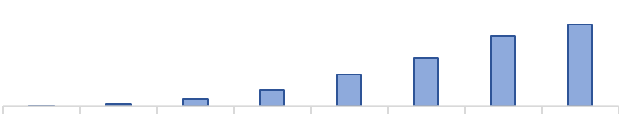}}\end{minipage} & \begin{minipage}[b]{0.32\columnwidth}\centering\raisebox{-.4\height}{\includegraphics[width=\linewidth]{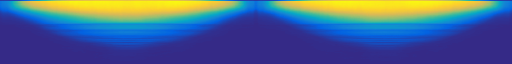}}\end{minipage}
\\\midrule
terrain & static & $8.122\%$ & \begin{minipage}[b]{0.23\columnwidth}\centering\raisebox{-.4\height}{\includegraphics[width=\linewidth]{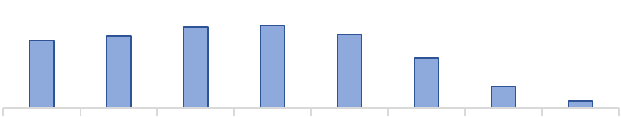}}\end{minipage} & \begin{minipage}[b]{0.32\columnwidth}\centering\raisebox{-.4\height}{\includegraphics[width=\linewidth]{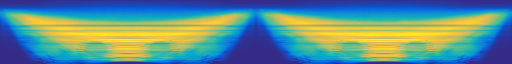}}\end{minipage}
\\\midrule
fence & static & $7.827\%$ & \begin{minipage}[b]{0.23\columnwidth}\centering\raisebox{-.4\height}{\includegraphics[width=\linewidth]{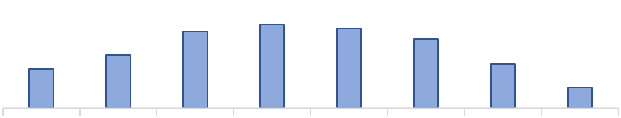}}\end{minipage} & \begin{minipage}[b]{0.32\columnwidth}\centering\raisebox{-.4\height}{\includegraphics[width=\linewidth]{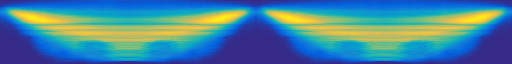}}\end{minipage}
\\\midrule
car & dynamic & $4.657\%$ & \begin{minipage}[b]{0.23\columnwidth}\centering\raisebox{-.4\height}{\includegraphics[width=\linewidth]{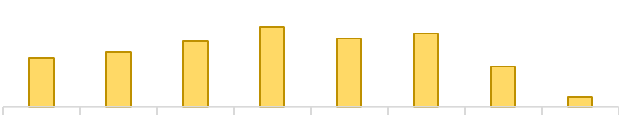}}\end{minipage} &
\begin{minipage}[b]{0.32\columnwidth}\centering\raisebox{-.4\height}{\includegraphics[width=\linewidth]{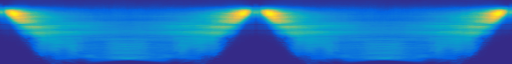}}\end{minipage}
\\\midrule
parking & static & $1.681\%$ & \begin{minipage}[b]{0.23\columnwidth}\centering\raisebox{-.4\height}{\includegraphics[width=\linewidth]{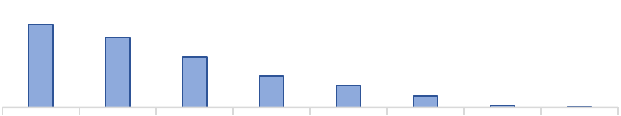}}\end{minipage} &
\begin{minipage}[b]{0.32\columnwidth}\centering\raisebox{-.4\height}{\includegraphics[width=\linewidth]{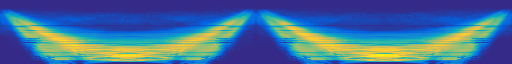}}\end{minipage}
\\\midrule
trunk & static & $0.580\%$ & \begin{minipage}[b]{0.23\columnwidth}\centering\raisebox{-.4\height}{\includegraphics[width=\linewidth]{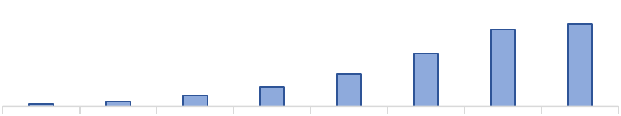}}\end{minipage} & \begin{minipage}[b]{0.32\columnwidth}\centering\raisebox{-.4\height}{\includegraphics[width=\linewidth]{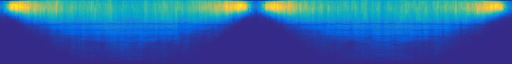}}\end{minipage}
\\\midrule
pole & static & $0.296\%$ & \begin{minipage}[b]{0.23\columnwidth}\centering\raisebox{-.4\height}{\includegraphics[width=\linewidth]{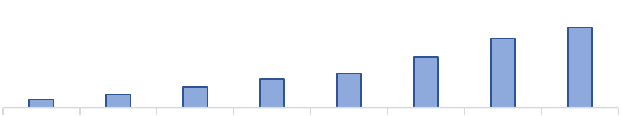}}\end{minipage} & \begin{minipage}[b]{0.32\columnwidth}\centering\raisebox{-.4\height}{\includegraphics[width=\linewidth]{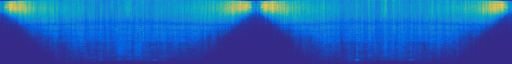}}\end{minipage}
\\\midrule
truck & dynamic & $0.193\%$ & \begin{minipage}[b]{0.23\columnwidth}\centering\raisebox{-.4\height}{\includegraphics[width=\linewidth]{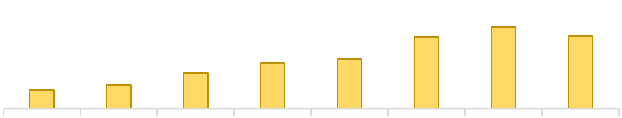}}\end{minipage} & \begin{minipage}[b]{0.32\columnwidth}\centering\raisebox{-.4\height}{\includegraphics[width=\linewidth]{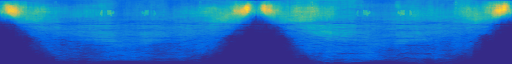}}\end{minipage}
\\\midrule
traffic-sign & static & $0.061\%$ & \begin{minipage}[b]{0.23\columnwidth}\centering\raisebox{-.4\height}{\includegraphics[width=\linewidth]{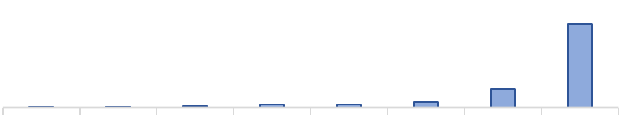}}\end{minipage} & \begin{minipage}[b]{0.32\columnwidth}\centering\raisebox{-.4\height}{\includegraphics[width=\linewidth]{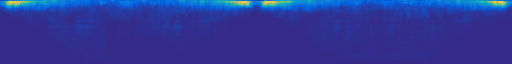}}\end{minipage}
\\\midrule
motorcycle & dynamic & $0.045\%$ & \begin{minipage}[b]{0.23\columnwidth}\centering\raisebox{-.4\height}{\includegraphics[width=\linewidth]{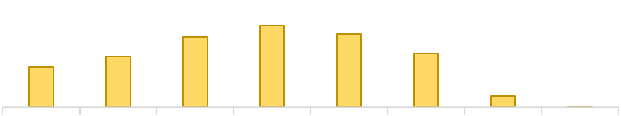}}\end{minipage} & \begin{minipage}[b]{0.32\columnwidth}\centering\raisebox{-.4\height}{\includegraphics[width=\linewidth]{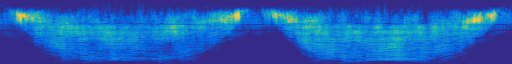}}\end{minipage}
\\\midrule
person & dynamic & $0.036\%$ & \begin{minipage}[b]{0.23\columnwidth}\centering\raisebox{-.4\height}{\includegraphics[width=\linewidth]{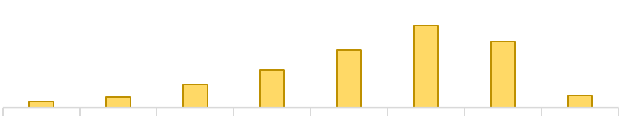}}\end{minipage} & \begin{minipage}[b]{0.32\columnwidth}\centering\raisebox{-.4\height}{\includegraphics[width=\linewidth]{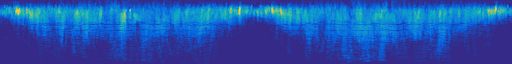}}\end{minipage}
\\\midrule
bicycle & dynamic & $0.018\%$ & \begin{minipage}[b]{0.23\columnwidth}\centering\raisebox{-.4\height}{\includegraphics[width=\linewidth]{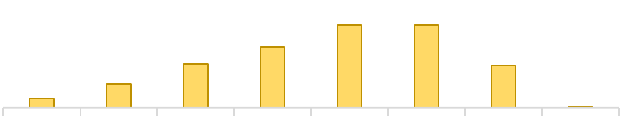}}\end{minipage} & \begin{minipage}[b]{0.32\columnwidth}\centering\raisebox{-.4\height}{\includegraphics[width=\linewidth]{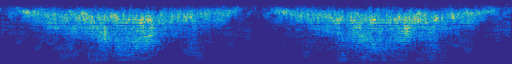}}\end{minipage}
\\
\bottomrule
\end{tabular}}
\label{tab:spatial-prior}
\end{table}

\noindent\textbf{Spatial Prior Formulation.}
The spatial distribution of objects and backgrounds within a LiDAR scene significantly influences their representations in the point cloud data. Objects and backgrounds at different distances and orientations from the sensor exhibit distinct spatial patterns and can be leveraged to reduce the prediction uncertainty in unlabeled scenarios. Specifically, we define a spatial area $a \in A$ where LiDAR points and their semantic labels inside this area (denoted as $\Xcrop$ and $\Ycrop$, respectively) exhibit lower variation. This is quantified by a smaller conditional entropy $H(\Xcrop, \Ycrop|A)$, where $A$ represents different spatial regions within the data.

\noindent\textbf{Entropy Minimization.}
Given the spatial prior, our objective is to minimize the entropy of predictions within predefined areas. Formally, we express the entropy condition as follows:
\begin{align}\label{eq:entropy}
    \ev_\theta[H(\Xcrop, \Ycrop|A)] =c\,,
\end{align}
where $c$ is a small constant and $\theta$ represents the model parameters. Similar to the classic entropy minimization framework \cite{grandvalet2004semi}, the constraint in Eq.~\ref{eq:entropy} is transformed into a probabilistic model where the model parameter distribution is guided by the principle of maximum entropy:
\begin{equation}
\begin{split}
\label{eq:structure_prior}
    P(\theta) \propto \exp(-\lambda H(\Xcrop, \Ycrop|A)) \propto \exp(-\lambda H(\Ycrop|\Xcrop, A))\,,
\end{split}
\end{equation}
where $\lambda>0$ acts as the Lagrange multiplier associated with the entropy constraint, which corresponds to a constant $c$. $H(\Xcrop| A)$ has been ignored for being independent of the model parameter $\theta$. Here, we consider Eq.~\ref{eq:structure_prior} as the formal formulation of the spatial prior and discuss how to empirically compute it in the following sections.

\noindent\textbf{Marginalization}.
To utilize the spatial prior defined in Eq.~\ref{eq:structure_prior}, we empirically compute the entropy $H(\Ycrop|\Xcrop, A)$ of the LiDAR points \emph{inside} area $A$ as follows:
\begin{equation}
\begin{split}\label{eq:empirical-entropy}
    \Hemp(\Ycrop|\Xcrop, A) = \evemp_{\Xcrop, \Ycrop, A}[\Pemp(\Ycrop|\Xcrop, A)\log \Pemp(\Ycrop|\Xcrop, A)]\,,
\end{split}
\end{equation}
where $\hat{.}$ denotes the empirical estimation. The end-to-end LiDAR segmentation model $\mathcal{G}_{\theta}$ (with parameters $\theta$) usually takes full-sized data as inputs during the inference. Therefore, to compute $\Pemp(\Ycrop|\Xcrop, A)$ in Eq.~\ref{eq:empirical-entropy}, we first pad the data \emph{outside} the area to obtain the full-sized data. Here, we denote the data \emph{outside} the area as $\Xcomp$; we then let the model infer $P(\Ycrop|\Xcrop, \Xcomp, A)$, and finally marginalize $\Xcomp$ as:
\begin{align}\label{eq:marginalization}
    \Pemp(\Ycrop|\Xcrop, A) = \evemp_{\Xcomp}[\Pemp(\Ycrop|\Xcrop, \Xcomp, A)]\,.
\end{align}
It is worth noting that the generative distribution of the padding $P(\Xcomp)$ can be directly obtained from the dataset.

\noindent\textbf{Training Objectives.}
Finally, we train the segmentation model $\mathcal{G}_{\theta}$ using the standard maximum-a-posteriori (MAP) estimation. We maximize the posterior that can be computed by Eq.~\ref{eq:structure_prior}, Eq.~\ref{eq:empirical-entropy}, and Eq.~\ref{eq:marginalization}, which is formulated as follows:
\begin{equation}
\label{eq:our-ssl}
\begin{split}
    &C(\theta) = L(\theta) - \lambda \Hemp(\Ycrop|\Xcrop, A) 
                = L(\theta) \\ &- \lambda\evemp_{\Xcrop, \Ycrop, A}[\Pemp(\Ycrop|\Xcrop, A)\log \Pemp(\Ycrop|\Xcrop, A)]\,.
\end{split}
\end{equation}
Here, $L(\theta)$ is the likelihood function which can be computed using labeled data, \emph{i.e.}, the conventional supervised learning. Minimizing $\Hemp(\Ycrop|\Xcrop, A)$ requires the marginal probability $P(\Ycrop|\Xcrop, A)$ to be confident, which further requires $P(\Ycrop|\Xcrop, \Xcomp, A)$ to be both confident and consistent with respect to different outside data $\Xcomp$. To sum up, our proposed semi-supervised learning framework in Eq.~\ref{eq:our-ssl} encourages the segmentation model to make confident and consistent predictions in a predefined area, regardless of the data outside the area. The predefined area set $A$ determines the ``strength'' of the prior. When setting $A$ to the full area (\emph{i.e.}, the whole point cloud), our framework degrades to the classic entropy minimization framework \cite{grandvalet2004semi}.

\noindent\textbf{Practical Implementations.}
Implementing our proposed prior-based semi-supervised learning framework effectively involves three critical steps, which are:
\begin{itemize}
    \item \emph{Step 1)}: Identify and select an appropriate partition set $A$ that encapsulates a strong spatial prior, which is essential for guiding the learning process;
    \item \emph{Step 2)}: Efficiently compute the marginal probability, \emph{i.e.}, $\Pemp(\Ycrop|\Xcrop, A)$, which is fundamental for understanding the distribution of labels within specified spatial regions;
    \item \emph{Step 3)}: Efficiently minimize the marginal entropy, represented as $\Hemp(\Ycrop|\Xcrop, A)$, to enhance the consistency and confidence of the predictions across unlabeled data.
\end{itemize}
A detailed and effective implementation of these steps is proposed in the subsequent section, showcasing their practical applicability and impact on the overall framework.

\subsection{LaserMix}
\label{sec:lasermix}

\begin{wrapfigure}{r}{0.58\textwidth}
    \begin{minipage}{\linewidth}
    \centering
    \vspace{-0.4cm}
    \includegraphics[width=\linewidth]{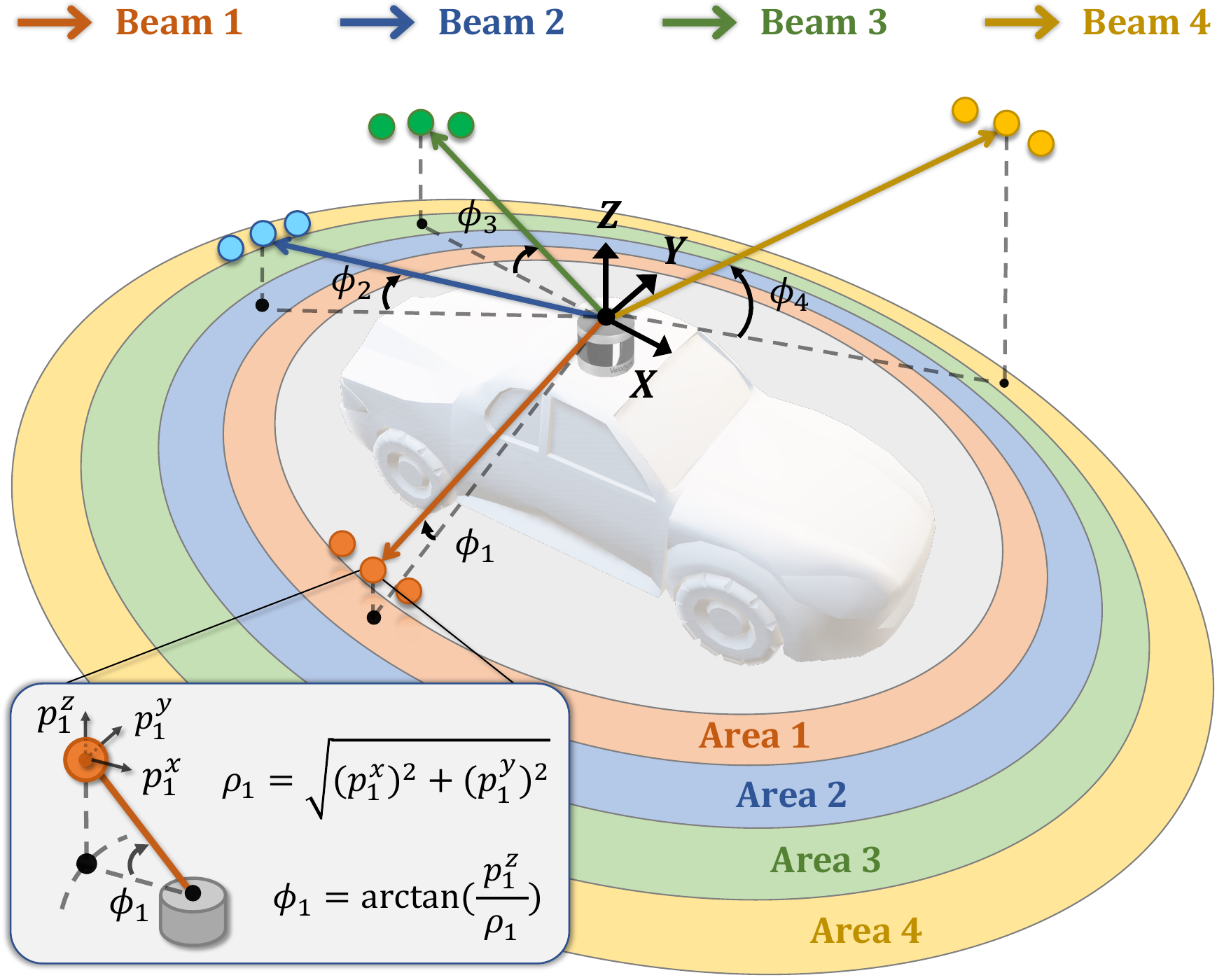}
    \vspace{-0.4cm}
    \caption{\textbf{Laser partition example}. We group LiDAR points $(p^x_i, p^y_i, p^z_i)$ whose inclinations $\phi_i$ are within the same inclination range into the same area, as depicted in color regions.}
    \label{fig:inclination}
    \end{minipage}
\end{wrapfigure}

\noindent\textbf{LiDAR Scene Partitions.}
The prevailing rotating LiDAR sensors deploy a fixed number (\emph{e.g.}, $32$, $64$, and $128$) of laser beams which are emitted isotropically around the ego-vehicle with predefined inclination angles as shown in Figure~\ref{fig:inclination}. To delineate distinct and proper spatial areas $A$, we propose to partition the LiDAR point cloud based on these laser beams. Specifically, each point captured by a particular laser beam aligns at a consistent inclination angle relative to the sensor plane. For point $i$, its inclination $\phi_i$ is calculated as follows:
\begin{align}
   \phi_i = \mathrm{arctan}(\frac{p^z_i}{\sqrt{(p^x_i)^2 + (p^y_i)^2 }})\,,
\end{align}
where $(p^x, p^y, p^z)$ represent the Cartesian coordinates of the LiDAR points. For any two LiDAR scans, $x_1$ and $x_2$, we first group all points from each scan by their inclination angles. More concretely, to establish $m$ non-overlapping areas, a set of $m+1$ inclination angles $\Phi=\{\phi_0, \phi_1, \phi_2, ..., \phi_m\}$ will be evenly sampled within the range of the minimum and maximum inclination angles in the dataset, and the area set $A=\{a_1, a_2, ..., a_m\}$ can then be confined by bounding area $a_i$ in the inclination range $[\phi_{i-1}, \phi_i)$. It is important to note that the range of inclination angles varies depending on the specific configurations of different LiDAR sensors.

\noindent\emph{Role in our framework:} The proposed laser-based partitioning aims to effectively ``excite'' a strong spatial prior in the LiDAR point cloud, as described by \emph{Step 1} in our semi-supervised learning framework. As shown in Figure~\ref{fig:teaser-prior} and Table~\ref{tab:spatial-prior}, this partitioning reveals clear distribution patterns in the semantic classes detected by each laser beam. Despite being an empirical choice, we will show in later sections that our laser-based partitioning method significantly outperforms other partition choices, including random points (\emph{MixUp}-like partition~\cite{MixUp}), random areas (\emph{CutMix}-like partition~\cite{CutMix}), and other heuristics like azimuth $\alpha$ (sensor horizontal direction) or radius $r$ (sensor range direction) partitions.

\noindent\textbf{LiDAR Scene Mixing.}
In the pursuit of refining the control over spatial priors, we introduce LaserMix, a sophisticated LiDAR mixing strategy tailored to optimize the manipulation of spatial data from multiple LiDAR scans. 
LaserMix mixes the aforementioned laser-partitioned areas $A$ from two scans in an intertwining way, \emph{i.e.}, one takes from odd-indexed areas $A_1=\{a_1, a_3, ...\}$ and the other takes from even-indexed areas $A_2=\{a_2, a_4, ...\}$, so that each area's neighbor will be from the other scan. More formally, we define the LaserMix operation as follows:
\begin{align}
\begin{split}
\tilde{x}_1, \tilde{x}_2 = \textrm{LaserMix}(x_1, x_2)\,, \\
\tilde{x}_1 = x_1^{a_1} \cup x_2^{a_2} \cup x_1^{a_3} \cup\cdots, \\
\tilde{x}_2 = x_2^{a_1} \cup x_1^{a_2} \cup x_2^{a_3} \cup\cdots, \\
\end{split}
\label{eq:lasermix}
\end{align}
where $x_i^{a_j}$ is the data crop of $x_i$ confined within area $a_j$. Correspondingly, the semantic labels are mixed in the same way as Eq.~\ref{eq:lasermix}. It is worth highlighting that LaserMix is directly applied to the LiDAR point clouds and is thus agnostic to the various LiDAR representations, such as range view \cite{RangeNet++}, bird's eye view \cite{PolarNet}, raw points \cite{RandLa-Net}, sparse voxel \cite{Cylinder3D}, and multi-view fusion \cite{SPVNAS} representations. This versatility allows LaserMix to be effectively integrated across a broad spectrum of existing LiDAR segmentation frameworks, without necessitating any modifications to the underlying data or network structures.
While LaserMix partitions the LiDAR point cloud for enhancing feature learning, concerns may arise regarding the integrity of object instances, such as cars, when these are split across different segments. However, given the large spatial area typically involved, the likelihood of significant object partitioning is low. Moreover, LaserMix's use of both ground truth and high-confidence pseudo labels provides robust supervision, ensuring accurate object interpretation even when partitioning occurs. This design allows LaserMix to leverage the benefits of data augmentation without compromising object integrity, thereby enhancing the robustness and generalization of the model.

\noindent\emph{Role in our framework:} Central to our semi-supervised learning framework, LaserMix streamlines the computation of marginal probability $P(\Ycrop|\Xcrop, A)$ -- a process otherwise computationally intensive in typical scenarios -- as described by \emph{Step 2} in our framework. The cost for directly computing the marginal probability in Eq.~\ref{eq:marginalization} on real-world LiDAR data is prohibitive; we need to iterate through all areas in $A$ and all outside data in $\Xcomp$, which requires $|A|\cdot|\Xcomp|$ predictions in total. To reduce the training overhead, we take advantage of the fact that a prediction in an area will be largely affected by its neighboring areas and let $\Xcomp$ fill only the neighbors instead of all the remaining areas. LaserMix mixes two scans by \emph{intertwining} the areas so that the neighbors of each area are filled with data from the other scan. As a result, we obtain the prediction on all areas $A$ of two scans from only two predictions, which on average reduces the cost from $|A|$ to $1$. The scan before and after mixing counts as two data fillings, therefore $|\Xcomp|=2$. Overall, the training overhead is reduced from $|A|\cdot|\Xcomp|$ to $2$: only one prediction on original data and one additional prediction on mixed data are required for each LiDAR scan. During training, the memory consumption for a batch will be $2\times$ compared to a standard semi-supervised learning framework, and the training speed will not be affected.

\begin{figure}[t]
    \begin{center}
    \includegraphics[width=\linewidth]{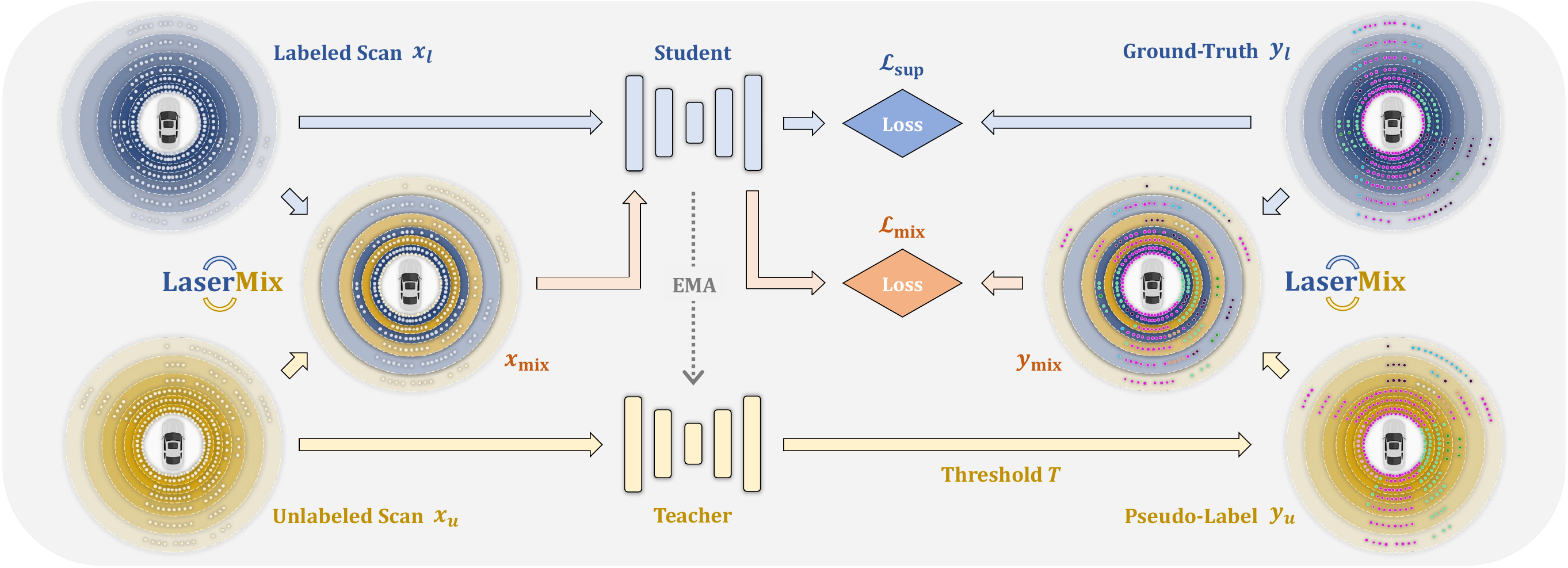}
    \end{center}
    \vspace{-0.2cm}
    \caption{\textbf{Overview of our baseline consistency regularization framework}. We feed the labeled scan $x_{l}$ into the Student network to compute the supervised loss $\mathcal{L}_{\text{sup}}$ (\emph{w/} ground truth $y_{l}$). The unlabeled scan $x_{u}$ and the generated pseudo-label $y_{u}$ are mixed with $(x_{l},y_{l})$ via LaserMix (Section~\ref{sec:lasermix}) to produce mixed data sample $(x_{\text{mix}},y_{\text{mix}})$, which is then fed into the Student network to compute the mixing loss $\mathcal{L}_{\text{mix}}$. Additionally, we encourage the consistency between the Student network and the Teacher network by computing the mean teacher loss $\mathcal{L}_{\text{mt}}$ over their predictions, where the Teacher network's parameters are updated by the exponential moving average of that of the Student network. During inference, only the Teacher network is needed, which maintains the same computational cost as the conventional LiDAR segmentation pipeline.}
    \label{fig:framework}
\end{figure}

\subsection{3D Scene Consistency Regularization}
\label{sec:consistency_framework}
\noindent\textbf{Dual-Branch Consistency.}
We design a consistency regularization framework based on our three-step procedures to enhance the data efficiency in 3D scene understanding, as depicted in Figure~\ref{fig:framework}. The framework incorporates a dual-branch architecture, consisting of a Student network $\mathcal{G}_{\theta}^{s}$ and a Teacher network $\mathcal{G}_{\theta}^{t}$. $\mathcal{G}_{\theta}^{s}$ and $\mathcal{G}_{\theta}^{t}$ take certain LiDAR representations (\emph{e.g.}, range images or voxel grids) as the input and make predictions. During the training phase, each batch contains an equal mix of labeled and unlabeled data. We collect the predictions from both $\mathcal{G}_{\theta}^{s}$ and $\mathcal{G}_{\theta}^{t}$, and generate the pseudo-labels from the Teacher network's predictions with a predefined confidence threshold $T$. For labeled data, the cross-entropy loss $\mathcal{L}_{\text{sup}}$ is calculated between the predictions of the Student network and the ground truth. Concurrently, LaserMix is applied to mix each unlabeled scan with a randomly selected labeled scan, together with their corresponding pseudo-labels and ground truth. Next, we use the Student network $\mathcal{G}_{\theta}^{s}$ to predict the mixed data and compute the cross-entropy loss $\mathcal{L}_{\text{mix}}$ (\emph{w/} mixed labels). 

\noindent\textbf{Optimization Objectives.}
The point-wise cross-entropy loss for a labeled or unlabeled LiDAR point cloud $x$ and its corresponding ground truth or pseudo-labels $y$ on the LiDAR segmentation network $\mathcal{G}_{\theta}$ is calculated as follows:
\begin{align}
    \mathcal{L}_{\text{ce}} = \frac{1}{|x|}\sum_{i=1}^{|x|}\text{CrossEntropy}(y^{(i)}, \mathcal{G}^{(i)}_\theta(x))\,,
\label{equ:sup}
\end{align}
where $(i)$ denotes the $i$-th point in the point cloud. Moreover, our framework is compatible with the mean teacher consistency regularization \cite{MeanTeacher}, where the exponential moving average (EMA) of the Student network $\mathcal{G}_{\theta}^{s}$ is used to update the parameters of Teacher network $\mathcal{G}_{\theta}^{t}$, along with a temperature coefficient. The $\ell2$ loss between the predictions from two networks, \emph{i.e.}, $\mathcal{L}_{\text{mt}}$, is calculated as follows:
\begin{align}
    \mathcal{L}_{\text{mt}} = \frac{1}{|x|}\sum_{i=1}^{|x|}||\mathcal{G}_{\theta}^{s,(i)}(x) - \mathcal{G}_{\theta}^{t,(i)}(x)||^2_2\,,
\label{equ:mt}
\end{align}
where $||\cdot||^2_2$ denotes the $\ell2$ norm. The overall loss function of this consistency regularization framework is $\mathcal{L} = \mathcal{L}_{\text{sup}} + \lambda_\textrm{mix} \mathcal{L}_{\text{mix}} + \lambda_\textrm{mt}\mathcal{L}_{\text{mt}}$,
where $\lambda_\textrm{mix}$ and $\lambda_\textrm{mt}$ are loss weights. The Teacher network $\mathcal{G}_{\theta}^{t}$ is used during inference due to its empirically observed stability. Compared to conventional fully supervised learning, there is no additional computational burden in the inference phase.

\noindent\emph{Role in our framework:} Our consistency regularization pipeline is designed to effectively minimize the marginal entropy for data-efficient 3D scene understanding, as described in \emph{Step 3} of our framework. Since the objective for minimizing the entropy has a challenging optimization landscape, pseudo-labeling emerges as a resort in practice~\cite{lee2013pseudo}. Unlike conventional pseudo-label optimization in semi-supervised learning, which only aims to encourage the predictions to be confident, minimizing the marginal entropy requires all predictions to be both confident and consistent. To this end, we leverage both ground truth and pseudo-labels as anchors to encourage the 3D scene understanding model's predictions to be confident and consistent with these supervision signals.

\section{LaserMix++}
\label{sec:lasermix++}

To leverage interactions between the camera and LiDAR, we first elaborate on the correspondence between LiDAR and cameras (Section~\ref{sec:2d-3d-correspond}). We then propose a multi-modal LaserMix operation (Section~\ref{sec:multi-modal-lasermix}), a cross-sensor feature distillation (Section~\ref{sec:c2l}), and language-driven knowledge guidance (Section~\ref{sec:lkg}). Finally, we describe the overall LaserMix++ pipeline for enhancing data-efficient 3D scene understanding (Section~\ref{sec:pipeline}).

Prevailing driving perception systems often consist of a spectrum of sensors to ensure safe operations \cite{nuScenes,sun2020waymoOpen,SemanticKITTI}. A typical sensor setup incorporated by existing perception systems contains both LiDAR and surrounding cameras. Both 2D (trained \emph{w/} RGB images) and 3D (trained \emph{w/} LiDAR point clouds) perception models are leveraged, which can complement each other, especially in challenging conditions such as low light, adverse weather, and motion perturbations \cite{yeong2021survey,chitta2023transfuser,kong2023robo3D,xie2023robobev}. In this work, instead of solely relying on LiDAR point clouds to train semi-supervised learning models, we propose to leverage the abundant RGB images from cameras as additional resources for a more holistic, multi-modal data-efficient 3D scene understanding. This enhanced framework, dubbed LaserMix++, does not require any image annotations, thus maintaining the same data efficiency as the baseline framework described in Section~\ref{sec:consistency_framework}.

\subsection{2D-3D Correspondences}
\label{sec:2d-3d-correspond}
Assuming that the driving perception system consists of one LiDAR and one camera sensor\footnote{This simple configuration can be easily extended to both single-LiDAR multi-camera and multi-LiDAR multi-camera scenarios.} that have been well calibrated and synchronized, we can obtain, at a certain data acquisition frequency, a pair of LiDAR point cloud $x_{\text{p}}=(p^x, p^y, p^z)$ and camera image $x_{\text{img}}$. To establish 2D-3D correspondences for a given pair $\{x_{\text{p}}, x_{\text{img}}\}$, we first project point cloud $x_{\text{p}}$ onto the image plane $(p^u, p^v) $ based on sensor calibration parameters:
\begin{equation}
    [p^u, p^v, 1]^{\text{T}} = \frac{1}{p^z} \times \Gamma_{K} \times \Gamma_{c \leftarrow l} \times [p^x, p^y, p^z]^{\text{T}}\,,
    \label{equ:projection}
\end{equation}
where $\Gamma_{K}$ denotes the camera intrinsic matrix and $\Gamma_{c \leftarrow l}$ is the transformation matrix from the LiDAR to the camera. 

It is worth mentioning that the correspondences between points and pixels do not always hold due to the possible mis-overlap between the field-of-view of the LiDAR and the camera. To handle this, we generate a correspondence mask $\mathcal{M}$ to establish valid point-pixel relationships. The process begins by projecting LiDAR points onto the image plane using the known extrinsic and intrinsic camera parameters, allowing us to determine which LiDAR points have corresponding pixels in the image. If a LiDAR point $x_\text{p}$ projects onto a valid pixel location, the corresponding entry in the mask $\mathcal{M}$ is set to $1$. For LiDAR points that project outside the image boundaries or onto invalid regions, the corresponding entry in $\mathcal{M}$ is set to $0$. These entries are then padded with zeros in the pixel correspondence data. This automated process leverages existing geometric transformations and does not require additional manual data annotation, ensuring scalability and efficiency.

\subsection{Multi-Modal LaserMix}
\label{sec:multi-modal-lasermix}
Different from the pure positional information encoded in the LiDAR point cloud, the image pixels provide extra texture information that could be supplementary to the 3D scene understanding task. To leverage such visual guidance for data-efficient learning, we associate the image pixels with the LiDAR points during LaserMix. Similar to Eq.~\ref{eq:lasermix}, such a multi-modal data mixing process between the LiDAR and camera can be defined as:

\begin{align}
\begin{split}
\tilde{\sigma}_1, \tilde{\sigma}_2 &= \textrm{Multi-Modal LaserMix}[ \{x_{\text{p}}, x_{\text{img}}\}_1, \{x_{\text{p}}, x_{\text{img}}\}_2 ]\,, \\
\tilde{\sigma}_1 &= \{x_{\text{p}}, x_{\text{img}}\}_1^{a_1} \cup \{x_{\text{p}}, x_{\text{img}}\}_2^{a_2} \cup \{x_{\text{p}}, x_{\text{img}}\}_1^{a_3} \cup\cdots, \\
\tilde{\sigma}_2 &= \{x_{\text{p}}, x_{\text{img}}\}_2^{a_1} \cup \{x_{\text{p}}, x_{\text{img}}\}_1^{a_2} \cup \{x_{\text{p}}, x_{\text{img}}\}_2^{a_3} \cup\cdots. \\
\end{split}
\label{eq:mm-lasermix}
\end{align}
The area set $\{ a_1, a_2, ..., a_m \}$ can be obtained in the same way as Section~\ref{sec:lasermix}. The multi-modal variant of LaserMix leverages off-the-shelf visual information from synchronized camera images to assist the consistency regularization in Section~\ref{sec:consistency_framework}. Such an approach further enhances the effect of spatial priors in driving scenes, since the LiDAR points have been ``painted'' with extra texture information from the images. The use of the correspondence mask $\mathcal{M}$ ensures that only valid LiDAR point-pixel pairs are utilized, maintaining the consistency of the multi-modal information. After obtaining mixed LiDAR-image pairs, we propose the following two operations in the LaserMix++ framework to assist the data-efficient 3D scene understanding task with such multi-modal inputs.

\subsection{Camera-to-LiDAR Feature Distillation}
\label{sec:c2l}
Over the past few years, image segmentation models have become both more efficient and highly accurate on large-scale benchmarks \cite{lin2014coco,zhang2019object365,zhou2017ade20k,Cityscapes,Geiger2012CVPR}. Most recent models are trained across a wide spectrum of image collections and, as a result, demonstrate promising zero-shot generalizability to unseen domains \cite{kirillov2023sam,zhang2023openSeeD,zou2023xcoder}. This versatile capability opens up new possibilities for driving perceptions from RGB cameras \cite{yan2024survey}. Leveraging these advances, we aim to transfer the semantic richness of pre-trained image models to LiDAR-based 3D scene understanding in a data-efficient manner, avoiding the need for ground truth image labels.

Specifically, given $\{ x_{\text{p}}, x_{\text{img}} \}$ pairs, we extract the point cloud and image features using the Student network's backbone and a pre-trained image segmentation backbone $\hat{\mathcal{G}}_{\xi}^{\text{img}}$, parameterized by $\xi$:
\begin{align}
\begin{split}
    F_{\text{p}} = \mathcal{H}_{\zeta}^{s}(\hat{\mathcal{G}}_{\hat{\theta}}^{s}(x_{\text{p}}))\,,~~~
    F_{\text{img}} = \hat{\mathcal{G}}_{\xi}^{\text{img}}(x_{\text{img}})\,,
\end{split}
\label{equ:extract_feature}
\end{align}
where $\hat{\mathcal{G}}_{\hat{\theta}}$ with parameters $\hat{\theta}$ is the backbone of the Student network $\mathcal{G}_{\theta}^{t}$, and $\mathcal{H}_{\zeta}^s$ is a projection layer mapping the dimension of point cloud features to match that of image features. Based on the sensor calibration parameters from Eq.~\ref{equ:projection}, we align the features to create paired point-image features $\{F_{\text{p}}, \Tilde{F}_\text{img}\}$. A key aspect of our approach is ensuring that the integration of image-based semantics does not diminish the unique spatial features of the point cloud.

To achieve this goal, we introduce a feature alignment strategy augmented with residual connections. This design allows the network to enhance point cloud features with semantic information from the image while maintaining the point cloud's inherent geometrical properties.
We then define the camera-to-LiDAR feature distillation loss $\mathcal{L}_\text{c2l}$ to minimize the cosine distance between the paired features:
\begin{align}
    \mathcal{L}_{\text{c2l}} = \frac{1}{\sum_{i=1}^{|x_{\text{p}}|} \mathcal{M}^{(i)}} \sum_{i=1}^{|x_{\text{p}}|} \left(\mathcal{M}^{(i)} \cdot \left( 1 - \langle F_{\text{p}}^{(i)}, \Tilde{F}_{\text{img}}^{(i)} \rangle \right)\right),
\label{equ:c2l}
\end{align}
where $\langle , \rangle$ calculates the inner product. The use of cosine distance ensures that the semantic alignment process enriches the point cloud features without overwhelming them, thereby achieving a balanced integration of both semantic and geometric information. This approach facilitates the transfer of semantically coherent features from images to LiDAR, enhancing the LiDAR segmentation model's performance in a data-efficient manner.

\subsection{Language-Driven Knowledge Guidance}
\label{sec:lkg}
Vision-language models, such as CLIP \cite{radford2021clip}, have demonstrated significant success in enabling open-vocabulary predictions by aligning visual and textual information. This capability is highly beneficial for driving scenarios, where a broader understanding beyond fixed label sets is required. Recent work in open-vocabulary image segmentation has further leveraged these models to train large-scale multi-task networks that can effectively align with textual inputs \cite{zou2023seem,zhang2023openSeeD,zou2023xcoder}. In our framework, we harness these capabilities to provide enriched supervision signals for 3D scene understanding using unlabeled LiDAR point clouds.

Given class names as text prompts, we use the CLIP text encoder $\mathcal{G}^{\text{txt}}_{\varrho}$ to extract text embeddings $F^{'}_{\text{txt}}$, parameterized by $\varrho$. These embeddings, combined with the image features $F_{\text{img}}$ from Eq.~\ref{equ:extract_feature}, are processed through the pre-trained open-vocabulary image segmentation head $\mathcal{H}_{\varsigma}^{\text{img}}$ and the LiDAR segmentation model $\mathcal{G}_{\theta}^{s}$:
\begin{align}
    \begin{split}
        F^{'}_{\text{p}} = \mathcal{G}_{\theta}^{s}(x_{\text{p}})\,,~~~
        F^{{\prime}}_{\text{img}} = \mathcal{H}{\varsigma}^{\text{img}}(F_{\text{img}}) \circ F^{{\prime}}_{\text{txt}}\,,
    \end{split}
\end{align}
where $\varsigma$ represents the parameters of $\mathcal{H}_{\varsigma}^{\text{img}}$. $F^{'}_{\text{p}}$ and $F^{'}_{\text{img}}$ denote non-probabilistic outputs from the models, which are then paired as $\{ F^{'}_{\text{p}}, \Tilde{F}^{'}_{\text{img}} \}$.
Symbol $\circ$ represents the alignment operation that combines the image feature vector $\mathcal{H}{\varsigma}^{\text{img}}(F_{\text{img}})$ with the text feature $F^{\prime}_{\text{txt}}$.
In our actual implementation, we use the text and image encoders from CLIP \cite{radford2021clip} to extract the text embedding and image embedding, respectively. We then use a cosine similarity loss to align the text and image features, resulting in text-aligned $F^{'}_{\text{img}}$.

To integrate semantic cues derived from text into the point cloud domain effectively, we employ a weighted feature fusion approach that respects the spatial integrity of the LiDAR data. This strategy, along with residual connections, ensures that the semantic information enhances the understanding of the scene without compromising the distinct structural characteristics of the point cloud. By maintaining this balance, our approach provides a richer and more robust understanding of the 3D environment.

To realize the language-driven knowledge guidance objective, we minimize the cosine distance between the point features and the text-aligned image features, that is:
\begin{align}
    \mathcal{L}_{\text{lkg}} = \frac{1}{\sum_{i=1}^{|x_{\text{p}}|} \mathcal{M}^{(i)}} \sum_{i=1}^{|x_{\text{p}}|} \mathcal{M}^{(i)} \cdot \left( 1 - \langle F_{\text{p}}^{\prime,(i)}, \Tilde{F}^{\prime,(i)}_{\text{img}} \rangle \right)\,.
\label{equ:lkg}
\end{align}
This loss function not only aligns the features but also ensures that the point cloud’s unique spatial features are preserved during the distillation process, promoting a harmonious integration of multi-modal information.

\begin{wrapfigure}{r}{0.58\textwidth}
    \begin{minipage}{\linewidth}
    \centering
    \includegraphics[width=\linewidth]{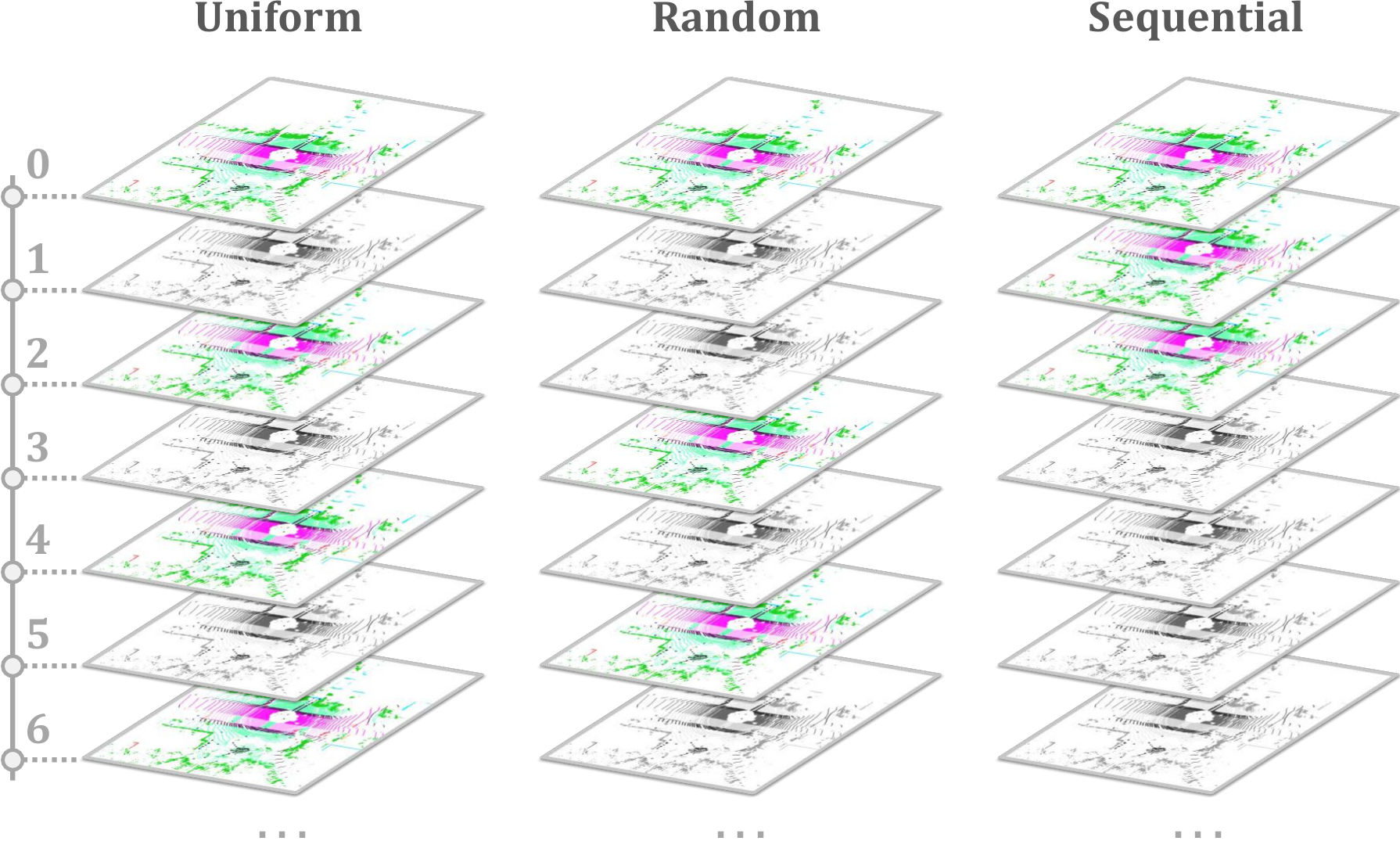}
    \vspace{-0.5cm}
    \caption{\textbf{Data splitting strategies} for data-efficient 3D scene understanding. The labeled (color) and unlabeled (gray-scale) LiDAR scans can be split via uniform (left), random (middle), and sequential (right) sampling strategies, respectively.}
    \label{fig:split}
    \end{minipage}
    \vspace{-0.2cm}
\end{wrapfigure}
\noindent\emph{Role in our framework:} 
Our enhanced framework is designed to leverage both LiDAR and image data for data-efficient 3D scene understanding, without needing image labels from the target driving datasets. We first propose the multi-modal LaserMix to enable richer interactions across sensors. To align point cloud and image features, we adapt a pretrained image segmentation model to the LiDAR segmentation backbone. To obtain auxiliary supervision signals for unlabeled scans, we propose to generate text-aligned non-probabilistic outputs from images and match them with those of the LiDAR point clouds. As we will show in the following sections, these two modules contribute to significant performance improvements for data-efficient 3D scene understanding.

\subsection{Overall Framework}
\label{sec:pipeline}
Incorporating everything together, we now present the overall LaserMix++ framework. Based on the 3D scene consistency regularization baseline in Section~\ref{sec:consistency_framework}, our approach seamlessly integrates optimization of objectives as follows:
\begin{itemize}
    \item The conventional supervision signals from labeled data, as in Eq.~\ref{equ:sup}.
    \item The mixing-based cross-sensor consistency from multi-modal LaserMix, as in Eq.~\ref{eq:mm-lasermix}.
    \item The consistency regularization between Student and Teacher networks, as in Eq.~\ref{equ:mt}.
    \item The camera-to-LiDAR feature distillation between images and LiDAR point clouds, as in Eq.~\ref{equ:c2l}.
    \item The auxiliary supervisions for unlabeled data, obtained from text-driven knowledge guidance in Eq.~\ref{equ:lkg}.
\end{itemize}
The overall objective aims to minimize the following losses:
\begin{align}
    \mathcal{L} = \mathcal{L}_{\text{sup}} + \lambda_{\text{mix}}\mathcal{L}_{\text{mix}} + \lambda_{\text{mt}}\mathcal{L}_{\text{mt}} + \lambda_{\text{c2l}}\mathcal{L}_{\text{c2l}} + \lambda_{\text{lkg}}\mathcal{L}_{\text{lkg}} \,,
\label{equ:final}
\end{align}
where $\lambda_{\text{c2l}}$ and $\lambda_{\text{lkg}}$ denote the loss weights of the camera-to-LiDAR feature distillation loss and the language-driven knowledge guidance loss, respectively.
\section{Experiments}
\label{sec:experiments}

In this section, we conduct thorough comparative and ablation experiments to validate the effectiveness and superiority of the proposed LaserMix++ framework.

\begin{table}[t]
\centering
\caption{\textbf{Benchmarking results} among state-of-the-art approaches using the LiDAR \emph{range view}, \emph{bird's eye view (BEV)}, \emph{sparse voxel}, and \emph{multi-view fusion} backbones. A unified backbone setup is used across representations, except for GPC \cite{GPC} and LiM3D \cite{li23lim3d}, which adopt extra modules. All mIoU scores are given in percentage ($\%$). The 
\textcolor{gray}{{\emph{sup.-only}}}, \textcolor{lightblue}{{\emph{best}}}, and \textcolor{brown}{{\emph{second best}}} scores under each data split within each group are shaded with \textcolor{gray}{{\emph{gray}}}, \textcolor{lightblue}{{\emph{blue}}}, and \textcolor{brown}{{\emph{yellow}}}, respectively.}
\vspace{-0.2cm}
\setlength{\tabcolsep}{4pt}
\resizebox{0.9\linewidth}{!}{
\begin{tabular}{c|r|r|c|p{20pt}<{\centering}p{24pt}<{\centering}p{20pt}<{\centering}p{24pt}<{\centering}|p{20pt}<{\centering}p{24pt}<{\centering}p{20pt}<{\centering}p{24pt}<{\centering}|p{20pt}<{\centering}p{24pt}<{\centering}p{20pt}<{\centering}p{24pt}<{\centering}}
\toprule
\multirow{2}{*}{\textbf{Repr.}} & \multirow{2}{*}{\textbf{Method}} & \multirow{2}{*}{\textbf{Venue}} & \multirow{2}{*}{\textbf{Backbone}} & 
\multicolumn{4}{c|}{\textbf{nuScenes}~\cite{Panoptic-nuScenes}} & \multicolumn{4}{c|}{\textbf{SemanticKITTI}~\cite{SemanticKITTI}} & \multicolumn{4}{c}{\textbf{ScribbleKITTI}~\cite{ScribbleKITTI}}
\\
& & & & \textbf{1\%} & \textbf{10\%} & \textbf{20\%} & \textbf{50\%} & \textbf{1\%} & \textbf{10\%} & \textbf{20\%} & \textbf{50\%} & \textbf{1\%} & \textbf{10\%} & \textbf{20\%} & \textbf{50\%}
\\\midrule\midrule
\multirow{9}{*}{\rotatebox{90}{Range View}} &
\cellcolor{gray!10}\emph{Sup.-only} & \cellcolor{gray!10}- & \cellcolor{gray!10}FIDNet & \cellcolor{gray!10}$38.3$ & \cellcolor{gray!10}$57.5$ & \cellcolor{gray!10}$62.7$ & \cellcolor{gray!10}$67.6$ & \cellcolor{gray!10}$36.2$ & \cellcolor{gray!10}$52.2$ & \cellcolor{gray!10}$55.9$ & \cellcolor{gray!10}$57.2$ & \cellcolor{gray!10}$33.1$ & \cellcolor{gray!10}$47.7$ & \cellcolor{gray!10}$49.9$ & \cellcolor{gray!10}$52.5$
\\\cmidrule{2-16}
& MeanTeacher~\cite{MeanTeacher} & NeurIPS'17 & \multirow{5}{*}{FIDNet} & $42.1$ & $60.4$ & $65.4$ & $69.4$ & $37.5$ & $53.1$ & $56.1$ & $57.4$ & $34.2$ & $49.8$ & $51.6$ & $53.3$
\\
& CBST~\cite{CBST} & ECCV'18 & & $40.9$ & $60.5$ & $64.3$ & $69.3$ & $39.9$ & $53.4$ & $56.1$ & $56.9$ & $35.7$ & $50.7$ & $52.7$ & $54.6$
\\
& CutMix-Seg~\cite{CutMix-Seg} & BMVC'20 & & $43.8$ & $63.9$ & $64.8$ & $69.8$ & $37.4$ & $54.3$ & $56.6$ & $57.6$ & $36.7$ & $50.7$ & $52.9$ & $54.3$
\\
& CPS~\cite{CPS} & CVPR'21 & & $40.7$ & $60.8$ & $64.9$ & $68.0$ & $36.5$ & $52.3$ & $56.3$ & $57.4$ & $33.7$ & $50.0$ & $52.8$ & $54.6$
\\
& LaserMix~\cite{kong2023laserMix} & CVPR'23 & & \cellcolor{yellow!12.5}$49.5$ & \cellcolor{yellow!12.5}$68.2$ & \cellcolor{yellow!12.5}$70.6$ & \cellcolor{yellow!12.5}$73.0$ & \cellcolor{yellow!12.5}$47.4$ & \cellcolor{yellow!12.5}$60.1$ & \cellcolor{yellow!12.5}$61.0$ & \cellcolor{yellow!12.5}$62.6$ & \cellcolor{yellow!12.5}$45.7$ & \cellcolor{yellow!12.5}$55.5$ & \cellcolor{yellow!12.5}$56.8$ & \cellcolor{yellow!12.5}$58.7$
\\\cmidrule{2-16}
& \textbf{LaserMix++} & \textbf{Ours} & \multirow{2}{*}{FIDNet} & \cellcolor{lightblue!9}$\mathbf{51.6}$ & \cellcolor{lightblue!9}$\mathbf{69.8}$ & \cellcolor{lightblue!9}$\mathbf{71.7}$ & \cellcolor{lightblue!9}$\mathbf{73.7}$ & \cellcolor{lightblue!9}$\mathbf{50.1}$ & \cellcolor{lightblue!9}$\mathbf{61.9}$ & \cellcolor{lightblue!9}$\mathbf{62.4}$ & \cellcolor{lightblue!9}$\mathbf{63.7}$ & \cellcolor{lightblue!9}$\mathbf{47.1}$ & \cellcolor{lightblue!9}$\mathbf{57.1}$ & \cellcolor{lightblue!9}$\mathbf{58.0}$ & \cellcolor{lightblue!9}$\mathbf{59.1}$
\\
& \emph{Improv.} $\uparrow$ & - & & \small{\textcolor{lightblue}{$\mathbf{+2.1}$}} & \small{\textcolor{lightblue}{$\mathbf{+1.6}$}} & \small{\textcolor{lightblue}{$\mathbf{+1.1}$}} & \small{\textcolor{lightblue}{$\mathbf{+0.7}$}} & \small{\textcolor{lightblue}{$\mathbf{+2.2}$}} & \small{\textcolor{lightblue}{$\mathbf{+1.8}$}} & \small{\textcolor{lightblue}{$\mathbf{+1.4}$}} & \small{\textcolor{lightblue}{$\mathbf{+1.1}$}} & \small{\textcolor{lightblue}{$\mathbf{+1.4}$}} & \small{\textcolor{lightblue}{$\mathbf{+1.6}$}} & \small{\textcolor{lightblue}{$\mathbf{+1.2}$}} & \small{\textcolor{lightblue}{$\mathbf{+0.4}$}}
\\\midrule\midrule
\multirow{7}{*}{\rotatebox{90}{BEV}} &
\cellcolor{gray!10}\emph{Sup.-only} & \cellcolor{gray!10}- & \cellcolor{gray!10}PolarNet & \cellcolor{gray!10}$50.9$ & \cellcolor{gray!10}$67.5$ & \cellcolor{gray!10}$69.5$ & \cellcolor{gray!10}$71.0$ & \cellcolor{gray!10}$45.1$ & \cellcolor{gray!10}$54.6$ & \cellcolor{gray!10}$55.6$ & \cellcolor{gray!10}$56.5$ & \cellcolor{gray!10}$42.6$ & \cellcolor{gray!10}$52.8$ & \cellcolor{gray!10}$53.4$ & \cellcolor{gray!10}$54.4$
\\\cmidrule{2-16}
& MeanTeacher~\cite{MeanTeacher} & NeurIPS'17 & \multirow{3}{*}{PolarNet} & $51.9$ & $68.1$ & $69.7$ & $71.1$ & $47.4$ & $55.6$ & $56.6$ & $57.1$ & $43.7$ & $53.4$ & $54.4$ & $54.9$
\\
& CPS~\cite{CPS} & CVPR'21 & & $52.1$ & $67.7$ & $69.8$ & $71.2$ & $46.9$ & $54.9$ & $56.0$ & $56.9$ & $44.0$ & $53.5$ & $54.4$ & $55.1$
\\
& LaserMix~\cite{kong2023laserMix} & CVPR'23 & & \cellcolor{yellow!12.5}$54.0$ & \cellcolor{yellow!12.5}$69.5$ & \cellcolor{yellow!12.5}$70.8$ & \cellcolor{yellow!12.5}$71.9$ & \cellcolor{yellow!12.5}$51.0$ & \cellcolor{yellow!12.5}$57.7$ & \cellcolor{yellow!12.5}$58.6$ & \cellcolor{yellow!12.5}$60.0$ & \cellcolor{yellow!12.5}$45.7$ & \cellcolor{yellow!12.5}$55.5$ & \cellcolor{yellow!12.5}$56.0$ & \cellcolor{yellow!12.5}$56.6$
\\\cmidrule{2-16}
& \textbf{LaserMix++} & \textbf{Ours} & \multirow{2}{*}{PolarNet} & \cellcolor{lightblue!9}$\mathbf{56.5}$ & \cellcolor{lightblue!9}$\mathbf{71.5}$ & \cellcolor{lightblue!9}$\mathbf{71.8}$ & \cellcolor{lightblue!9}$\mathbf{72.7}$ & \cellcolor{lightblue!9}$\mathbf{54.0}$ & \cellcolor{lightblue!9}$\mathbf{59.9}$ & \cellcolor{lightblue!9}$\mathbf{60.6}$ & \cellcolor{lightblue!9}$\mathbf{62.3}$ & \cellcolor{lightblue!9}$\mathbf{48.3}$ & \cellcolor{lightblue!9}$\mathbf{57.8}$ & \cellcolor{lightblue!9}$\mathbf{58.6}$ & \cellcolor{lightblue!9}$\mathbf{58.8}$
\\
& \emph{Improv.} $\uparrow$ & - & & \small{\textcolor{lightblue}{$\mathbf{+2.5}$}} & \small{\textcolor{lightblue}{$\mathbf{+2.0}$}} & \small{\textcolor{lightblue}{$\mathbf{+1.0}$}} & \small{\textcolor{lightblue}{$\mathbf{+0.8}$}} & \small{\textcolor{lightblue}{$\mathbf{+3.0}$}} & \small{\textcolor{lightblue}{$\mathbf{+2.2}$}} & \small{\textcolor{lightblue}{$\mathbf{+2.0}$}} & \small{\textcolor{lightblue}{$\mathbf{+2.3}$}} & \small{\textcolor{lightblue}{$\mathbf{+2.6}$}} & \small{\textcolor{lightblue}{$\mathbf{+2.3}$}} & \small{\textcolor{lightblue}{$\mathbf{+2.6}$}} & \small{\textcolor{lightblue}{$\mathbf{+2.2}$}}
\\\midrule\midrule

\multirow{19}{*}{\rotatebox{90}{Voxel}} &
\cellcolor{gray!10}\emph{Sup.-only} & \cellcolor{gray!10}- & \cellcolor{gray!10}Cylinder3D & \cellcolor{gray!10}$50.9$ & \cellcolor{gray!10}$65.9$ & \cellcolor{gray!10}$66.6$ & \cellcolor{gray!10}$71.2$ & \cellcolor{gray!10}$45.4$ & \cellcolor{gray!10}$56.1$ & \cellcolor{gray!10}$57.8$ & \cellcolor{gray!10}$58.7$ & \cellcolor{gray!10}$39.2$ & \cellcolor{gray!10}$48.0$ & \cellcolor{gray!10}$52.1$ & \cellcolor{gray!10}$53.8$
\\\cmidrule{2-16}
& MeanTeacher~\cite{MeanTeacher} & NeurIPS'17 & \multirow{8}{*}{Cylinder3D} & $51.6$ & $66.0$ & $67.1$ & $71.7$ & $45.4$ & $57.1$ & $59.2$ & $60.0$ & $41.0$ & $50.1$ & $52.8$ & $53.9$
\\
& CBST~\cite{CBST} & ECCV'18 & & $53.0$ & $66.5$ & $69.6$ & $71.6$ & $48.8$ & $58.3$ & $59.4$ & $59.7$ & $41.5$ & $50.6$ & $53.3$ & $54.5$
\\
& CPS~\cite{CPS} & CVPR'21 & & $52.9$ & $66.3$ & $70.0$ & $72.5$ &  $46.7$ & $58.7$ & $59.6$ & $60.5$ & $41.4$ & $51.8$ & $53.9$ & $54.8$
\\
& GPC~\cite{GPC} & ICCV'21 & & - & - & - & - & - & $49.9$ & $58.8$ & - & - & - & - & -
\\
& CRB~\cite{ScribbleKITTI} & CVPR'22 & & - & - & - & - & - & $58.7$ & $59.1$ & $60.9$ & - & $54.2$ & $56.5$ & $58.9$
\\
& LaserMix~\cite{kong2023laserMix} & CVPR'23 & & \cellcolor{yellow!12.5}$55.3$ & \cellcolor{yellow!12.5}$69.9$ & \cellcolor{yellow!12.5}$71.8$ & \cellcolor{yellow!12.5}$73.2$ & $50.6$ & $60.0$ & $61.9$ & $62.3$ & \cellcolor{yellow!12.5}$44.2$ & $53.7$ & $55.1$ & $56.8$
\\
& LiM3D~\cite{li23lim3d} & CVPR'23 & & - & - & - & - & - & \cellcolor{yellow!12.5}$61.6$ & \cellcolor{yellow!12.5}$62.6$ & \cellcolor{yellow!12.5}$62.8$ & - & \cellcolor{lightblue!9}$60.3$ & \cellcolor{lightblue!9}$60.5$ & \cellcolor{lightblue!9}$60.9$
\\
& ImageTo360 \cite{reichardt2023imageto360} & ICCVW'23 & & - & - & - & - & \cellcolor{yellow!12.5}$54.1$ & $60.0$ & $62.2$ & \cellcolor{lightblue!9}$65.0$ & - & - & - & -
\\\cmidrule{2-16}
& \textbf{LaserMix++} & \textbf{Ours} & \multirow{2}{*}{Cylinder3D} & \cellcolor{lightblue!9}$\mathbf{58.5}$ & \cellcolor{lightblue!9}$\mathbf{71.1}$ & \cellcolor{lightblue!9}$\mathbf{72.8}$ & \cellcolor{lightblue!9}$\mathbf{74.0}$ & \cellcolor{lightblue!9}$\mathbf{56.2}$ & \cellcolor{lightblue!9}$\mathbf{62.3}$ & \cellcolor{lightblue!9}$\mathbf{62.9}$ & \cellcolor{yellow!12.5}$\mathbf{63.4}$ & \cellcolor{lightblue!9}$\mathbf{47.3}$ & \cellcolor{yellow!12.5}$\mathbf{56.7}$ & \cellcolor{yellow!12.5}$\mathbf{57.6}$ & \cellcolor{yellow!12.5}$\mathbf{59.8}$
\\
& \emph{Improv.} $\uparrow$ & - & & \small{\textcolor{lightblue}{$\mathbf{+3.2}$}} & \small{\textcolor{lightblue}{$\mathbf{+1.2}$}} & \small{\textcolor{lightblue}{$\mathbf{+1.0}$}} & \small{\textcolor{lightblue}{$\mathbf{+0.8}$}} & \small{\textcolor{lightblue}{$\mathbf{+5.6}$}} & \small{\textcolor{lightblue}{$\mathbf{+0.7}$}} & \small{\textcolor{lightblue}{$\mathbf{+0.3}$}} & \small{\textcolor{lightblue}{$\mathbf{+0.6}$}} & \small{\textcolor{lightblue}{$\mathbf{+3.1}$}} & \small{\textcolor{brown}{$\mathbf{-3.6}$}} & \small{\textcolor{brown}{$\mathbf{-2.9}$}} & \small{\textcolor{brown}{$\mathbf{-1.1}$}}
\\\cmidrule{2-16}

& \cellcolor{gray!10}\emph{Sup.-only} & \cellcolor{gray!10}- & \cellcolor{gray!10}MinkUNet & \cellcolor{gray!10}$58.3$ & \cellcolor{gray!10}$71.0$ & \cellcolor{gray!10}$73.0$ & \cellcolor{gray!10}$75.1$ & \cellcolor{gray!10}$53.9$ & \cellcolor{gray!10}$64.0$ & \cellcolor{gray!10}$64.6$ & \cellcolor{gray!10}$65.4$ & \cellcolor{gray!10}$48.6$ & \cellcolor{gray!10}$57.7$ & \cellcolor{gray!10}$58.5$ & \cellcolor{gray!10}$60.0$
\\\cmidrule{2-16}
& MeanTeacher~\cite{MeanTeacher} & NeurIPS'17 & \multirow{3}{*}{MinkUNet} & $60.1$ & $71.7$ & $73.4$ & $75.2$ & $56.1$ & $64.7$ & $65.4$ & $66.0$ & $49.7$ & $59.4$ & $60.0$ & $61.7$
\\
& CPS~\cite{CPS} & CVPR'21 & & $59.8$ & $71.6$ & $73.4$ & $75.1$ & $54.7$ & $64.1$ & $65.5$ & $66.2$ & $50.1$ & $59.6$ & $60.3$ & $61.6$
\\
& LaserMix~\cite{kong2023laserMix} & CVPR'23 & & \cellcolor{yellow!12.5}$62.8$ & \cellcolor{yellow!12.5}$73.6$ & \cellcolor{yellow!12.5}$74.8$ & \cellcolor{yellow!12.5}$76.1$ & \cellcolor{yellow!12.5}$60.9$ & \cellcolor{yellow!12.5}$66.6$ & \cellcolor{yellow!12.5}$67.2$ & \cellcolor{yellow!12.5}$68.0$ & \cellcolor{yellow!12.5}$57.2$ & \cellcolor{yellow!12.5}$61.1$ & \cellcolor{yellow!12.5}$61.4$ & \cellcolor{yellow!12.5}$62.4$ 
\\\cmidrule{2-16}
& \textbf{LaserMix++} & \textbf{Ours} & \multirow{2}{*}{MinkUNet} & \cellcolor{lightblue!9}$\mathbf{64.7}$ & \cellcolor{lightblue!9}$\mathbf{74.6}$ & \cellcolor{lightblue!9}$\mathbf{75.6}$ & \cellcolor{lightblue!9}$\mathbf{76.6}$ & \cellcolor{lightblue!9}$\mathbf{63.1}$ & \cellcolor{lightblue!9}$\mathbf{67.9}$ & \cellcolor{lightblue!9}$\mathbf{68.2}$ & \cellcolor{lightblue!9}$\mathbf{68.7}$ & \cellcolor{lightblue!9}$\mathbf{61.0}$ & \cellcolor{lightblue!9}$\mathbf{64.2}$ & \cellcolor{lightblue!9}$\mathbf{64.8}$ & \cellcolor{lightblue!9}$65.1$
\\
& \emph{Improv.} $\uparrow$ & - & & \small{\textcolor{lightblue}{$\mathbf{+1.9}$}} & \small{\textcolor{lightblue}{$\mathbf{+1.0}$}} & \small{\textcolor{lightblue}{$\mathbf{+1.2}$}} & \small{\textcolor{lightblue}{$\mathbf{+0.5}$}} & \small{\textcolor{lightblue}{$\mathbf{+2.2}$}} & \small{\textcolor{lightblue}{$\mathbf{+1.3}$}} & \small{\textcolor{lightblue}{$\mathbf{+1.0}$}} & \small{\textcolor{lightblue}{$\mathbf{+0.7}$}} & \small{\textcolor{lightblue}{$\mathbf{+3.8}$}} & \small{\textcolor{lightblue}{$\mathbf{+3.1}$}} & \small{\textcolor{lightblue}{$\mathbf{+3.4}$}} & \small{\textcolor{lightblue}{$\mathbf{+2.7}$}}
\\\midrule\midrule

\multirow{8}{*}{\rotatebox{90}{Fusion}} &
\cellcolor{gray!10}\emph{Sup.-only} & \cellcolor{gray!10}- & \cellcolor{gray!10}SPVCNN & \cellcolor{gray!10}$57.9$ & \cellcolor{gray!10}$71.7$ & \cellcolor{gray!10}$73.0$ & \cellcolor{gray!10}$74.6$ & \cellcolor{gray!10}$52.7$ & \cellcolor{gray!10}$64.1$ & \cellcolor{gray!10}$64.5$ & \cellcolor{gray!10}$65.1$ & \cellcolor{gray!10}$47.2$ & \cellcolor{gray!10}$57.3$ & \cellcolor{gray!10}$58.2$ & \cellcolor{gray!10}$58.8$
\\\cmidrule{2-16}
& MeanTeacher~\cite{MeanTeacher} & NeurIPS'17 & \multirow{4}{*}{SPVCNN} & $59.4$ & $72.5$ & $73.1$ & $74.7$ & $54.4$ & $64.8$ & $65.2$ & $65.7$ & $49.9$ & $58.3$ & $58.6$ & $59.1$
\\
& CPS~\cite{CPS} & CVPR'21 & & $58.7$ & $72.0$ & $73.2$ & $74.7$ & $54.6$ & $64.6$ & $65.3$ & $65.9$ & $48.7$ & $58.0$ & $58.4$ & $59.0$
\\
& LaserMix~\cite{kong2023laserMix} & CVPR'23 & & \cellcolor{yellow!12.5}$63.2$ & \cellcolor{yellow!12.5}$74.1$ & \cellcolor{yellow!12.5}$74.6$ & \cellcolor{yellow!12.5}$75.8$ & \cellcolor{yellow!12.5}$60.3$ & \cellcolor{yellow!12.5}$66.6$ & \cellcolor{yellow!12.5}$67.0$ & \cellcolor{yellow!12.5}$67.6$ & \cellcolor{yellow!12.5}$57.1$ & \cellcolor{yellow!12.5}$60.8$ & \cellcolor{yellow!12.5}$60.7$ & \cellcolor{yellow!12.5}$61.0$
\\
& ImageTo360 \cite{reichardt2023imageto360} & ICCVW'23 & & - & - & - & - & $59.5$ & $62.4$ & $64.2$ & $66.1$ & - & - & - & -
\\\cmidrule{2-16}
& \textbf{LaserMix++} & \textbf{Ours} & \multirow{2}{*}{SPVCNN} & \cellcolor{lightblue!9}$\mathbf{65.3}$ & \cellcolor{lightblue!9}$\mathbf{75.3}$ & \cellcolor{lightblue!9}$\mathbf{75.2}$ & \cellcolor{lightblue!9}$\mathbf{76.3}$ & \cellcolor{lightblue!9}$\mathbf{63.2}$ & \cellcolor{lightblue!9}$\mathbf{67.5}$ & \cellcolor{lightblue!9}$\mathbf{67.7}$ & \cellcolor{lightblue!9}$\mathbf{68.6}$ & \cellcolor{lightblue!9}$\mathbf{60.6}$ & \cellcolor{lightblue!9}$\mathbf{63.6}$ & \cellcolor{lightblue!9}$\mathbf{65.0}$ & \cellcolor{lightblue!9}$\mathbf{66.2}$
\\
& \emph{Improv.} $\uparrow$ & - & & \small{\textcolor{lightblue}{$\mathbf{+3.1}$}} & \small{\textcolor{lightblue}{$\mathbf{+1.2}$}} & \small{\textcolor{lightblue}{$\mathbf{+0.6}$}} & \small{\textcolor{lightblue}{$\mathbf{+0.5}$}} & \small{\textcolor{lightblue}{$\mathbf{+2.9}$}} & \small{\textcolor{lightblue}{$\mathbf{+0.9}$}} & \small{\textcolor{lightblue}{$\mathbf{+0.7}$}} & \small{\textcolor{lightblue}{$\mathbf{+1.0}$}} & \small{\textcolor{lightblue}{$\mathbf{+3.5}$}} & \small{\textcolor{lightblue}{$\mathbf{+2.8}$}} & \small{\textcolor{lightblue}{$\mathbf{+4.3}$}} & \small{\textcolor{lightblue}{$\mathbf{+5.2}$}} 
\\
\bottomrule
\end{tabular}}
\label{tab:benchmark}
\end{table}
\begin{table*}[t]
\centering
\caption{\textbf{Robustness enhancement effect analysis} among different mixing-based 3D scene augmentation methods on the Robo3D \cite{kong2023robo3D} benchmark for LiDAR semantic segmentation (\emph{w/} a MinkUNet \cite{choy2019minkowski} backbone) and 3D object detection (\emph{w/} a CenterPoint \cite{centerpoint} backbone) tasks. All mIoU/mAP/mCE/mRR scores are given in percentage ($\%$). The 
\textcolor{gray}{{\emph{baseline}}}, \textcolor{lightblue}{{\emph{best}}}, and \textcolor{brown}{{\emph{second best}}} scores under each evaluation metric are shaded with \textcolor{gray}{{\emph{gray}}}, \textcolor{lightblue}{{\emph{blue}}}, and \textcolor{brown}{{\emph{yellow}}}, respectively.}
\vspace{-0.2cm}
\resizebox{\linewidth}{!}{
\begin{tabular}{c|r|r|c|p{30pt}<{\centering}p{30pt}<{\centering}|p{26.5pt}<{\centering}p{26.5pt}<{\centering}p{26.5pt}<{\centering}p{26.5pt}<{\centering}p{26.5pt}<{\centering}p{26.5pt}<{\centering}p{26.5pt}<{\centering}p{26.5pt}<{\centering}}
\toprule
\textbf{Set} & \textbf{Method} & \textbf{Venue} & \textbf{Backbone} & \textbf{mCE}~\small{$\downarrow$} & \textbf{mRR}~\small{$\uparrow$} & \textbf{Fog} & \textbf{Rain} & \textbf{Snow} & \textbf{Motio} & \textbf{Beam} & \textbf{Cross} & \textbf{Echo} & \textbf{Sensor}
\\\midrule\midrule
\multirow{8}{*}{\rotatebox{90}{SemanticKITTI-C}} & \cellcolor{gray!10}None & \cellcolor{gray!10}- & \cellcolor{gray!10}MinkUNet & \cellcolor{gray!10}$100.0$ & \cellcolor{gray!10}$81.9$ & \cellcolor{gray!10}$55.9$ & \cellcolor{gray!10}$54.0$ & \cellcolor{gray!10}$53.3$ & \cellcolor{gray!10}$32.9$ & \cellcolor{gray!10}$56.3$ & \cellcolor{gray!10}$58.3$ & \cellcolor{gray!10}$54.4$ & \cellcolor{gray!10}$46.1$ 
\\\cmidrule{2-14}
& Common & - & \multirow{4}{*}{MinkUNet} & $110.6$ & $83.4$ & $38.9$ & $57.1$ & $52.0$ & $41.2$ & $49.4$ & $55.0$ & $51.7$ & $41.3$
\\
& Mix3D~\cite{Mix3D} & 3DV'21 & & $96.7$ & \cellcolor{yellow!12.5}$88.0$ & $57.3$ & $54.4$ & $56.3$ & $42.9$ & $55.8$ & \cellcolor{yellow!12.5}$59.0$ & $52.9$ & $48.1$
\\
& PolarMix~\cite{xiao2022polarmix} & NeurIPS'22 & & $86.8$ & $86.9$ & $61.0$ & $63.8$ & $61.0$ & \cellcolor{yellow!12.5}$50.0$ & $61.5$ & $58.9$ & $55.9$ & $53.6$
\\
& LaserMix~\cite{kong2023laserMix} & CVPR'23 & & \cellcolor{yellow!12.5}$83.3$ & $87.0$ & \cellcolor{yellow!12.5}$62.7$ & \cellcolor{yellow!12.5}$66.3$ & \cellcolor{yellow!12.5}$62.1$ & $48.9$ & \cellcolor{yellow!12.5}$66.3$ & $56.9$ & \cellcolor{yellow!12.5}$58.0$ & \cellcolor{yellow!12.5}$57.8$
\\\cmidrule{2-14}
& \textbf{LaserMix++} & \textbf{Ours} & \multirow{2}{*}{MinkUNet} & \cellcolor{lightblue!9}$\mathbf{80.9}$ & \cellcolor{lightblue!9}$\mathbf{88.1}$ & \cellcolor{lightblue!9}$\mathbf{63.4}$ & \cellcolor{lightblue!9}$\mathbf{66.4}$ & \cellcolor{lightblue!9}$\mathbf{63.2}$ & \cellcolor{lightblue!9}$\mathbf{52.8}$ & \cellcolor{lightblue!9}$\mathbf{66.5}$ & \cellcolor{lightblue!9}$\mathbf{59.5}$ & \cellcolor{lightblue!9}$\mathbf{58.7}$ & \cellcolor{lightblue!9}$\mathbf{58.3}$
\\
& \emph{Improv.} $\uparrow$ & - & & \textcolor{lightblue}{\small{$\mathbf{-2.4}$}} & \textcolor{lightblue}{\small{$\mathbf{+0.1}$}} & \textcolor{lightblue}{\small{$\mathbf{+0.7}$}} & \textcolor{lightblue}{\small{$\mathbf{+0.1}$}} & \textcolor{lightblue}{\small{$\mathbf{+1.1}$}} & \textcolor{lightblue}{\small{$\mathbf{+2.8}$}} & \textcolor{lightblue}{\small{$\mathbf{+0.2}$}} & \textcolor{lightblue}{\small{$\mathbf{+0.5}$}} & \textcolor{lightblue}{\small{$\mathbf{+0.7}$}} & \textcolor{lightblue}{\small{$\mathbf{+0.5}$}}
\\\midrule\midrule

\multirow{8}{*}{\rotatebox{90}{WOD-C}} & \cellcolor{gray!10}None & \cellcolor{gray!10}- & \cellcolor{gray!10}CenterPoint & \cellcolor{gray!10}$100.0$ & \cellcolor{gray!10}$83.3$ & \cellcolor{gray!10}$43.1$ & \cellcolor{gray!10}$62.8$ & \cellcolor{gray!10}$58.6$ & \cellcolor{gray!10}$43.5$ & \cellcolor{gray!10}$54.4$ & \cellcolor{gray!10}$60.3$ & \cellcolor{gray!10}$57.0$ & \cellcolor{gray!10}$44.0$
\\\cmidrule{2-14}
& Common & - & \multirow{4}{*}{CenterPoint} & $110.6$ & $83.4$ & $38.9$ & $57.1$ & $52.0$ & $41.2$ & $49.4$ & $55.0$ & $51.7$ & $41.3$
\\
& GT Sampling & - & & $115.5$ & $80.2$ & $38.5$ & $55.6$ & $51.2$ & $40.0$ & $45.8$ & $51.1$ & $49.7$ & $36.6$
\\
& PolarMix~\cite{xiao2022polarmix} & NeurIPS'22 & & $101.0$ & \cellcolor{yellow!12.5}$83.7$ & $42.7$ & $62.7$ & $58.7$ & $43.2$ & $53.5$ & \cellcolor{yellow!12.5}$59.2$ & $56.4$ & \cellcolor{yellow!12.5}$43.8$
\\
& LaserMix~\cite{kong2023laserMix} & CVPR'23 & & \cellcolor{yellow!12.5}$100.5$ & $82.9$ & \cellcolor{yellow!12.5}$43.6$ & \cellcolor{yellow!12.5}$63.2$ & \cellcolor{yellow!12.5}$59.2$ & \cellcolor{yellow!12.5}$43.4$ & \cellcolor{yellow!12.5}$53.8$ & $58.9$ & \cellcolor{yellow!12.5}$56.6$ & $43.5$
\\\cmidrule{2-14}
& \textbf{LaserMix++} & \textbf{Ours} & \multirow{2}{*}{CenterPoint} & \cellcolor{lightblue!9}$\mathbf{98.2}$ & \cellcolor{lightblue!9}$\mathbf{84.3}$ & \cellcolor{lightblue!9}$\mathbf{45.1}$ & \cellcolor{lightblue!9}$\mathbf{63.6}$ & \cellcolor{lightblue!9}$\mathbf{60.1}$ & \cellcolor{lightblue!9}$\mathbf{45.2}$ & \cellcolor{lightblue!9}$\mathbf{54.3}$ & \cellcolor{lightblue!9}$\mathbf{59.8}$ & \cellcolor{lightblue!9}$\mathbf{57.8}$ & \cellcolor{lightblue!9}$\mathbf{45.1}$
\\
& \emph{Improv.} $\uparrow$ & - &  & \textcolor{lightblue}{\small{$\mathbf{-2.3}$}} & \textcolor{lightblue}{\small{$\mathbf{+0.6}$}} & \textcolor{lightblue}{\small{$\mathbf{+1.5}$}} & \textcolor{lightblue}{\small{$\mathbf{+0.4}$}} & \textcolor{lightblue}{\small{$\mathbf{+0.9}$}} & \textcolor{lightblue}{\small{$\mathbf{+1.8}$}} & \textcolor{lightblue}{\small{$\mathbf{+0.5}$}} & \textcolor{lightblue}{\small{$\mathbf{+0.6}$}} & \textcolor{lightblue}{\small{$\mathbf{+1.2}$}} & \textcolor{lightblue}{\small{$\mathbf{+1.3}$}}
\\\bottomrule
\end{tabular}}
\label{tab:robustness}
\end{table*}

\subsection{Datasets}
We utilize three data-efficient 3D scene understanding benchmarks for our experiments: nuScenes \cite{Panoptic-nuScenes}, SemanticKITTI \cite{SemanticKITTI}, and ScribbleKITTI \cite{ScribbleKITTI}. nuScenes \cite{Panoptic-nuScenes} and SemanticKITTI \cite{SemanticKITTI} are the two most popular multi-modal driving perception datasets, with 29,130 and 19,130 training scans and 6,019 and 4,071 validation scans, respectively. ScribbleKITTI \cite{ScribbleKITTI}, a derivative of SemanticKITTI \cite{SemanticKITTI}, features the same number of scans but is annotated with line scribbles, as opposed to full annotations. We employ a range of labeled training scans -- 1\%, 10\%, 20\%, and 50\% -- treating the rest as unlabeled to align with standard semi-supervised settings from the image segmentation community. Additionally, we extend our evaluations to the Cityscapes dataset \cite{Cityscapes}, following the semi-supervised splits commonly used in previous studies \cite{CPS,CCT,GCT} -- 1/16, 1/8, 1/4, and 1/2 data splits -- to assess the generality of our methods across both LiDAR and image modalities. Our approaches can also be applied to robust 3D perception. We verify this on the SemanticKITTI-C and WOD-C benchmarks from Robo3D \cite{kong2023robo3D}, which are constructed based on the SemanticKITTI \cite{SemanticKITTI} and Waymo Open \cite{sun2020waymoOpen} datasets, respectively.

\subsection{Experimental Setups}
\noindent\textbf{3D Backbone Configurations.}
To validate that LaserMix++ is universally applicable to different LiDAR representations, we conduct experiments using a total of five backbones, including FIDNet \cite{FIDNet} (range view), PolarNet \cite{PolarNet}, MinkUNet \cite{choy2019minkowski} and Cylinder3D \cite{Cylinder3D} (sparse voxel), and SPVCNN \cite{SPVNAS} (multi-view fusion). The input resolution of range images is set to 32$\times$1920 for nuScenes \cite{Panoptic-nuScenes} and 64$\times$2048 for SemanticKITTI \cite{SemanticKITTI} and ScribbleKITTI \cite{ScribbleKITTI}. The grid cell size of bird's eye view methods is set to [480, 360, 32]. The voxel size of voxel-based methods is fixed as [240, 180, 20] for all three datasets. The same voxel size is applied to the voxel branch in the multi-view fusion backbone.

\noindent\textbf{Implementation Details}. Our LaserMix++ framework is established based on the MMDetection3D codebase \cite{mmdet3d}. We denote the supervised-only baseline as \emph{sup.-only} in our experiments. For semi-supervised learning, the labeled/unlabeled data are selected in four ways as shown in Figure~\ref{fig:split}: 1) random sampling, 2) uniform sampling, 3) sequential sampling, and 4) ST-RFD \cite{li23lim3d} sampling strategies. Due to the lack of previous works, we also compare consistency regularization \cite{MeanTeacher,CPS,CutMix-Seg} and entropy minimization \cite{CBST} methods from semi-supervised image segmentation. The number of spatial area sets $m$ in multi-modal LaserMix is uniformly sampled from 2 to 6. The loss weights $\lambda_{\text{mix}}$, $\lambda_{\text{mt}}$, $\lambda_{\text{c2l}}$, and $\lambda_{\text{lkg}}$ in Eq.~\ref{equ:final} are set to 2.0, 250, 1.5, and 1.0, respectively. All experiments are implemented using PyTorch on eight NVIDIA A100 GPUs. All baseline models are trained with a batch size of 2 on each GPU, along with the AdamW optimizer \cite{AdamW}, OneCycle learning rate schedule \cite{OneCycle}, and a learning rate of 0.008. We do not include any type of test-time augmentation or model ensemble during the evaluation. For additional details, kindly refer to the Appendix.

\noindent\textbf{Evaluation Protocol.}
We follow the Realistic Evaluation Protocol \cite{Realistic-Evaluation} when building our benchmarks. Under each setting, the configurations are unified to ensure a fair comparison among different semi-supervised learning algorithms.

\noindent\textbf{Evaluation Metrics.}
In this work, we report the intersection-over-union (IoU) for each semantic class and the mean IoU (mIoU) scores across all semantic classes in semi-supervised learning benchmarks. For out-of-distribution robustness assessment, we follow the Robo3D \cite{kong2023robo3D} protocol to report the mean corruption error (mCE) and mean resilience rate (mRR).

\subsection{Comparative Study}
\noindent\textbf{Improvements over Baselines.}
In Table~\ref{tab:benchmark}, we compare the proposed LaserMix++ with the LaserMix \cite{kong2023laserMix} and \emph{sup.-only} baselines on nuScenes \cite{Panoptic-nuScenes}, SemanticKITTI \cite{SemanticKITTI}, and ScribbleKITTI \cite{ScribbleKITTI}. Since our approaches can be universally applied to different LiDAR representations, we also set up the benchmark using various backbones from the literature. We observe consistent improvements achieved across all different settings by integrating multi-modal learning for data-efficient 3D scene understanding. On average, there is a 2\% to 3\% performance gain brought by LaserMix++ over the previous framework. The improvements are especially predominant when the number of labeled data is extremely limited (\emph{e.g.}, 1\%), which verifies the effectiveness of our approaches. As for the \emph{sup.-only} baselines, the gains vary under different backbones. LaserMix++ yields up to 13\% mIoU improvements across three datasets when using the range view backbone \cite{FIDNet}. Similar trends hold for the bird's eye view \cite{PolarNet}, sparse voxel \cite{Cylinder3D,choy2019minkowski}, and multi-view fusion \cite{SPVNAS} backbones, where the mIoU gains under the 1\% split are around 6\%, 8\%, and 10\%, respectively. The results strongly verify the effectiveness of our framework and further highlight the importance of leveraging unlabeled data for LiDAR semantic segmentation.

\begin{wraptable}{r}{0.5\textwidth}
\centering
\vspace{-0.5cm}
    \caption{\textbf{Benchmarking results} among existing 3D representation learning methods pre-trained, linear-probed (LP), and fine-tuned on nuScenes \cite{Panoptic-nuScenes}. All mIoU scores are given in percentage ($\%$). The \textcolor{gray}{{\emph{sup.-only}}}, \textcolor{lightblue}{{\emph{best}}}, and \textcolor{brown}{{\emph{second best}}} scores are shaded with \textcolor{gray}{{\emph{gray}}}, \textcolor{lightblue}{{\emph{blue}}}, and \textcolor{brown}{{\emph{yellow}}}, respectively.}
\vspace{-0.2cm}
\resizebox{\linewidth}{!}{
\begin{tabular}{r|p{20.5pt}<{\centering}p{20.5pt}<{\centering}p{20.5pt}<{\centering}p{20.5pt}<{\centering}p{20.5pt}<{\centering}p{20.5pt}<{\centering}}
\toprule
\multirow{2}{*}{\textbf{Method}} & \multicolumn{6}{c}{\textbf{nuScenes}~\cite{Panoptic-nuScenes}}
\\
& \textbf{LP} & \textbf{1\%} & \textbf{5\%} & \textbf{10\%} & \textbf{20\%} & \textbf{Full}
\\\midrule\midrule

\cellcolor{gray!10}\emph{Sup.-only} & \cellcolor{gray!10}$81.0$ & \cellcolor{gray!10}$30.3$ & \cellcolor{gray!10}$47.8$ & \cellcolor{gray!10}$56.2$ & \cellcolor{gray!10}$65.5$ & \cellcolor{gray!10}$74.7$
\\\midrule
PointContrast~\cite{xie2020pointContrast} & $21.9$ & $32.5$ & - & - & - & - 
\\
DepthContrast~\cite{zhang2021depthContrast} & $22.1$ & $31.7$ & - & - & - & -
\\
PPKT~\cite{liu2021ppkt} & $35.9$ & $37.8$ & $53.7$ & $60.3$ & $67.1$ & $74.5$
\\
SLidR~\cite{sautier2022slidr} & $38.8$ & $38.3$ & $52.5$ & $59.8$ & $66.9$ & $74.8$
\\
ST-SLidR~\cite{mahmoud2023st} & $40.5$ & $40.8$ & $54.7$ & $60.8$ & $67.7$ & $75.1$
\\
Seal~\cite{liu2023segment} & $45.0$ & $45.8$ & $55.6$ & $63.0$ & $68.4$ & $75.6$
\\\midrule
\emph{w/} LaserMix~\cite{kong2023laserMix} & \cellcolor{yellow!12.5}- & \cellcolor{yellow!12.5}$48.4$ & \cellcolor{yellow!12.5}$57.8$ & \cellcolor{yellow!12.5}$65.5$ & \cellcolor{yellow!12.5}$70.8$ & \cellcolor{yellow!12.5}$77.1$
\\
\emph{Improv.} $\uparrow$ & - & \textcolor{brown}{\small{$\mathbf{+2.6}$}} & \textcolor{brown}{\small{$\mathbf{+2.2}$}} & \textcolor{brown}{\small{$\mathbf{+2.5}$}} & \textcolor{brown}{\small{$\mathbf{+2.4}$}} & \textcolor{brown}{\small{$\mathbf{+1.5}$}}
\\\midrule
\textbf{\emph{w/} LaserMix++ (Ours)} & \cellcolor{lightblue!9}- & \cellcolor{lightblue!9}$\mathbf{49.9}$ & \cellcolor{lightblue!9}$\mathbf{58.5}$ & \cellcolor{lightblue!9}$\mathbf{66.7}$ & \cellcolor{lightblue!9}$\mathbf{71.6}$ & \cellcolor{lightblue!9}$\mathbf{77.9}$ 
\\
\emph{Improv.} $\uparrow$ & - & \textcolor{lightblue}{\small{$\mathbf{+1.5}$}} & \textcolor{lightblue}{\small{$\mathbf{+0.7}$}} & \textcolor{lightblue}{\small{$\mathbf{+1.2}$}} & \textcolor{lightblue}{\small{$\mathbf{+0.8}$}} & \textcolor{lightblue}{\small{$\mathbf{+0.8}$}}
\\\bottomrule
\end{tabular}}
\label{tab:representation}
\vspace{-0.1cm}
\end{wraptable}
\noindent\textbf{Compare with State of the Arts.}
We compare LaserMix++ with recent arts from the literature, \emph{i.e.}, GPC \cite{GPC}, CRB \cite{ScribbleKITTI}, and Lim3D \cite{li23lim3d}, and show results (using the Cylinder3D backbone) in Table~\ref{tab:benchmark}. As can be seen, LaserMix++ exhibits much better results than GPC \cite{GPC} and \cite{ScribbleKITTI} across various scenarios. The most recent work, Lim3D \cite{li23lim3d}, achieves the best performance on ScribbleKITTI \cite{ScribbleKITTI}. However, its results were obtained by voting from multiple augmented test-time ensembles, which created a different evaluation protocol from ours. In addition to the above, we also reproduced several popular algorithms \cite{MeanTeacher,CBST,CPS} from the semi-supervised image segmentation domain. The results in Table~\ref{tab:benchmark} verify that these methods, albeit competitive in 2D tasks, only yield sub-par performance in the semi-supervised LiDAR semantic segmentation benchmark. This highlights the importance of exploiting the LiDAR data structure for data-efficient 3D scene understanding. 

\begin{figure}[t]
    \begin{center}
    \includegraphics[width=0.98\textwidth]{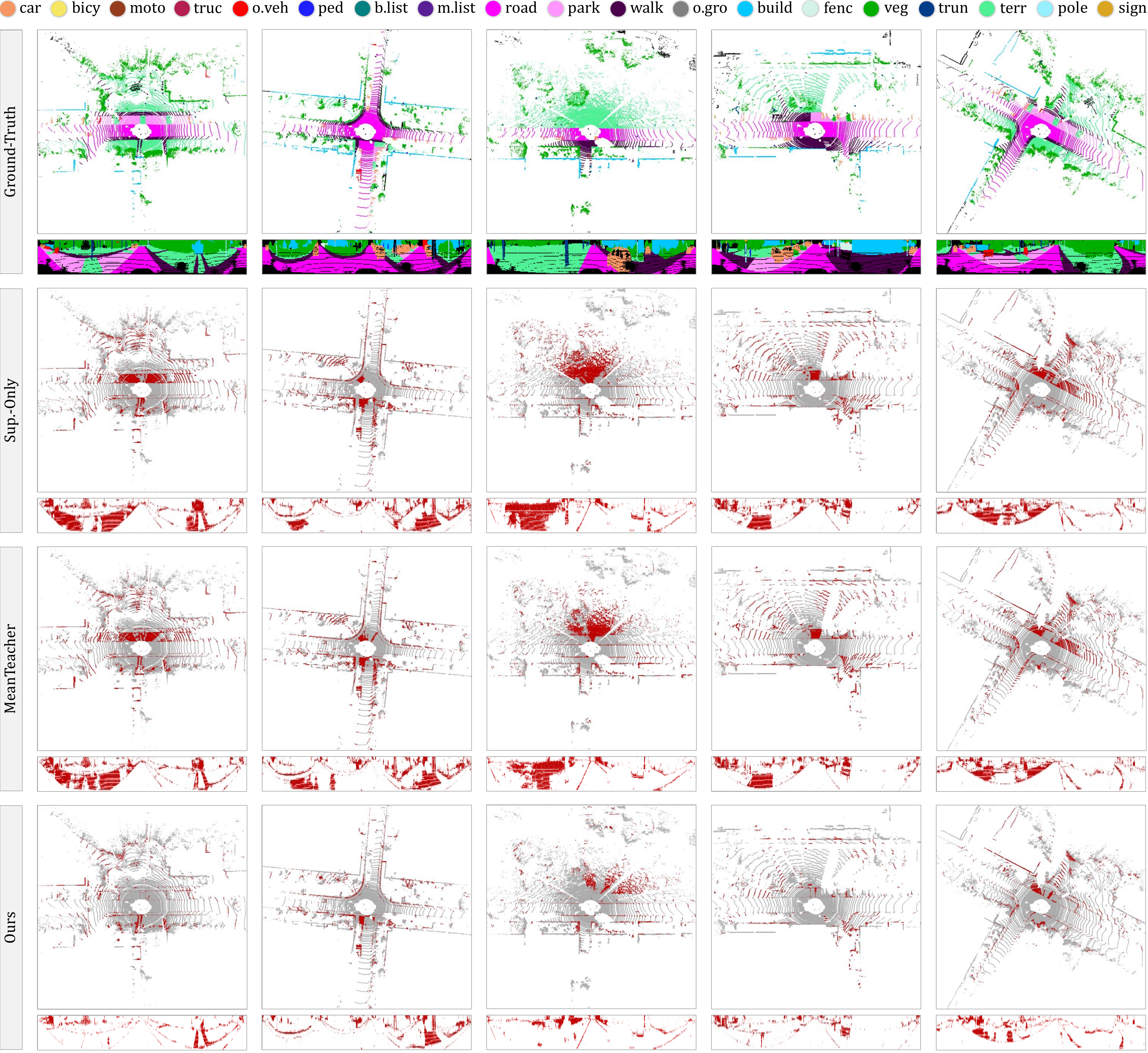}
    \end{center}
    \vspace{-0.2cm}
    \caption{\textbf{Qualitative assessments} of state-of-the-art data-efficient 3D scene understanding models from the LiDAR \emph{bird's eye view} and \emph{range view} on the validation set of SemanticKITTI~\cite{SemanticKITTI}. To highlight the differences, the \textcolor{correct}{\textbf{correct}} and \textcolor{incorrect}{\textbf{incorrect}} predictions are painted in \textcolor{correct}{\textbf{gray}} and \textcolor{red}{\textbf{red}}. Best viewed in colors and zoomed-in for additional details.}
    \label{fig:qualitative}
\end{figure}

\begin{table*}[t]
\caption{
    \textbf{Ablation study} for different components in the \textbf{LaserMix++} framework (\emph{w/} a FIDNet \cite{FIDNet} backbone) on the official \emph{val} sets of SemanticKITTI \cite{SemanticKITTI} and nuScenes \cite{Panoptic-nuScenes}. Settings: \textbf{(a)} The \textcolor{gray}{{\emph{sup.-only}}} results. \textbf{(b)} The \emph{baseline} \cite{MeanTeacher} results; \textbf{(c)} The \emph{baseline} \cite{MeanTeacher} results with multi-modal operations (camera-to-LiDAR feature distillation and language-driven knowledge guidance). \textbf{(d)} The LaserMix results \emph{w/} Student net supervision (SS); \textbf{(e)} The LaserMix results \emph{w/} Teacher net supervision (TS). \textbf{(f)} The LaserMix++ results \emph{w/} the camera-to-LiDAR feature distillation ($\mathcal{L}_{\text{c2l}}$). \textbf{(g)} The LaserMix++ results \emph{w/} the language-driven knowledge guidance ($\mathcal{L}_{\text{lkg}}$). \textbf{(h)} The complete LaserMix++ configuration results. All mIoU scores are given in percentage ($\%$).}
\vspace{-0.2cm}
\resizebox{\linewidth}{!}{
\begin{tabular}{c|cc|cc|cc|cccc|cccc}
\toprule
\multirow{2}{*}{\textbf{\#}} & \multirow{2}{*}{$\mathcal{L}_{\text{mt}}$} & \multirow{2}{*}{$\mathcal{L}_{\text{mix}}$} & \multirow{2}{*}{SS} & \multirow{2}{*}{TS} & \multirow{2}{*}{$\mathcal{L}_{\text{c2l}}$} & \multirow{2}{*}{$\mathcal{L}_{\text{lkg}}$} & \multicolumn{4}{c|}{\textbf{SemanticKITTI}~\cite{SemanticKITTI}} & \multicolumn{4}{c}{\textbf{nuScenes}~\cite{Panoptic-nuScenes}}
\\
& & & & & & & \textbf{1\%} & \textbf{10\%} & \textbf{20\%} & \textbf{50\%} & \textbf{1\%} & \textbf{10\%} & \textbf{20\%} & \textbf{50\%}
\\\midrule\midrule
a & \cellcolor{gray!10}\textcolor{gray}{\xmark} & \cellcolor{gray!10}\textcolor{gray}{\xmark} & \cellcolor{gray!10}\textcolor{gray}{\xmark} & \cellcolor{gray!10}\textcolor{gray}{\xmark} & \cellcolor{gray!10}\textcolor{gray}{\xmark} & \cellcolor{gray!10}\textcolor{gray}{\xmark} & 
\cellcolor{gray!10}$36.2$ \textcolor{gray}{\small{$\mathbf{-1.3}$}} & \cellcolor{gray!10}$52.2$ \textcolor{gray}{\small{$\mathbf{-0.9}$}} & \cellcolor{gray!10}$55.9$ \textcolor{gray}{\small{$\mathbf{-0.2}$}} & \cellcolor{gray!10}$57.2$ \textcolor{gray}{\small{$\mathbf{-0.2}$}} &
\cellcolor{gray!10}$38.3$ \textcolor{gray}{\small{$\mathbf{-3.8}$}} & \cellcolor{gray!10}$57.5$ \textcolor{gray}{\small{$\mathbf{-2.9}$}} & \cellcolor{gray!10}$62.7$ \textcolor{gray}{\small{$\mathbf{-2.7}$}} & \cellcolor{gray!10}$67.6$ \textcolor{gray}{\small{$\mathbf{-1.8}$}}
\\\midrule\midrule

b & \textcolor{brown}{\cmark} & \textcolor{gray}{\xmark} & \textcolor{gray}{\xmark} & \textcolor{gray}{\xmark} & \textcolor{gray}{\xmark} & \textcolor{gray}{\xmark} &
$37.5$ \textcolor{brown}{\small{$\mathbf{+0.0}$}} & $53.1$ \textcolor{brown}{\small{$\mathbf{+0.0}$}} & $56.1$ \textcolor{brown}{\small{$\mathbf{+0.0}$}} & $57.4$ \textcolor{brown}{\small{$\mathbf{+0.0}$}} & 
$42.1$ \textcolor{brown}{\small{$\mathbf{+0.0}$}} & $60.4$ \textcolor{brown}{\small{$\mathbf{+0.0}$}} & $65.4$ \textcolor{brown}{\small{$\mathbf{+0.0}$}} & $69.4$ \textcolor{brown}{\small{$\mathbf{+0.0}$}}
\\\midrule

    \multirow{2}{*}{c} & \textcolor{brown}{\cmark} & \textcolor{gray}{\xmark} & \textcolor{gray}{\xmark} & \textcolor{gray}{\xmark} & \textcolor{brown}{\cmark} & \textcolor{gray}{\xmark} & 
    $40.0$ \textcolor{brown}{\small{$\mathbf{+2.5}$}} & 
    $54.7$ \textcolor{brown}{\small{$\mathbf{+1.6}$}} &
    $56.6$ \textcolor{brown}{\small{$\mathbf{+0.5}$}} & 
    $57.9$ \textcolor{brown}{\small{$\mathbf{+0.5}$}} & 
    $43.9$ \textcolor{brown}{\small{$\mathbf{+1.8}$}} & 
    $61.6$ \textcolor{brown}{\small{$\mathbf{+1.2}$}} & 
    $66.2$ \textcolor{brown}{\small{$\mathbf{+0.8}$}} & 
    $69.9$ \textcolor{brown}{\small{$\mathbf{+0.5}$}}
    \\
    & \textcolor{brown}{\cmark} & \textcolor{gray}{\xmark} & \textcolor{gray}{\xmark} & \textcolor{gray}{\xmark} & \textcolor{gray}{\xmark} & \textcolor{brown}{\cmark} &
    $41.2$ \textcolor{brown}{\small{$\mathbf{+3.7}$}} & 
    $55.3$ \textcolor{brown}{\small{$\mathbf{+2.2}$}} & 
    $56.9$ \textcolor{brown}{\small{$\mathbf{+0.8}$}} & 
    $58.2$ \textcolor{brown}{\small{$\mathbf{+0.8}$}} & 
    $44.5$ \textcolor{brown}{\small{$\mathbf{+2.4}$}} & 
    $61.8$ \textcolor{brown}{\small{$\mathbf{+1.4}$}} & 
    $66.2$ \textcolor{brown}{\small{$\mathbf{+0.8}$}} & 
    $69.6$ \textcolor{brown}{\small{$\mathbf{+0.2}$}}
    \\\midrule\midrule

    \multirow{2}{*}{d} & \textcolor{gray}{\xmark} & \textcolor{brown}{\cmark} & \textcolor{brown}{\cmark} & \textcolor{gray}{\xmark} & \textcolor{gray}{\xmark} & \textcolor{gray}{\xmark} & $43.2$ \textcolor{brown} {\small{$\mathbf{+5.7}$}} & $57.1$ \textcolor{brown} {\small{$\mathbf{+4.0}$}} & $58.3$ \textcolor{brown} {\small{$\mathbf{+2.2}$}} & $59.8$ \textcolor{brown} {\small{$\mathbf{+2.4}$}} & $45.6$ \textcolor{brown} {\small{$\mathbf{+3.5}$}} & $64.3$ \textcolor{brown}{\small{$\mathbf{+3.9}$}} & $67.8$ \textcolor{brown}{\small{$\mathbf{+2.4}$}} & $71.6$ \textcolor{brown}{\small{$\mathbf{+2.2} $}} 
    \\
    & \textcolor{brown}{\cmark} & \textcolor{brown}{\cmark} & \textcolor{brown}{\cmark} & \textcolor{gray}{\xmark} & \textcolor{gray}{\xmark} & \textcolor{gray}{\xmark} & $45.3$ \textcolor{brown} {\small{$\mathbf{+7.8}$}} & $58.3$ \textcolor{brown} {\small{$\mathbf{+5.2}$}} & $58.8$ \textcolor{brown} {\small{$\mathbf{+2.7}$}} & $60.2$ \textcolor{brown} {\small{$\mathbf{+2.8}$}} & $47.0$ \textcolor{brown}{\small{$\mathbf{+4.9}$}} & $65.5$ \textcolor{brown}{\small{$\mathbf{+5.1}$}} & $69.5$ \textcolor{brown}{\small{$\mathbf{+4.1}$}} & $72.0$ \textcolor{brown}{\small{$\mathbf{+2.6}$}}
    \\\midrule

    \multirow{2}{*}{e} & \textcolor{gray}{\xmark} & \textcolor{brown}{\cmark} & \textcolor{gray}{\xmark} & \textcolor{brown}{\cmark} & \textcolor{gray}{\xmark} & \textcolor{gray}{\xmark} & $46.5$ \textcolor{brown} {\small{$\mathbf{+9.0}$}} & $59.3$ \textcolor{brown} {\small{$\mathbf{+6.2}$}} & $60.4$ \textcolor{brown} {\small{$\mathbf{+4.3}$}} & $61.9$ \textcolor{brown} {\small{$\mathbf{+4.5}$}} & $46.0$ \textcolor{brown}{\small{$\mathbf{+3.9}$}} & $64.1$ \textcolor{brown}{\small{$\mathbf{+3.7}$}} & $69.5$ \textcolor{brown}{\small{$\mathbf{+4.1}$}} & $72.3$ \textcolor{brown}{\small{$\mathbf{+2.9}$}}
    \\
    & \cellcolor{yellow!12.5}\textcolor{brown}{\cmark} & \cellcolor{yellow!12.5}\textcolor{brown}{\cmark} & \cellcolor{yellow!12.5}\textcolor{gray}{\xmark} & \cellcolor{yellow!12.5}\textcolor{brown}{\cmark} & \cellcolor{yellow!12.5}\textcolor{gray}{\xmark} & \cellcolor{yellow!12.5}\textcolor{gray}{\xmark} & \cellcolor{yellow!12.5}$\mathbf{47.4}$ \textcolor{brown}{\small{$\mathbf{+9.9}$}} & \cellcolor{yellow!12.5}$\mathbf{60.1}$ \textcolor{brown}{\small{$\mathbf{+7.0}$}} & \cellcolor{yellow!12.5}$\mathbf{61.0}$ \textcolor{brown}{\small{$\mathbf{+4.9}$}} & \cellcolor{yellow!12.5}$\mathbf{62.6}$ \textcolor{brown}{\small{$\mathbf{+5.2}$}} & \cellcolor{yellow!12.5}$\mathbf{49.5}$ \textcolor{brown}{\small{$\mathbf{+7.4}$}} & \cellcolor{yellow!12.5}$\mathbf{68.2}$ \textcolor{brown}{\small{$\mathbf{+7.8}$}} & \cellcolor{yellow!12.5}$\mathbf{70.6}$ \textcolor{brown}{\small{$\mathbf{+5.2}$}} & \cellcolor{yellow!12.5}$\mathbf{73.0}$ \textcolor{brown}{\small{$\mathbf{+3.6}$}}
    \\\midrule\midrule

    \multirow{2}{*}{f} & \textcolor{lightblue}{\cmark} & \textcolor{lightblue}{\cmark} & \textcolor{lightblue}{\cmark} & \textcolor{gray}{\xmark} & \textcolor{lightblue}{\cmark} & \textcolor{gray}{\xmark} & $48.3$ \textcolor{lightblue} {\small{$\mathbf{+0.9}$}} & $60.7$ \textcolor{lightblue} {\small{$\mathbf{+0.6}$}} & $61.4$ \textcolor{lightblue} {\small{$\mathbf{+0.4}$}} & $63.1$ \textcolor{lightblue} {\small{$\mathbf{+0.5}$}} & $50.1$ \textcolor{lightblue}{\small{$\mathbf{+0.6}$}} & $68.8$ \textcolor{lightblue}{\small{$\mathbf{+0.6}$}} & $71.0$ \textcolor{lightblue}{\small{$\mathbf{+0.4}$}} & $73.1$ \textcolor{lightblue}{\small{$\mathbf{+0.1}$}}
    \\
    & \textcolor{lightblue}{\cmark} & \textcolor{lightblue}{\cmark} & \textcolor{gray}{\xmark} & \textcolor{lightblue}{\cmark} & \textcolor{lightblue}{\cmark} & \textcolor{gray}{\xmark} & $48.6$ \textcolor{lightblue} {\small{$\mathbf{+1.2}$}} & $60.9$ \textcolor{lightblue} {\small{$\mathbf{+0.8}$}} & $61.6$ \textcolor{lightblue} {\small{$\mathbf{+0.6}$}} & $63.2$ \textcolor{lightblue} {\small{$\mathbf{+0.6}$}} &  $50.2$ \textcolor{lightblue}{\small{$\mathbf{+0.7}$}} & $69.0$ \textcolor{lightblue}{\small{$\mathbf{+0.8}$}} & $71.1$ \textcolor{lightblue}{\small{$\mathbf{+0.5}$}} & $73.2$ \textcolor{lightblue}{\small{$\mathbf{+0.2}$}}
    \\\midrule

    \multirow{2}{*}{g} & \textcolor{lightblue}{\cmark} & \textcolor{lightblue}{\cmark} & \textcolor{lightblue}{\cmark} & \textcolor{gray}{\xmark} & \textcolor{gray}{\xmark} & \textcolor{lightblue}{\cmark} & $49.6$ \textcolor{lightblue} {\small{$\mathbf{+2.2}$}} & $61.2$ \textcolor{lightblue} {\small{$\mathbf{+1.1}$}} & $61.9$ \textcolor{lightblue} {\small{$\mathbf{+0.9}$}} & $63.4$ \textcolor{lightblue} {\small{$\mathbf{+0.8}$}} & $50.8$ \textcolor{lightblue}{\small{$\mathbf{+1.3}$}} & $69.4$ \textcolor{lightblue}{\small{$\mathbf{+1.2}$}} & $71.3$ \textcolor{lightblue}{\small{$\mathbf{+0.7}$}} & $73.3$ \textcolor{lightblue}{\small{$\mathbf{+0.3}$}}
    \\
    & \textcolor{lightblue}{\cmark} & \textcolor{lightblue}{\cmark} & \textcolor{gray}{\xmark} & \textcolor{lightblue}{\cmark} & \textcolor{gray}{\xmark} & \textcolor{lightblue}{\cmark} & $49.7$ \textcolor{lightblue} {\small{$\mathbf{+2.3}$}} & $61.4$ \textcolor{lightblue} {\small{$\mathbf{+1.3}$}} & $62.2$ \textcolor{lightblue} {\small{$\mathbf{+1.2}$}} & $63.5$ \textcolor{lightblue} {\small{$\mathbf{+0.9}$}} & $51.0$ \textcolor{lightblue}{\small{$\mathbf{+1.5}$}} & $69.5$ \textcolor{lightblue}{\small{$\mathbf{+1.3}$}} & $71.5$ \textcolor{lightblue}{\small{$\mathbf{+0.9}$}} & $73.5$ \textcolor{lightblue}{\small{$\mathbf{+0.5}$}}
    \\\midrule

    h & \cellcolor{lightblue!9}\textcolor{lightblue}{\cmark} & \cellcolor{lightblue!9}\textcolor{lightblue}{\cmark} & \cellcolor{lightblue!9}\textcolor{gray}{\xmark} & \cellcolor{lightblue!9}\textcolor{lightblue}{\cmark} & \cellcolor{lightblue!9}\textcolor{lightblue}{\cmark} & \cellcolor{lightblue!9}\textcolor{lightblue}{\cmark} & \cellcolor{lightblue!9}$\mathbf{50.1}$ \textcolor{lightblue}{\small{$\mathbf{+2.7}$}} & \cellcolor{lightblue!9}$\mathbf{61.9}$ \textcolor{lightblue}{\small{$\mathbf{+1.8}$}} & \cellcolor{lightblue!9}$\mathbf{62.4}$ \textcolor{lightblue}{\small{$\mathbf{+1.4}$}} & \cellcolor{lightblue!9}$\mathbf{63.7}$ \textcolor{lightblue}{\small{$\mathbf{+1.1}$}} & \cellcolor{lightblue!9}$\mathbf{51.6}$ \textcolor{lightblue}{\small{$\mathbf{+2.1}$}} & \cellcolor{lightblue!9}$\mathbf{69.8}$ \textcolor{lightblue}{\small{$\mathbf{+1.6}$}} & \cellcolor{lightblue!9}$\mathbf{71.7}$ \textcolor{lightblue}{\small{$\mathbf{+1.1}$}} & \cellcolor{lightblue!9}$\mathbf{73.7}$ \textcolor{lightblue}{\small{$\mathbf{+0.7}$}}
\\\bottomrule
\end{tabular}}
\label{tab:ablation}
\end{table*}

\noindent\textbf{Compare with Fully Supervised Methods.}
As shown in Figure~\ref{fig:teaser-performance}, the comparisons of LaserMix++ over the prevailing LiDAR semantic segmentation methods \cite{Cylinder3D,PolarNet,SalsaNext,PolarStream} validate that our approaches are competitive to the fully supervised counterparts while only requiring 2$\times$ to 5$\times$ fewer annotations. Additionally, the results in Table~\ref{tab:benchmark} verify the strong augmentation and regularization ability of LaserMix++ again. Our approaches have yielded better results in the high-data regime and extreme low-data regime (\emph{i.e.}, $0.8\%$ human annotations on ScribbleKITTI~\cite{ScribbleKITTI}).

\begin{wraptable}{r}{0.5\textwidth}
\centering
\vspace{-0.6cm}
\caption{\textbf{Benchmark results} of different semi-supervised learning approaches on the Cityscapes~\cite{Cityscapes} dataset. (a) Methods using MeanTeacher \cite{MeanTeacher} as the backbone. (b) Methods using CPS \cite{CPS} as the backbone with a CutMix \cite{CutMix} augmentation. All mIoU scores are given in percentage ($\%$).}
\vspace{-0.2cm}
\label{tab:benchmark-cityscapes}
\setlength{\tabcolsep}{4pt}
\resizebox{\linewidth}{!}{
\begin{tabular}{p{10pt}<{\centering}|p{74pt}<{\centering}|cccc}
\toprule
\textbf{\#} & \textbf{Method} & $\mathbf{1/16}$ & $\mathbf{1/8}$ & $\mathbf{1/4}$ & $\mathbf{1/2}$ 
\\\midrule\midrule
\multirow{6}{*}{a} & \cellcolor{gray!10}MeanTeacher~\cite{MeanTeacher} & \cellcolor{gray!10}$66.1$ & \cellcolor{gray!10}$71.2$ & \cellcolor{gray!10}$74.4$ & \cellcolor{gray!10}$76.3$
\\\cmidrule{2-6}
& \textbf{\emph{w/} Ours} & \cellcolor{yellow!12.5}$\mathbf{68.7}$ \textcolor{brown}{\small{$\mathbf{+2.6}$}} & \cellcolor{yellow!12.5}$\mathbf{72.3}$ \textcolor{brown}{\small{$\mathbf{+1.1}$}} & \cellcolor{yellow!12.5}$\mathbf{75.7}$ \textcolor{brown}{\small{$\mathbf{+1.3}$}} & \cellcolor{yellow!12.5}$\mathbf{76.8}$ \textcolor{brown}{\small{$\mathbf{+0.5}$}}
\\\cmidrule{2-6}
& CCT~\cite{CCT} & $66.4$ & $72.5$ & $75.7$ & $76.8$
\\
& GCT~\cite{GCT} & $65.8$ & $71.3$ & $75.3$ & $77.1$
\\
& CPS~\cite{CPS} & $69.8$ & $74.4$ & $76.9$ & $78.6$
\\\midrule\midrule
\multirow{2.5}{*}{b} & \cellcolor{gray!10}CPS-CutMix~\cite{CPS} & \cellcolor{gray!10}$74.5$ & \cellcolor{gray!10}$76.6$ & \cellcolor{gray!10}$77.8$ & \cellcolor{gray!10}$78.8$
\\\cmidrule{2-6}
& \textbf{\emph{w/} Ours} & \cellcolor{lightblue!9}$\mathbf{75.5}$ \textcolor{lightblue}{\small{$\mathbf{+1.0}$}} & \cellcolor{lightblue!9}$\mathbf{77.1}$ \textcolor{lightblue}{\small{$\mathbf{+0.5}$}} & \cellcolor{lightblue!9}$\mathbf{78.3}$ \textcolor{lightblue}{\small{$\mathbf{+0.5}$}} & \cellcolor{lightblue!9}$\mathbf{79.1}$ \textcolor{lightblue}{\small{$\mathbf{+0.3}$}}
\\\bottomrule
\end{tabular}}
\label{tab:cityscapes}
\vspace{-0.7cm}
\end{wraptable}
\noindent\textbf{Qualitative Assessments.} Figure~\ref{fig:qualitative} displays visualizations of the 3D scene segmentation results for different semi-supervised learning algorithms on the validation set of nuScenes~\cite{Panoptic-nuScenes}, where each example covers a 50$\times$50 $\text{m}^2$ driving scene centered by the ego-vehicle. We observe that previous methods can only improve predictions in limited regions, while our approaches holistically eliminate false predictions in almost every region around the ego-vehicle. The consistency enlightened by our methods has yielded better 3D segmentation accuracy under annotation scarcity.

\noindent\textbf{Enhancing Representation Learning Effects.}
Recent explorations on self-supervised LiDAR semantic segmentation exhibit promising representation learning effects during model pretraining \cite{xie2020pointContrast,zhang2021depthContrast,liu2021ppkt,sautier2022slidr,mahmoud2023st,liu2023segment}. Such methods establish suitable self-supervised pretraining objectives and probe the qualitative aspects of representation learning via few-shot fine-tuning.

The results shown in Table~\ref{tab:representation} verify that our approaches are effective in enhancing the effects of representation learning during fine-tuning. combined with Seal \cite{liu2023segment}, LaserMix++ also achieves superior performance under full supervision (from 74.7\% mIoU to 77.9\% mIoU). This study further proves the versatility of our framework in handling different LiDAR semantic segmentation tasks.

\noindent\textbf{Enhancing Out-of-Distribution Robustness.}
The ability to tackle scenarios that are not observed during the training time is crucial for a 3D scene understanding model, especially under the autonomous driving context \cite{kong2023robodepth,xie2023robobev}. The fine-grained manipulations of LiDAR point clouds in LaserMix and LaserMix++ have the potential to enrich the training distribution and, in return, achieve better robustness. In this work, we conduct experiments on the Robo3D benchmark \cite{kong2023robo3D} to validate the robustness enhancement effect of our framework. 

As shown in Table~\ref{tab:robustness}, our approaches consistently yield lower mCE scores for both LiDAR semantic segmentation and 3D object detection compared to other mixing-based techniques, such as Mix3D \cite{Mix3D} and PolarMix \cite{xiao2022polarmix}.

\begin{wraptable}{r}{0.55\textwidth}
\centering
\vspace{-0.3cm}
\caption{\textbf{Ablation study} on laser beam partition strategies in LaserMix. Horizontal axis: inclination direction $\phi$; Vertical axis: azimuth direction $\alpha$. A ($i$-$\alpha$, $j$-$\phi$) strategy denotes that there are $i$ azimuth and $j$ inclination partitions in total.}
\vspace{-0.2cm}
\centering
\resizebox{\linewidth}{!}{
\begin{tabular}{p{38.45pt}<{\centering}|p{38.45pt}<{\centering}|p{38.45pt}<{\centering}|p{38.45pt}<{\centering}|p{38.45pt}<{\centering}|p{38.45pt}<{\centering}}
\toprule
\rowcolor[gray]{.95} Baseline & ($1\alpha$, $2\phi$) & ($1\alpha$, $3\phi$) & ($1\alpha$, $4\phi$) & ($1\alpha$, $5\phi$) & ($1\alpha$, $6\phi$)
\\\midrule
\begin{minipage}[b]{0.085\columnwidth}\centering\raisebox{-.37\height}{\includegraphics[width=\linewidth]{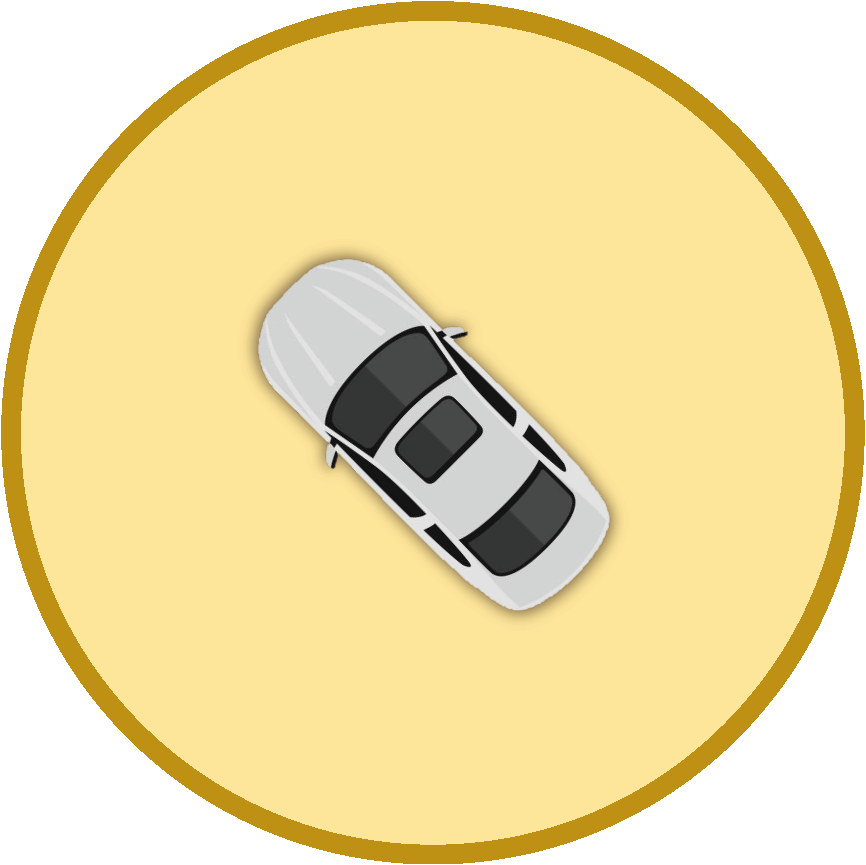}}\end{minipage} &
\begin{minipage}[b]{0.085\columnwidth}\centering\raisebox{-.37\height}{\includegraphics[width=\linewidth]{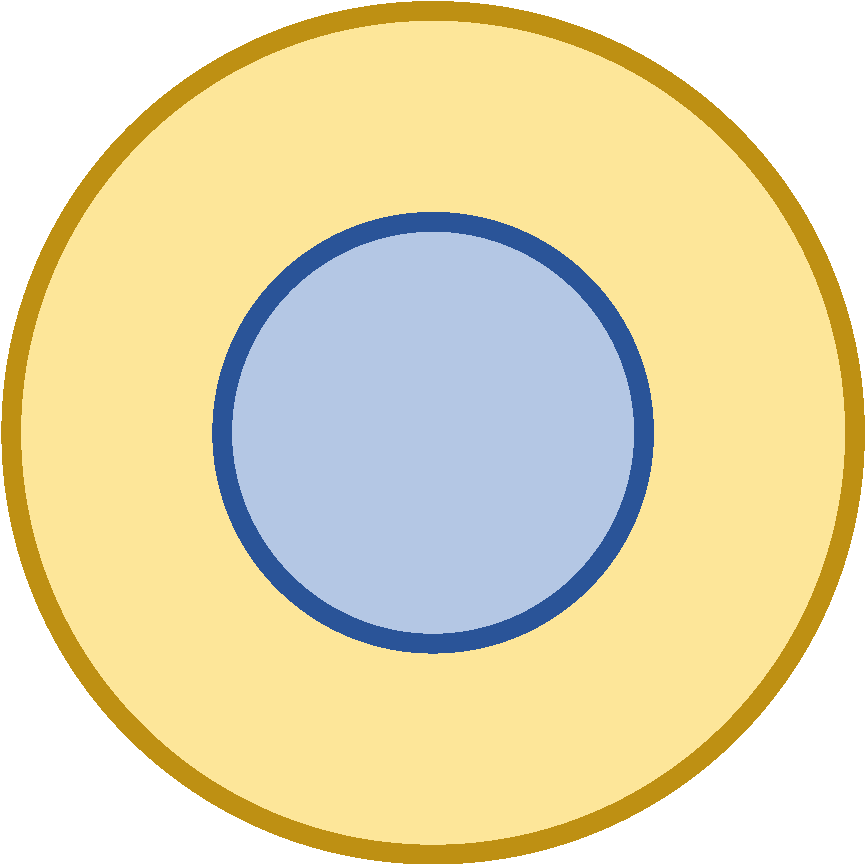}}\end{minipage} &
\begin{minipage}[b]{0.085\columnwidth}\centering\raisebox{-.37\height}{\includegraphics[width=\linewidth]{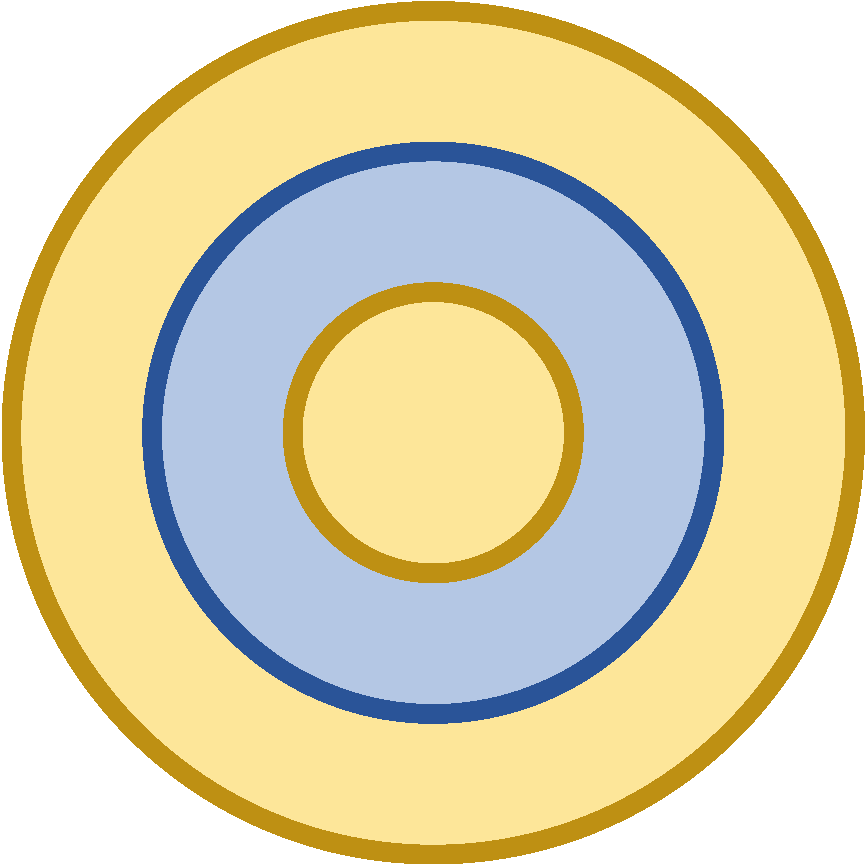}}\end{minipage} &
\begin{minipage}[b]{0.085\columnwidth}\centering\raisebox{-.37\height}{\includegraphics[width=\linewidth]{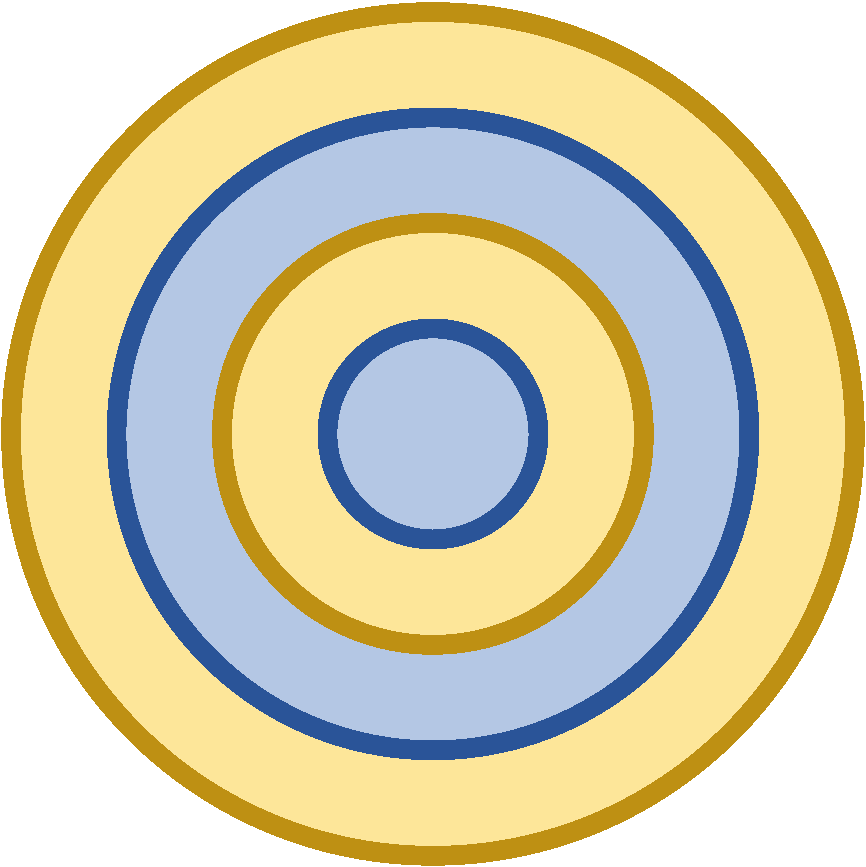}}\end{minipage} &
\begin{minipage}[b]{0.085\columnwidth}\centering\raisebox{-.37\height}{\includegraphics[width=\linewidth]{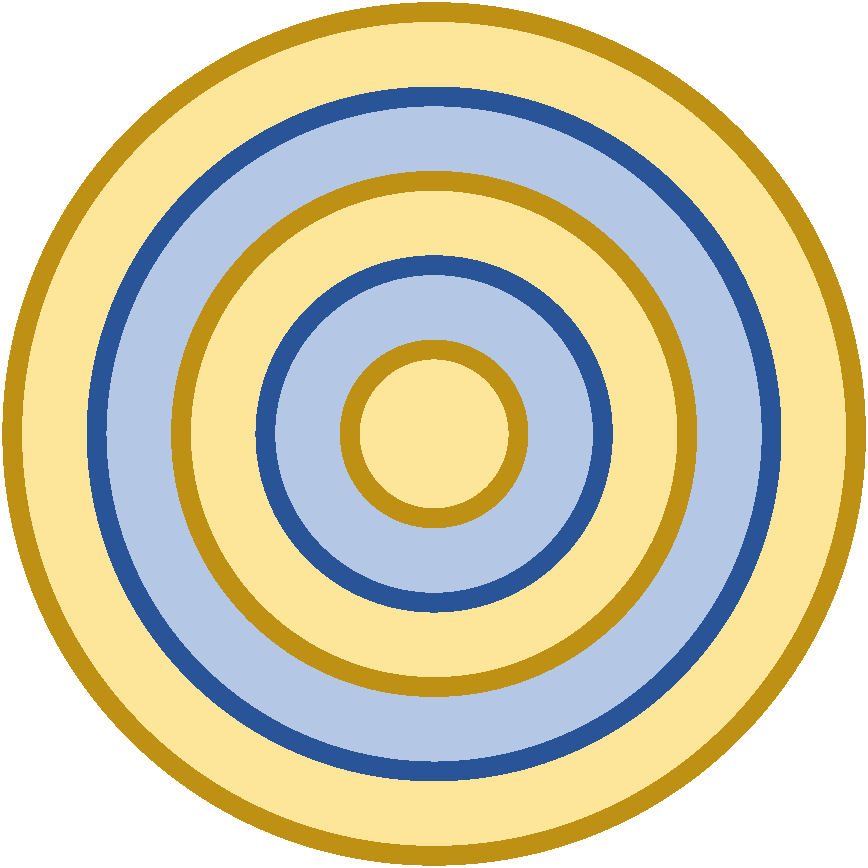}}\end{minipage} &
\begin{minipage}[b]{0.085\columnwidth}\centering\raisebox{-.37\height}{\includegraphics[width=\linewidth]{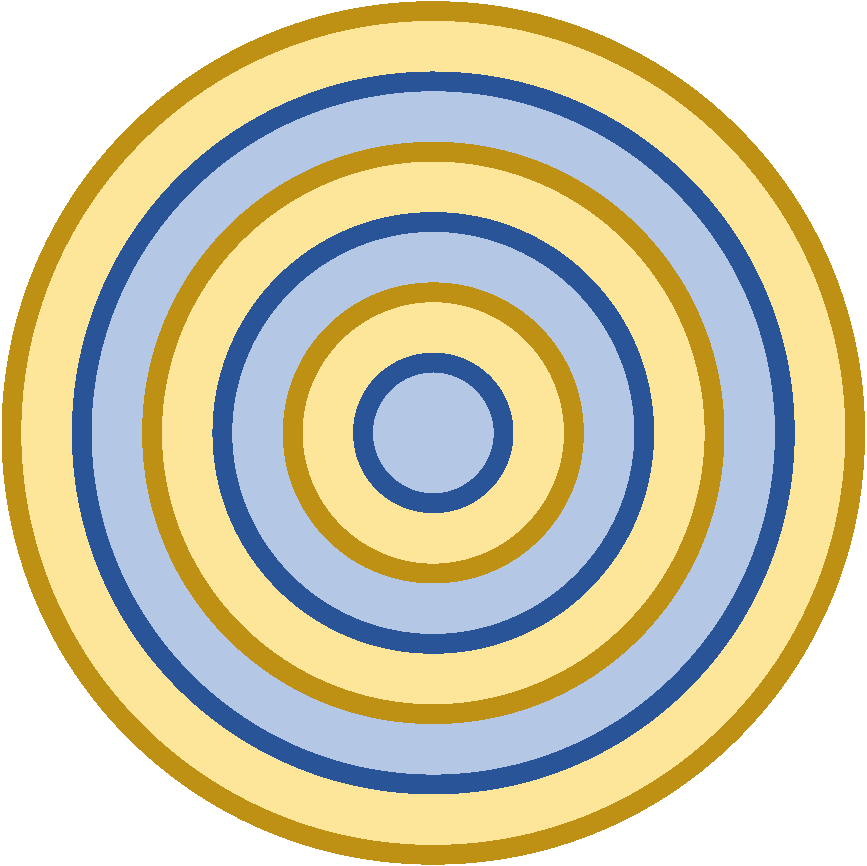}}\end{minipage} 
\\\midrule
$60.4$ & $63.5_\mathbf{{\textcolor{brown}{(+3.1)}}}$ & $65.2_\mathbf{{\textcolor{brown}{(+4.8)}}}$ & $66.5_\mathbf{{\textcolor{brown}{(+6.1)}}}$ & $66.2_\mathbf{{\textcolor{brown}{(+5.8)}}}$ & $65.4_\mathbf{{\textcolor{brown}{(+5.0)}}}$ 
\\\midrule
\rowcolor[gray]{.95} ($2\alpha$, $1\phi$) & ($2\alpha$, $2\phi$) & ($2\alpha$, $3\phi$) & ($2\alpha$, $4\phi$) & ($2\alpha$, $5\phi$) & ($2\alpha$, $6\phi$)
\\\midrule
\begin{minipage}[b]{0.085\columnwidth}\centering\raisebox{-.37\height}{\includegraphics[width=\linewidth]{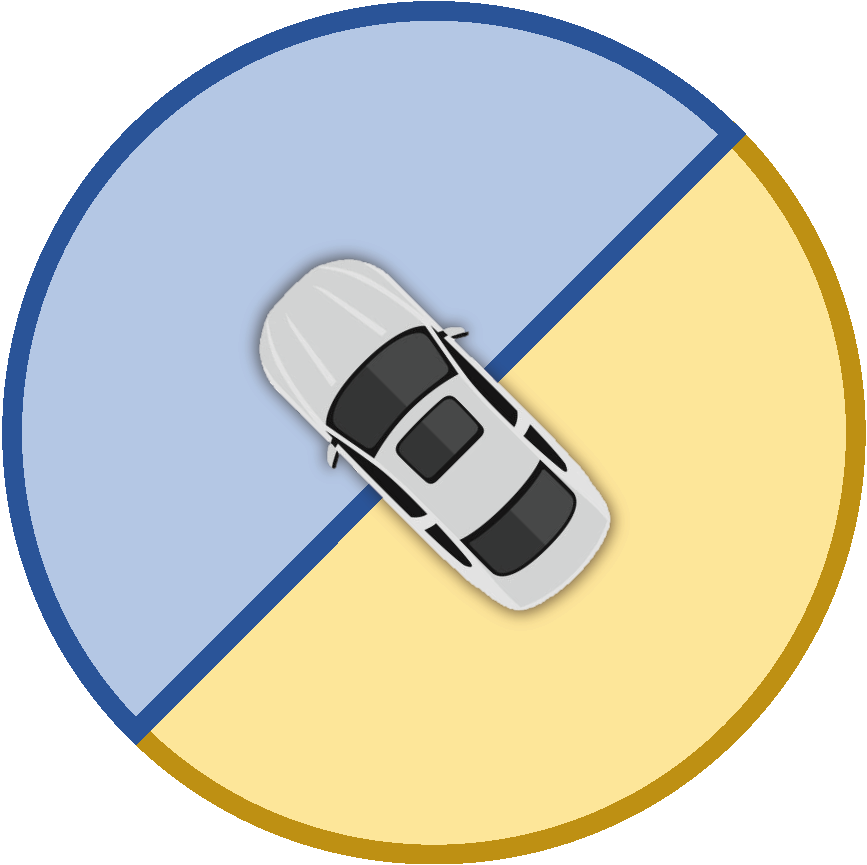}}\end{minipage} & \begin{minipage}[b]{0.085\columnwidth}\centering\raisebox{-.37\height}{\includegraphics[width=\linewidth]{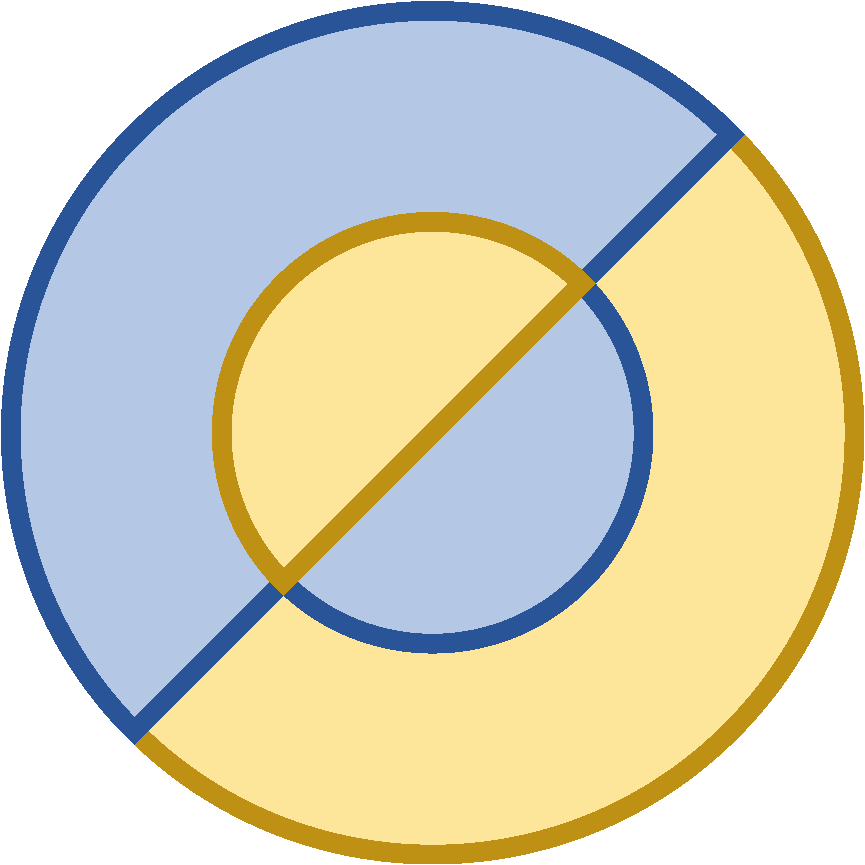}}\end{minipage} &
\begin{minipage}[b]{0.085\columnwidth}\centering\raisebox{-.37\height}{\includegraphics[width=\linewidth]{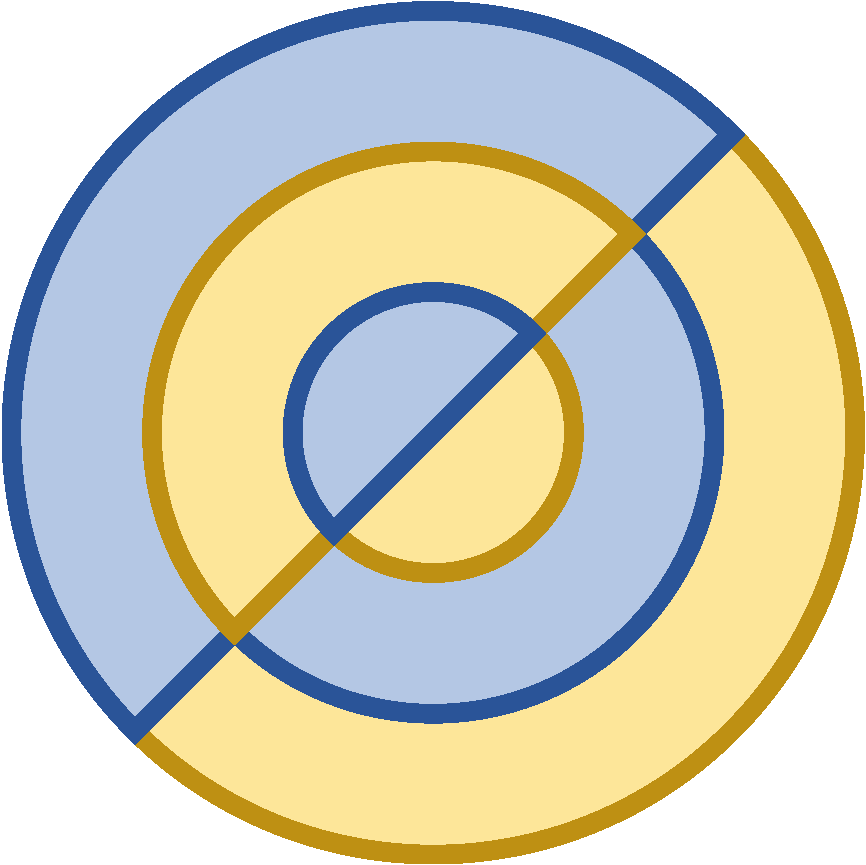}}\end{minipage} &
\begin{minipage}[b]{0.085\columnwidth}\centering\raisebox{-.37\height}{\includegraphics[width=\linewidth]{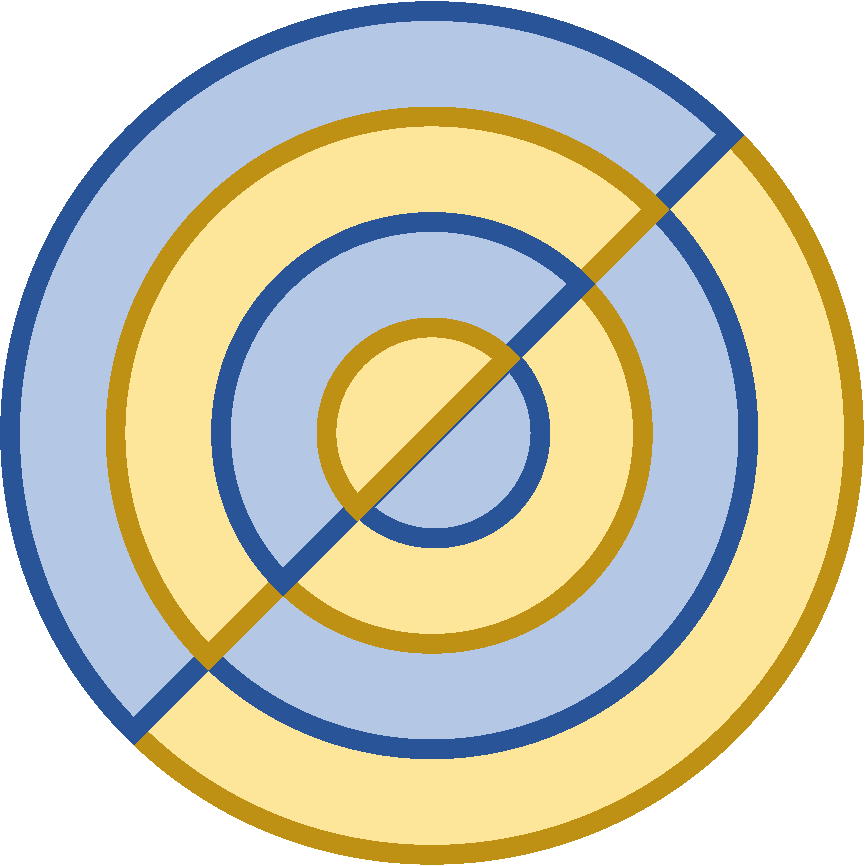}}\end{minipage} &
\begin{minipage}[b]{0.085\columnwidth}\centering\raisebox{-.37\height}{\includegraphics[width=\linewidth]{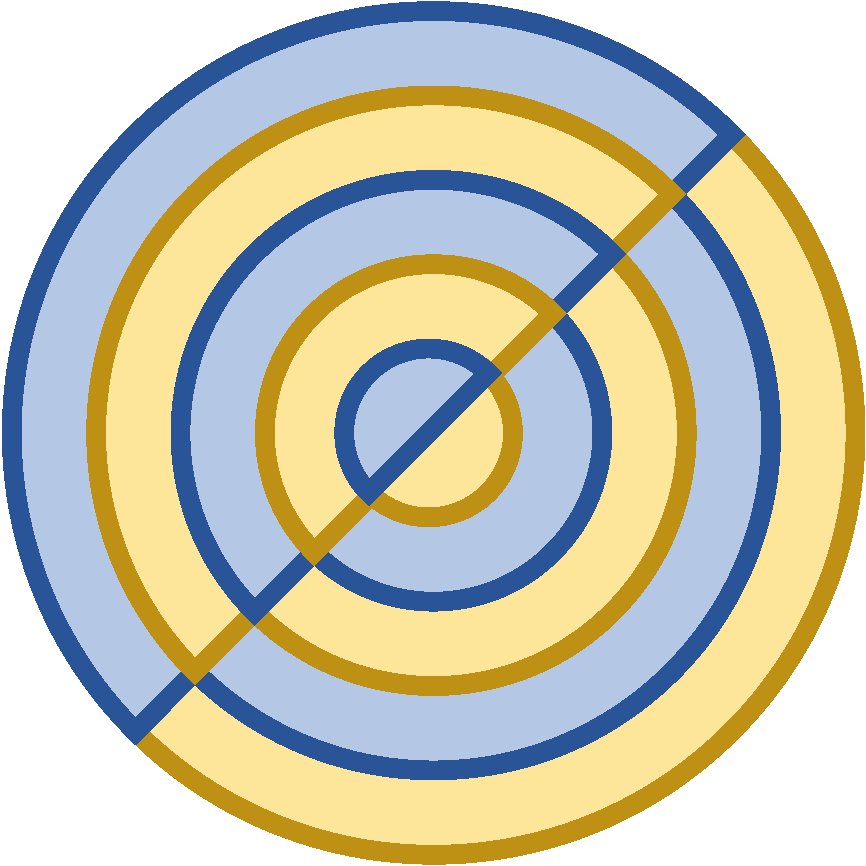}}\end{minipage} &
\begin{minipage}[b]{0.085\columnwidth}\centering\raisebox{-.37\height}{\includegraphics[width=\linewidth]{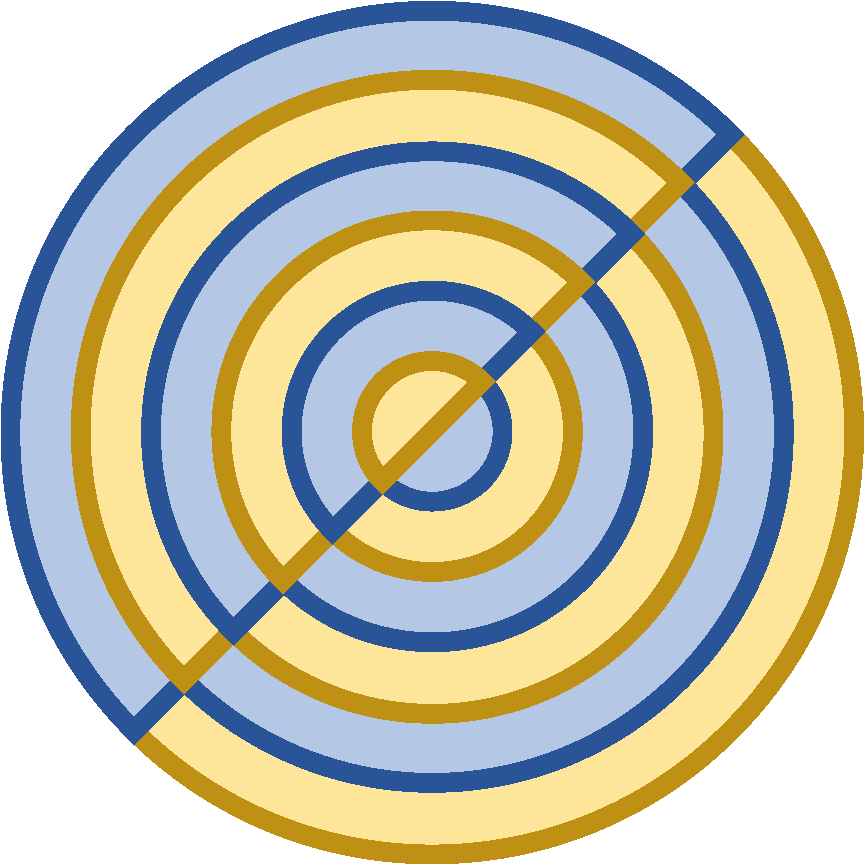}}\end{minipage} 
\\\midrule
$61.5_\mathbf{{\textcolor{brown}{(+1.1)}}}$ & $63.3_\mathbf{{\textcolor{brown}{(+2.9)}}}$ & $65.9_\mathbf{{\textcolor{brown}{(+5.5)}}}$ & $66.1_\mathbf{{\textcolor{brown}{(+5.7)}}}$ & $66.7_\mathbf{{\textcolor{brown}{(+6.3)}}}$ & $65.3_\mathbf{{\textcolor{brown}{(+4.9)}}}$ 
\\\midrule
\rowcolor[gray]{.95} ($3\alpha$, $1\phi$) & ($3\alpha$, $2\phi$) & ($3\alpha$, $3\phi$) & ($3\alpha$, $4\phi$) & ($3\alpha$, $5\phi$) & ($3\alpha$, $6\phi$)
\\\midrule
\begin{minipage}[b]{0.085\columnwidth}\centering\raisebox{-.37\height}{\includegraphics[width=\linewidth]{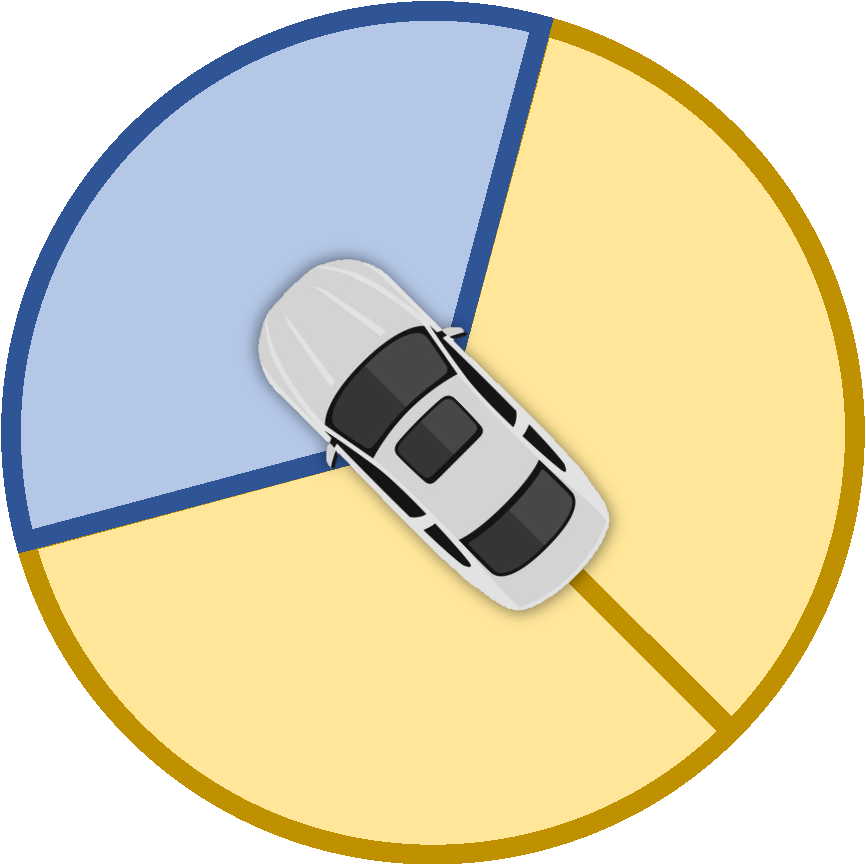}}\end{minipage} & \begin{minipage}[b]{0.085\columnwidth}\centering\raisebox{-.37\height}{\includegraphics[width=\linewidth]{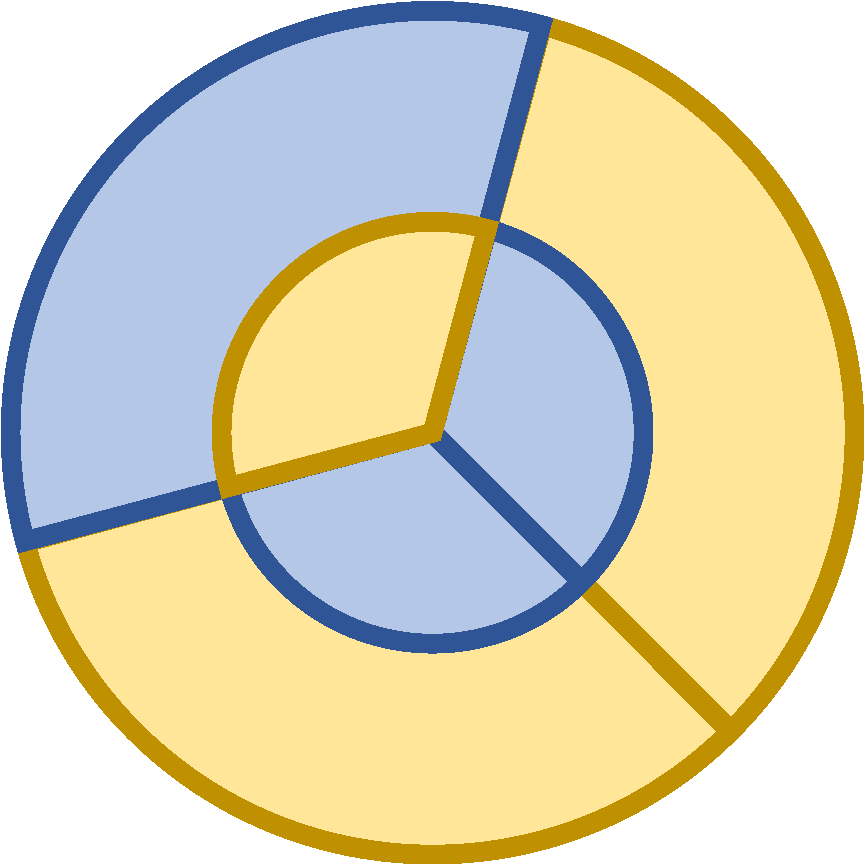}}\end{minipage} & \begin{minipage}[b]{0.085\columnwidth}\centering\raisebox{-.37\height}{\includegraphics[width=\linewidth]{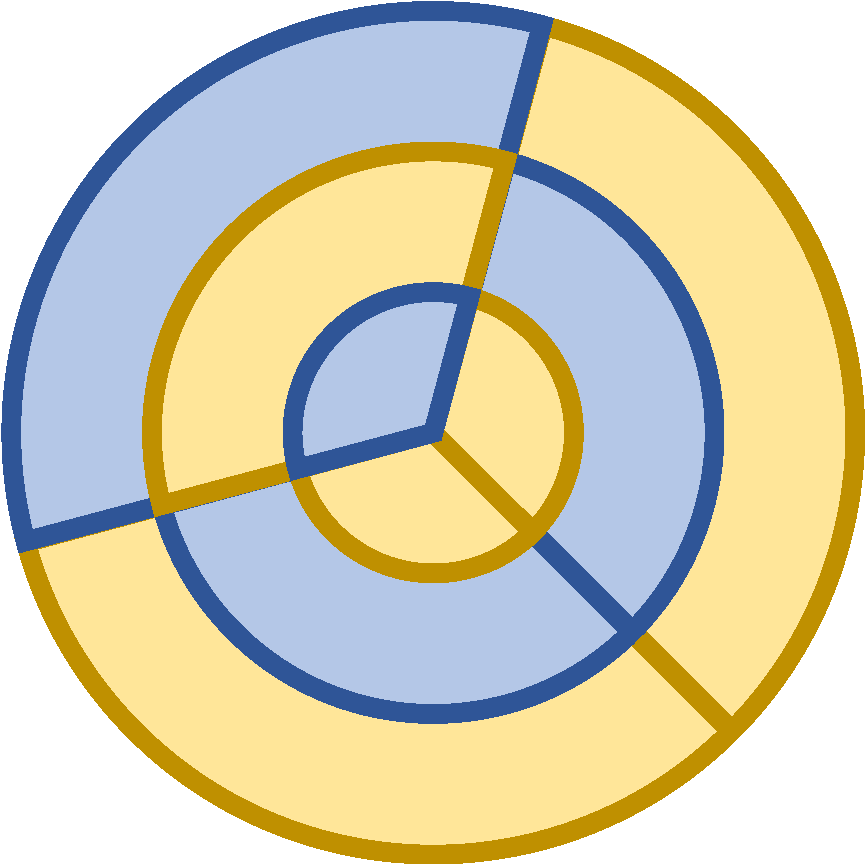}}\end{minipage} & \begin{minipage}[b]{0.085\columnwidth}\centering\raisebox{-.37\height}{\includegraphics[width=\linewidth]{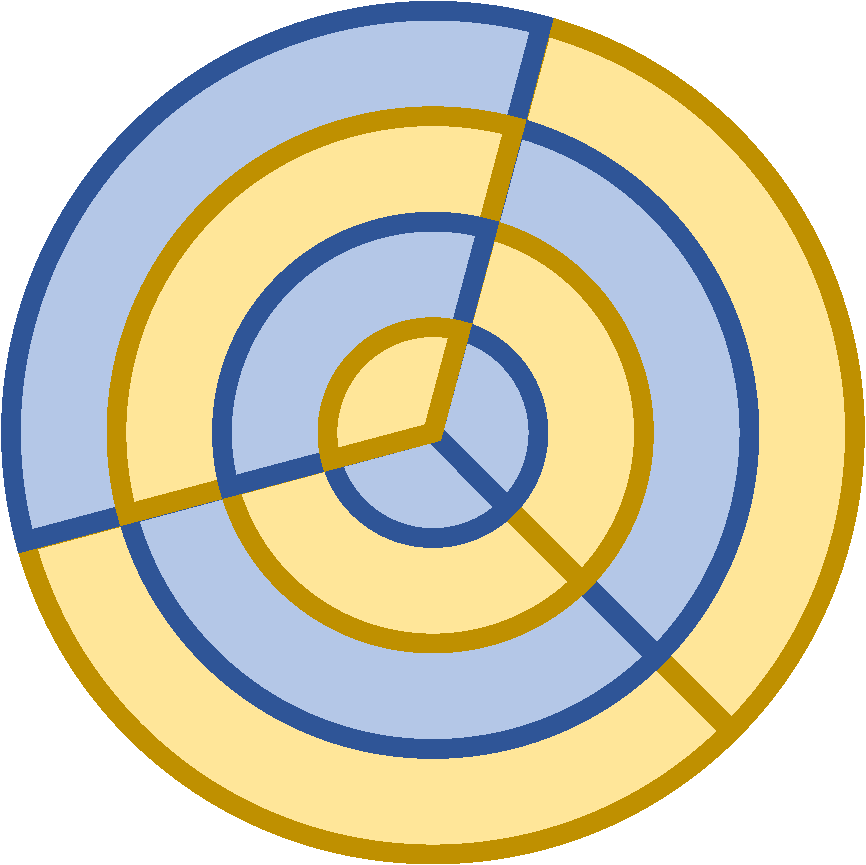}}\end{minipage} & \begin{minipage}[b]{0.085\columnwidth}\centering\raisebox{-.37\height}{\includegraphics[width=\linewidth]{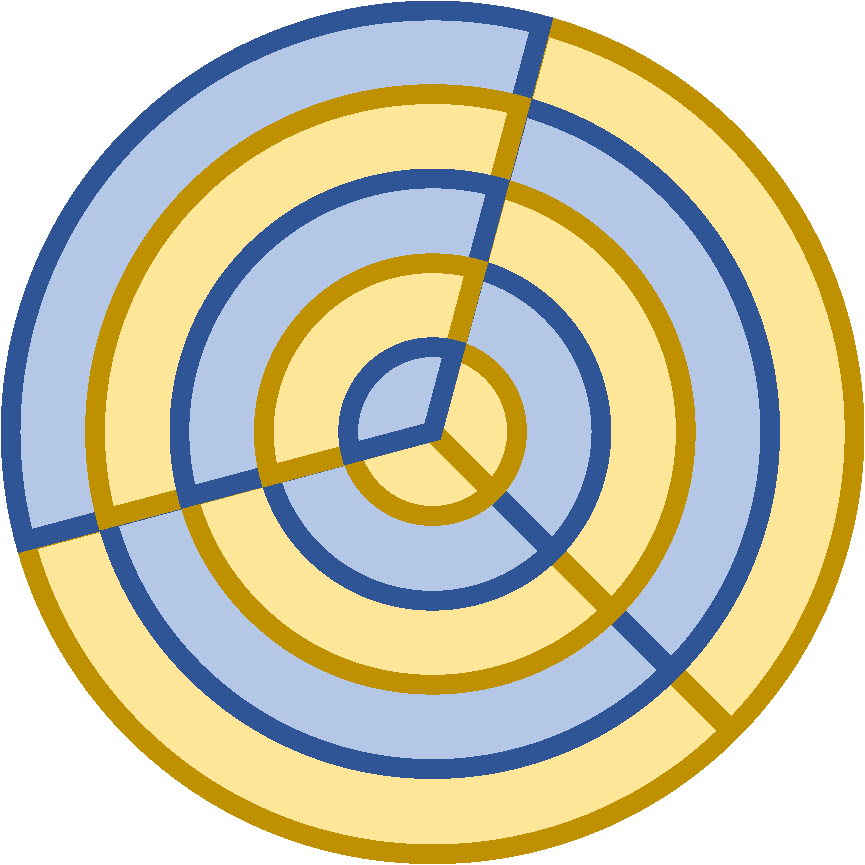}}\end{minipage} & \begin{minipage}[b]{0.085\columnwidth}\centering\raisebox{-.37\height}{\includegraphics[width=\linewidth]{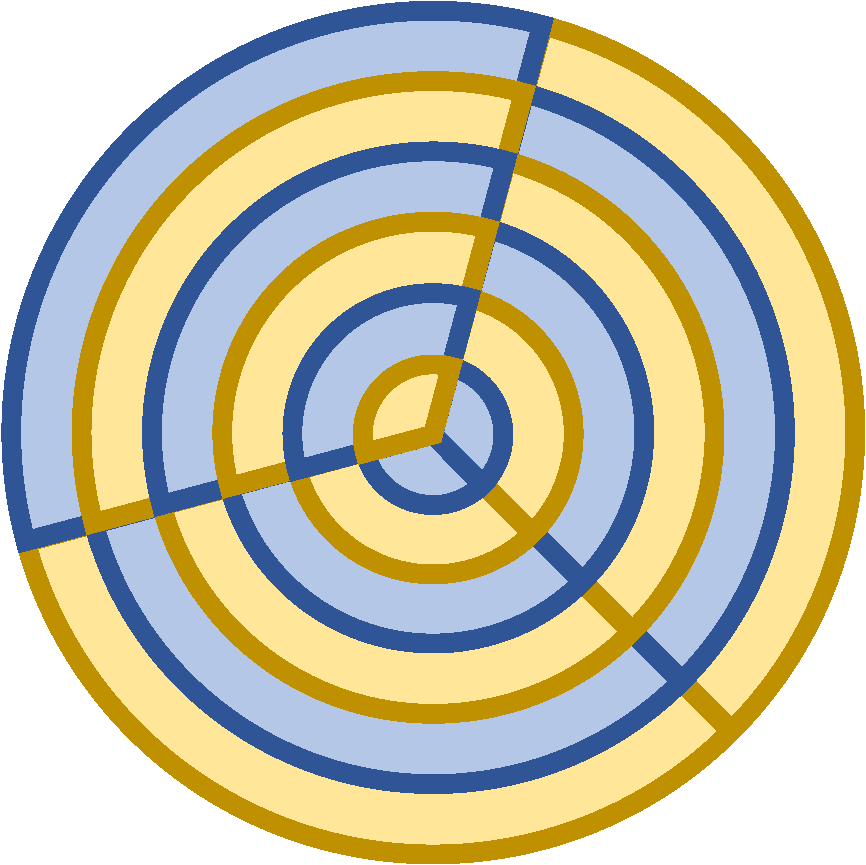}}\end{minipage} 
\\\midrule
$60.9_\mathbf{{\textcolor{brown}{(+0.6)}}}$ & $64.2_\mathbf{{\textcolor{brown}{(+3.8)}}}$ & $65.9_\mathbf{{\textcolor{brown}{(+5.5)}}}$ & $66.3_\mathbf{{\textcolor{brown}{(+5.9)}}}$ & $66.0_\mathbf{{\textcolor{brown}{(+5.6)}}}$ & $65.2_\mathbf{{\textcolor{brown}{(+4.8)}}}$ 
\\\midrule
\rowcolor[gray]{.95} ($4\alpha$, $1\phi$) & ($4\alpha$, $2\phi$) & ($4\alpha$, $3\phi$) & ($4\alpha$, $4\phi$) & ($4\alpha$, $5\phi$) & ($4\alpha$, $6\phi$)
\\\midrule
\begin{minipage}[b]{0.085\columnwidth}\centering\raisebox{-.37\height}{\includegraphics[width=\linewidth]{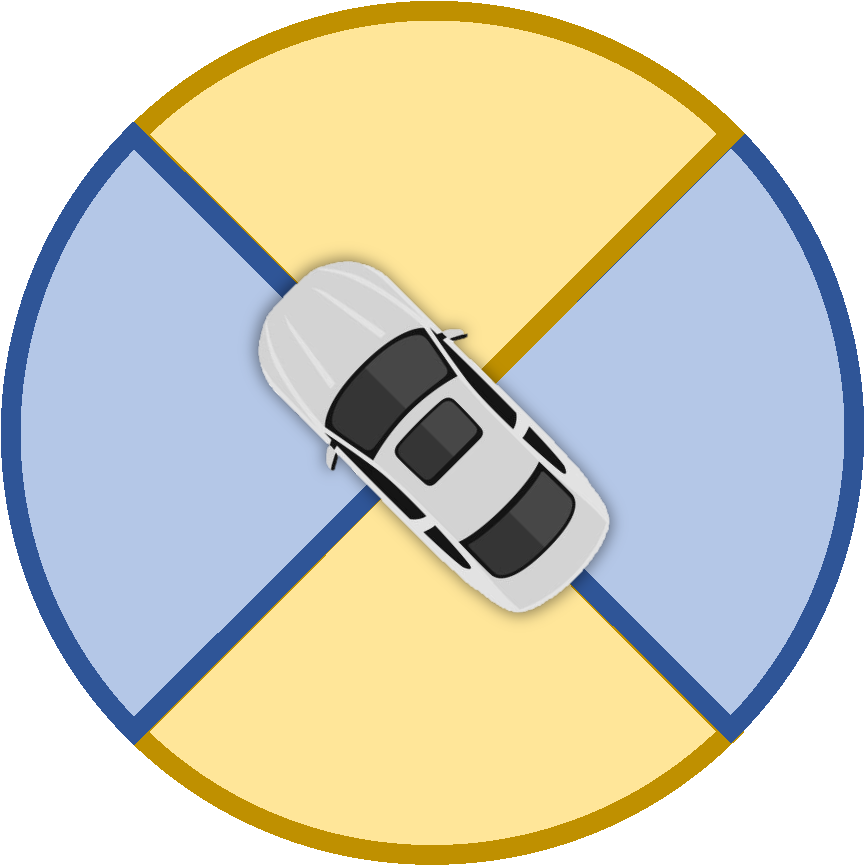}}\end{minipage} & \begin{minipage}[b]{0.085\columnwidth}\centering\raisebox{-.37\height}{\includegraphics[width=\linewidth]{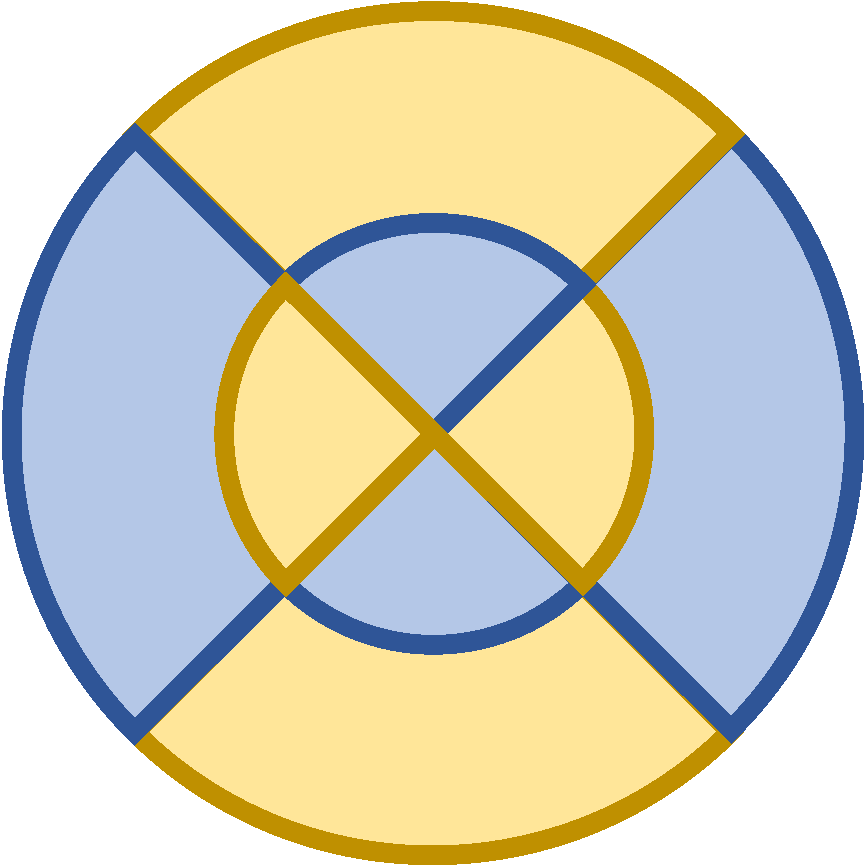}}\end{minipage} & \begin{minipage}[b]{0.085\columnwidth}\centering\raisebox{-.37\height}{\includegraphics[width=\linewidth]{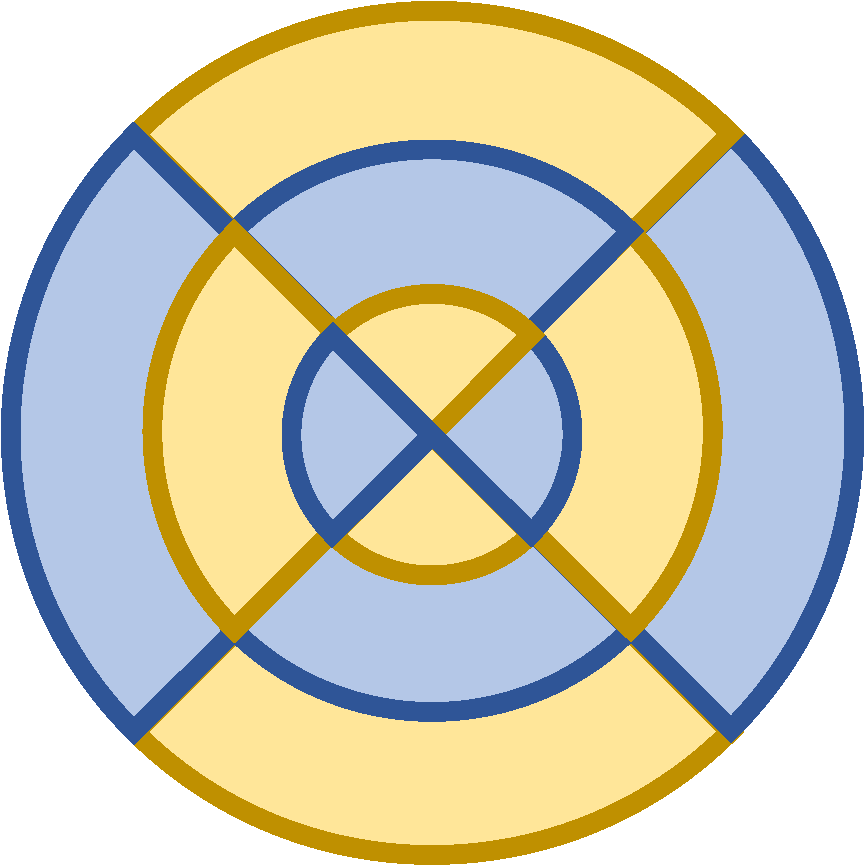}}\end{minipage} & \begin{minipage}[b]{0.085\columnwidth}\centering\raisebox{-.37\height}{\includegraphics[width=\linewidth]{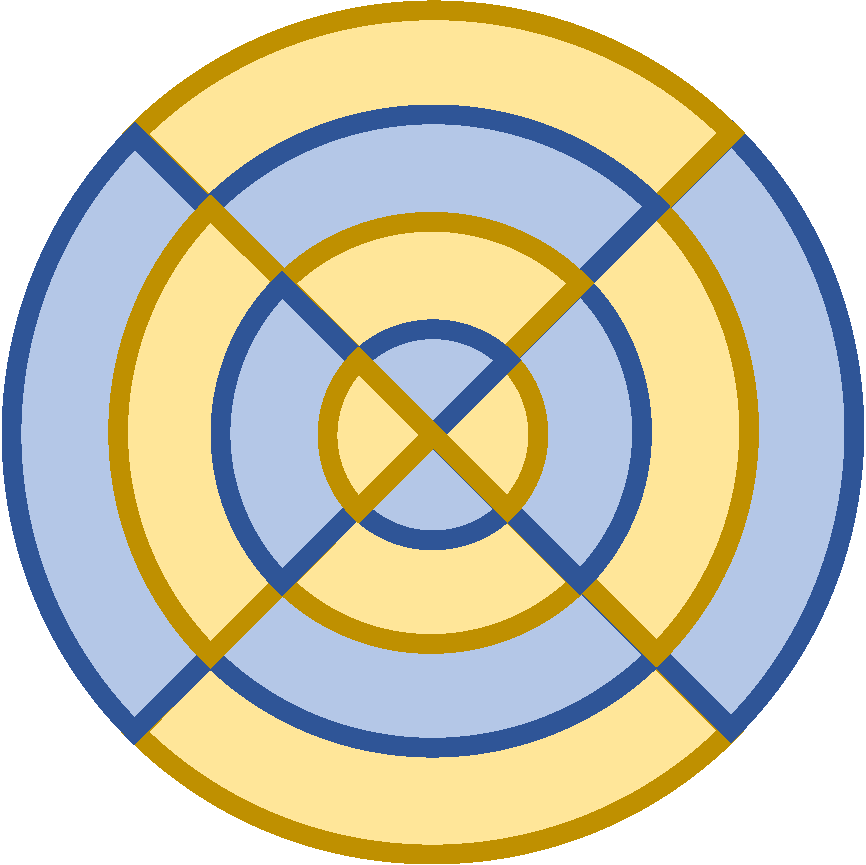}}\end{minipage} & \begin{minipage}[b]{0.085\columnwidth}\centering\raisebox{-.37\height}{\includegraphics[width=\linewidth]{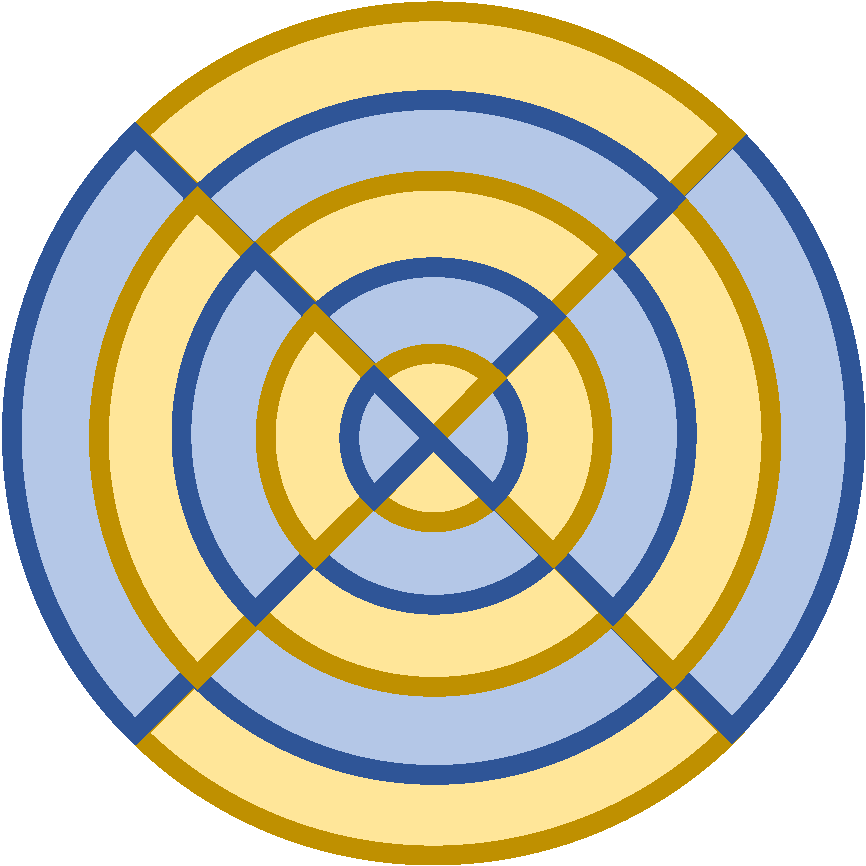}}\end{minipage} & \begin{minipage}[b]{0.085\columnwidth}\centering\raisebox{-.37\height}{\includegraphics[width=\linewidth]{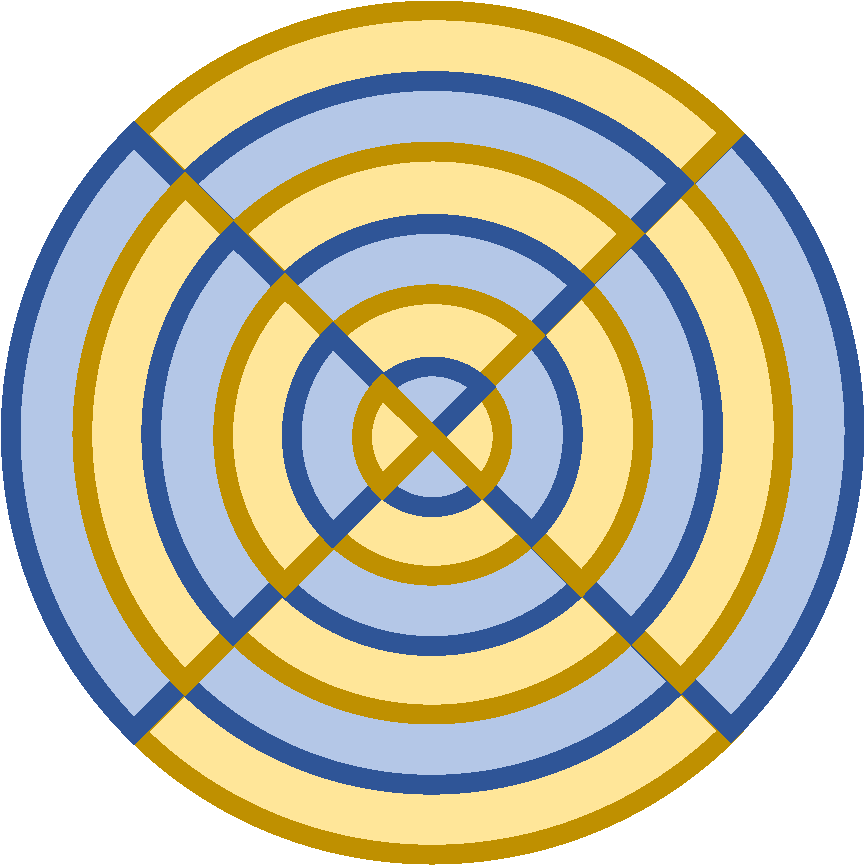}}\end{minipage} 
\\\midrule
$60.9_\mathbf{{\textcolor{brown}{(+0.6)}}}$ & $64.7_\mathbf{{\textcolor{brown}{(+4.3)}}}$ & $65.3_\mathbf{{\textcolor{brown}{(+4.9)}}}$ & $65.6_\mathbf{{\textcolor{brown}{(+5.2)}}}$ & $65.7_\mathbf{{\textcolor{brown}{(+5.3)}}}$ & $65.2_\mathbf{{\textcolor{brown}{(+4.8)}}}$ 
\\
\bottomrule
\end{tabular}}
\label{tab:granularity}
\vspace{-0.7cm}
\end{wraptable}
\noindent\textbf{Generalize to Image Segmentation.}
The proposed scene prior-based consistency regularization can be extended to image domains since they also reflect real-world driving scenarios. To validate this, we conduct experiments on Cityscapes \cite{Cityscapes} and show the results in Table~\ref{tab:cityscapes}. As can be seen, our approaches achieved improved performance over the strong image segmentation baselines \cite{MeanTeacher,CPS,CCT,GCT}. Such a generalizability ensures our approaches are universally applicable to different sensor modalities.

\subsection{Ablation Study}
In this section, we conduct several ablation studies to verify the effectiveness of each component. Without otherwise mentioned, we stick with the 10\% label budget setting and the range view backbone in our ablation experiments.

\begin{wraptable}{r}{0.5\textwidth}
\centering
\vspace{-0.3cm}
\caption{\textbf{Ablation study} on state-of-the-art 3D SSL methods (\emph{w/} the same Cylinder3D \cite{Cylinder3D} backbone except for LiM3D \cite{li23lim3d}) under different data splitting strategies, \ie, (1) random, (2) uniform, (3) sequential, and (4) spatio-temporal redundant frame downsampling (ST-RFD) strategies. All mIoU scores are given in percentage ($\%$). The \textcolor{gray}{{\emph{sup.-only}}}, \textcolor{lightblue}{{\emph{best}}}, and \textcolor{brown}{{\emph{second best}}} scores under each data split within each data splitting group are shaded with \textcolor{gray}{{\emph{gray}}}, \textcolor{lightblue}{{\emph{blue}}}, and \textcolor{brown}{{\emph{yellow}}}, respectively.}
\vspace{-0.2cm}
\resizebox{\linewidth}{!}{
\begin{tabular}{c|c|p{21.0pt}<{\centering}p{21.0pt}<{\centering}p{21.0pt}<{\centering}|p{21.0pt}<{\centering}p{21.0pt}<{\centering}p{21.0pt}<{\centering}}
\toprule
\multirow{2}{*}{\textbf{Split}} & \multirow{2}{*}{\textbf{Method}} & \multicolumn{3}{c|}{\textbf{SemanticKITTI}~\cite{SemanticKITTI}} & \multicolumn{3}{c}{\textbf{ScribbleKITTI}~\cite{ScribbleKITTI}}
\\
& & \textbf{1\%} & \textbf{10\%} & \textbf{20\%} & \textbf{1\%} & \textbf{10\%} & \textbf{20\%}
\\\midrule\midrule

\multirow{7}{*}{\rotatebox{90}{Random}} &
\cellcolor{gray!10}\emph{Sup.-only} & \cellcolor{gray!10}$45.4$ & \cellcolor{gray!10}$56.1$ & \cellcolor{gray!10}$57.8$ & \cellcolor{gray!10}$39.2$ & \cellcolor{gray!10}$48.0$ & \cellcolor{gray!10}$52.1$
\\\cmidrule{2-8}
& MT~\cite{MeanTeacher} & $45.4$ & $57.1$ & $59.2$ & $41.0$ & $50.1$ & $52.8$ 
\\
& CPS~\cite{CPS} & $46.7$ & $58.7$ & $59.6$ & $41.4$ & $51.8$ & $53.9$
\\
& LaserMix~\cite{kong2023laserMix} & \cellcolor{yellow!12.5}$50.6$ & $60.0$ & $61.9$ & \cellcolor{yellow!12.5}$44.2$ & $53.7$ & $55.1$
\\
& LiM3D~\cite{li23lim3d} & - & \cellcolor{yellow!12.5}$61.6$ & \cellcolor{yellow!12.5}$62.6$  & - & \cellcolor{lightblue!9}$60.3$ & \cellcolor{lightblue!9}$60.5$
\\\cmidrule{2-8}
& \textbf{LaserMix++} & \cellcolor{lightblue!9}$\mathbf{56.2}$ & \cellcolor{lightblue!9}$\mathbf{62.3}$ & \cellcolor{lightblue!9}$\mathbf{62.9}$ & \cellcolor{lightblue!9}$\mathbf{47.3}$ & \cellcolor{yellow!12.5}$\mathbf{56.7}$ & \cellcolor{yellow!12.5}$\mathbf{57.6}$
\\\midrule\midrule

\multirow{7}{*}{\rotatebox{90}{Uniform}} &
\cellcolor{gray!10}\emph{Sup.-only} & \cellcolor{gray!10}$44.7$ & \cellcolor{gray!10}$56.2$ & \cellcolor{gray!10}$57.7$ & \cellcolor{gray!10}$39.6$ & \cellcolor{gray!10}$49.9$ & \cellcolor{gray!10}$52.4$
\\\cmidrule{2-8}
& MT~\cite{MeanTeacher} & $45.7$ & $58.3$ & $60.1$ & $40.0$ & $51.8$ & $54.3$ 
\\
& CPS~\cite{CPS} & $45.6$ & $58.5$ & $59.8$ & $41.7$ & $50.1$ & $53.1$
\\
& LaserMix~\cite{kong2023laserMix} & \cellcolor{yellow!12.5}$50.8$ & \cellcolor{yellow!12.5}$61.4$ & \cellcolor{yellow!12.5}$62.6$ & \cellcolor{yellow!12.5}$44.0$ & $54.0$ & $55.9$
\\
& LiM3D~\cite{li23lim3d} & - & $61.3$ & $62.4$ & - & \cellcolor{lightblue!9}$60.6$ & \cellcolor{lightblue!9}$60.3$
\\\cmidrule{2-8}
& \textbf{LaserMix++} & \cellcolor{lightblue!9}$\mathbf{56.0}$ & \cellcolor{lightblue!9}$\mathbf{62.6}$ & \cellcolor{lightblue!9}$\mathbf{63.0}$ & \cellcolor{lightblue!9}$\mathbf{47.3}$ & \cellcolor{yellow!12.5}$\mathbf{57.3}$ & \cellcolor{yellow!12.5}$\mathbf{58.0}$
\\\midrule\midrule

\multirow{7}{*}{\rotatebox{90}{Sequential}} &
\cellcolor{gray!10}\emph{Sup.-only} & \cellcolor{gray!10}$17.6$ & \cellcolor{gray!10}$41.8$ & \cellcolor{gray!10}$50.3$ & \cellcolor{gray!10}$16.4$ & \cellcolor{gray!10}$38.4$ & \cellcolor{gray!10}$47.9$
\\\cmidrule{2-8}
& MT~\cite{MeanTeacher} & $17.0$ & $42.4$ & $50.9$ & $15.9$ & $38.6$ & $46.7$ 
\\
& CPS~\cite{CPS} & $17.6$ & $43.3$ & $50.9$ & $16.5$ & $38.7$ & $48.1$
\\
& LaserMix~\cite{kong2023laserMix} & \cellcolor{yellow!12.5}$18.1$ & \cellcolor{yellow!12.5}$47.7$ & \cellcolor{yellow!12.5}$56.3$ & \cellcolor{yellow!12.5}$16.9$ & \cellcolor{yellow!12.5}$42.4$ & \cellcolor{yellow!12.5}$48.2$
\\
& LiM3D~\cite{li23lim3d} & - & - & - & - & - & -
\\\cmidrule{2-8}
& \textbf{LaserMix++} & \cellcolor{lightblue!9}$\mathbf{19.1}$ & \cellcolor{lightblue!9}$\mathbf{49.5}$ & \cellcolor{lightblue!9}$\mathbf{57.3}$ & \cellcolor{lightblue!9}$\mathbf{18.6}$ & \cellcolor{lightblue!9}$\mathbf{44.7}$ & \cellcolor{lightblue!9}$\mathbf{48.6}$
\\\midrule\midrule

\multirow{7}{*}{\rotatebox{90}{ST-RFD \cite{li23lim3d}}} &
\cellcolor{gray!10}\emph{Sup.-only} & \cellcolor{gray!10}$47.2 $& \cellcolor{gray!10}$56.9$ & \cellcolor{gray!10}$58.1$ & \cellcolor{gray!10}$40.1$ & \cellcolor{gray!10}$54.1$ & \cellcolor{gray!10}$55.5$
\\\cmidrule{2-8}
& MT~\cite{MeanTeacher} & $49.1$ & $59.8$ & $60.2$ & $42.3$ & $55.0$ & $56.8$ 
\\
& CPS~\cite{CPS} & $48.7$ & $60.5$ & $59.0$ & $42.4$ & $55.2$ & $56.6$
\\
& LaserMix~\cite{kong2023laserMix} & \cellcolor{yellow!12.5}$53.3$ & \cellcolor{yellow!12.5}$62.6$ & $62.9$ & $44.7$ & $54.1$ & $57.1$
\\
& LiM3D~\cite{li23lim3d} & - & $62.2$ & \cellcolor{yellow!12.5}$63.1$ & - & \cellcolor{lightblue!9}$61.0$ & \cellcolor{lightblue!9}$61.2$
\\\cmidrule{2-8}
& \textbf{LaserMix++} & \cellcolor{lightblue!9}$\mathbf{57.5}$ & \cellcolor{lightblue!9}$\mathbf{63.1}$ & \cellcolor{lightblue!9}$\mathbf{63.2}$ & \cellcolor{lightblue!9}$\mathbf{48.0}$ & \cellcolor{yellow!12.5}$\mathbf{57.6}$ & \cellcolor{yellow!12.5}$\mathbf{58.5}$
\\\bottomrule
\end{tabular}}
\label{tab:split}
\vspace{-0.4cm}
\end{wraptable}
\noindent\textbf{Component Analyses.} The ablation results in Table~\ref{tab:ablation} validate that each of the proposed components contributes significantly to the overall improvement in data-efficient 3D scene understanding. The mixing-based consistency regularization establishes a stable yet effective baseline across different datasets (Section~\ref{sec:lasermix}). Meanwhile, using the Teacher network instead of the Student network to generate pseudo-labels tends to yield better results, as the formal temporally ensembles and encourages spatial consistency (Section~\ref{sec:consistency_framework}). Moreover, integrating multi-modal interactions between LiDAR and cameras consistently improves performance, as encouraged by the camera-to-LiDAR feature distillation (Section~\ref{sec:c2l}) and the language-driven knowledge guidance (Section~\ref{sec:lkg}). It is worth noting that all model configurations have achieved superior results than the baseline MeanTeacher~\cite{MeanTeacher}, which further emphasizes the effectiveness of our framework in tackling LiDAR segmentation under data-efficient learning setups.

\begin{figure}[t]
    \centering
    \begin{subfigure}[b]{0.518\textwidth}
        \centering
        \includegraphics[width=\textwidth]{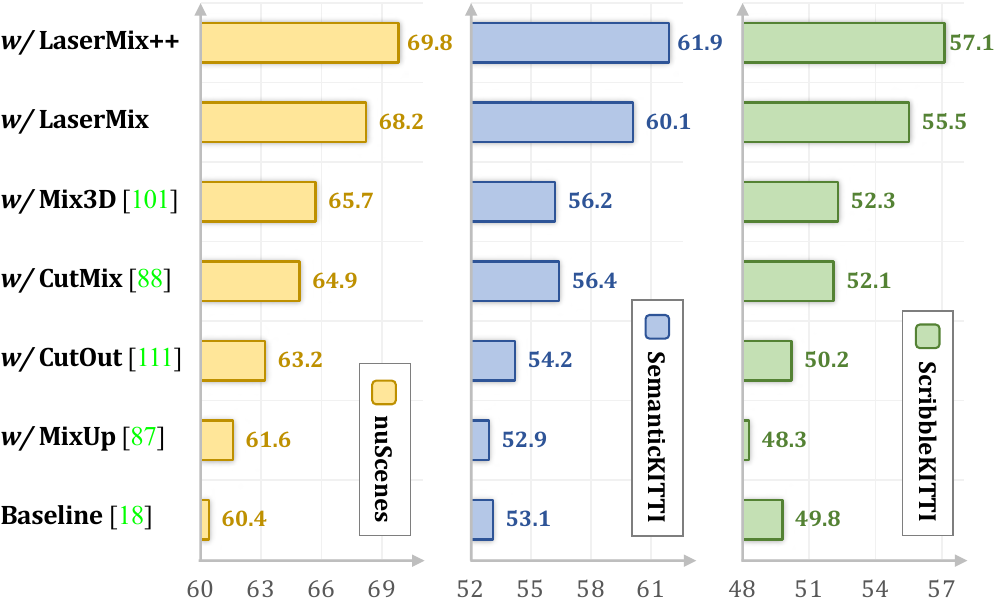}
        \caption{Mixing Strategy}
        \label{fig:ablation-mix}
    \end{subfigure}~~
    \begin{subfigure}[b]{0.223\textwidth}
        \centering
        \includegraphics[width=\textwidth]{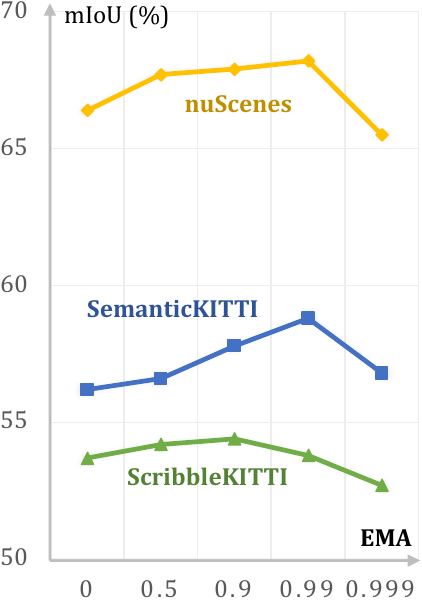}
        \caption{EMA}
        \label{fig:ablation-ema}
    \end{subfigure}~~~~~
    \begin{subfigure}[b]{0.223\textwidth}
        \centering
        \includegraphics[width=\textwidth]{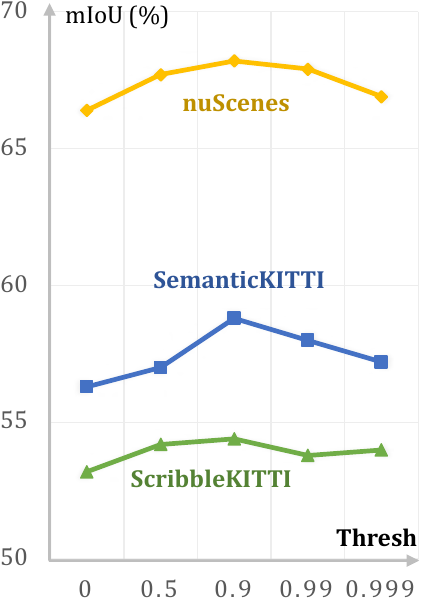}
        \caption{Threshold}
        \label{fig:ablation-threshold}
    \end{subfigure}
    \vspace{-0.5cm}
    \caption{\textbf{Ablation study} on \textbf{(a)} Different mixing-based techniques used in point partition \& mixing. \textbf{(b)} Different EMA decay rates between the Teacher and Student networks. \textbf{(c)} Different confidence thresholds $T$ that are used in the pseudo-label generation process.}
    \label{fig:ablation-stats}
\end{figure}

\noindent\textbf{Mixing Strategies.}
Another ablation experiment, depicted in Figure~\ref{fig:ablation-mix}, compares the performance of LaserMix and LaserMix++ against traditional mixing techniques, \emph{i.e.}, MixUp \cite{MixUp}, CutOut \cite{CutOut}, CutMix \cite{CutMix}, and Mix3D \cite{Mix3D}. While MixUp and CutMix manipulate random points and areas, respectively, they do not inherently leverage the structured nature of LiDAR data, where spatial relations significantly influence segmentation accuracy. CutMix shows improvement by utilizing the structural priors in scene segmentation; however, LaserMix and LaserMix++, which consider both area structure and precise spatial positioning, provide a more substantial boost in performance, outperforming CutMix by up to 3.3\% mIoU. CutOut can be considered as setting $\Xcomp$ to a dummy filling instead of sampling from datasets, and it leads to a considerable performance drop from CutMix.

\noindent\textbf{Alternative Heuristic Data Mixing.} 
Exploring further, we evaluate different heuristic approaches to LiDAR scan partitioning, including azimuth and radius-based splits. As shown in Table~\ref{tab:granularity}, azimuth splits do not correlate well with semantic distributions, offering no performance gains. In contrast, finer granularity in mixing generally improves results until it reaches a threshold beyond which it disrupts semantic coherence. This observation supports our use of inclination-based partitioning in LaserMix, which maintains semantic integrity better than purely radial or azimuthal approaches.

\noindent\textbf{Mix Unlabeled Data Only.}
To underscore that LaserMix extends beyond simple data augmentation, we experiment with mixing solely between unlabeled scans. This modification results in a slight performance drop (from 68.2\% to 66.9\% mIoU), yet still significantly outperforms traditional methods, highlighting the strong consistency regularization imparted by LaserMix even without direct labeled data interaction.

\noindent\textbf{Data Splitting Strategies.}
For a semi-supervised learning problem, the way of partitioning labeled and unlabeled data plays a crucial role. We compare four different splitting strategies (the first three are shown in Figure~\ref{fig:split}) that meet the requirements of driving data collection and perception, \emph{i.e.}, random, uniform, sequential sampling strategies, and the recent visual-encouraged ST-RFD \cite{li23lim3d} strategy. As shown in Table~\ref{tab:split}, methods under the random or uniform sampling achieved higher results than sequential sampling. This is because the former two introduce a more diverse sample distribution than sampling sequentially. ST-RFD, which picks scans using additional image cues, provides an even richer labeled data set. Under all four strategies, LaserMix++ consistently achieved better performance than the baseline LaserMix \cite{kong2023laserMix} and MeanTeacher \cite{MeanTeacher} frameworks and is competitive with Lim3D \cite{li23lim3d} on ScribbleKITTI \cite{ScribbleKITTI}.

\noindent\textbf{Dual-Branch Consistency.}
As shown in Figure~\ref{fig:ablation-ema}, our dual-branch consistency setup, leveraging different exponential moving average (EMA) decay rates, shows that a balance between 0.9 and 0.99 optimizes performance, while higher rates disrupt network consistency. This configuration corroborates the synergistic potential of the teacher-student architecture within our approaches, facilitating the integration of modern semi-supervised learning techniques.

\noindent\textbf{Confidence Threshold.}
The role of pseudo-labels is critical in our framework. As shown in Figure~\ref{fig:ablation-threshold}, adjusting the confidence threshold for pseudo-label generation shows that overly low thresholds degrade performance by enforcing consistency on unreliable labels. Conversely, high thresholds may diminish the mixing benefits. Optimal threshold tuning, around 0.9, is crucial and varies across datasets, underpinning the adaptability of our approaches.

\begin{wraptable}{r}{0.5\textwidth}
\centering
\vspace{-0.4cm}
\caption{
    \textbf{Fully-supervised LiDAR semantic segmentation results} on the \emph{val} sets of SemanticKITTI \cite{SemanticKITTI}, nuScenes \cite{Panoptic-nuScenes}, and ScribbleKITTI \cite{ScribbleKITTI}. All IoU scores are given in percentage ($\%$). Symbol \textcolor{lightblue}{\cmark} denotes that the model has been trained with LaserMix as the data augmentation during the training.}
    \vspace{-0.2cm}
    \resizebox{\linewidth}{!}{
    \begin{tabular}{r|c|p{20.0pt}<{\centering}p{20.0pt}<{\centering}|p{20.0pt}<{\centering}p{20.0pt}<{\centering}|p{20.0pt}<{\centering}p{20.0pt}<{\centering}}
    \toprule
    \multirow{2}{*}{\textbf{Method}} & \multirow{2}{*}{\textbf{Mix}} & \multicolumn{2}{c|}{\textbf{KITTI}} & \multicolumn{2}{c|}{\textbf{nuScenes}} & \multicolumn{2}{c}{\textbf{Scribble}}
    \\
    & & \textbf{mIoU} & \textbf{mAcc} & \textbf{mIoU} & \textbf{mAcc} & \textbf{mIoU} & \textbf{mAcc}
    \\\midrule\midrule
    \multirow{2}{*}{MinkUNet} \cite{choy2019minkowski} & \textcolor{gray}{\xmark} & $66.9$ & $92.4$ & $76.4$ & $94.1$ & $61.2$ & $88.5$
    \\
    & \textcolor{lightblue}{\cmark} & \cellcolor{lightblue!9}$\mathbf{68.1}$ & \cellcolor{lightblue!9}$\mathbf{92.6}$ & \cellcolor{lightblue!9}$\mathbf{76.9}$ & \cellcolor{lightblue!9}$\mathbf{94.3}$ & \cellcolor{lightblue!9}$\mathbf{62.0}$ & \cellcolor{lightblue!9}$\mathbf{89.0}$
    \\\midrule
    \multirow{2}{*}{Cylinder3D} \cite{Cylinder3D} & \textcolor{gray}{\xmark} & $63.7$ & $91.0$ & $75.8$ & $93.7$ & $58.8$ & $87.4$
    \\
    & \textcolor{lightblue}{\cmark} & \cellcolor{lightblue!9}$\mathbf{64.7}$ & \cellcolor{lightblue!9}$\mathbf{91.3}$ & \cellcolor{lightblue!9}$\mathbf{78.1}$ & \cellcolor{lightblue!9}$\mathbf{94.0}$ & \cellcolor{lightblue!9}$\mathbf{59.5}$ & \cellcolor{lightblue!9}$\mathbf{88.6}$
    \\\midrule
    \multirow{2}{*}{PolarNet} \cite{PolarNet} & \textcolor{gray}{\xmark} & $57.2$ & $91.0$ & $71.7$ & $93.1$ & $55.7$ & $87.6$
    \\
    & \textcolor{lightblue}{\cmark} & \cellcolor{lightblue!9}$\mathbf{59.6}$ & \cellcolor{lightblue!9}$\mathbf{91.2}$ & \cellcolor{lightblue!9}$\mathbf{72.1}$ & \cellcolor{lightblue!9}$\mathbf{93.2}$ & \cellcolor{lightblue!9}$\mathbf{56.5}$ & \cellcolor{lightblue!9}$\mathbf{88.1}$
    \\\midrule
    \multirow{2}{*}{FRNet} \cite{xu2023frnet} & \textcolor{gray}{\xmark} & $64.1$ & $92.2$ & $76.8$ & $93.4$ & $57.6$ & $88.3$
    \\
    & \textcolor{lightblue}{\cmark} & \cellcolor{lightblue!9}$\mathbf{66.4}$ & \cellcolor{lightblue!9}$\mathbf{92.4}$ & \cellcolor{lightblue!9}$\mathbf{77.6}$ & \cellcolor{lightblue!9}$\mathbf{93.9}$ & \cellcolor{lightblue!9}$\mathbf{59.6}$ & \cellcolor{lightblue!9}$\mathbf{88.8}$
\\\bottomrule
\end{tabular}}
\label{tab:fully}
\vspace{-0.4cm}
\end{wraptable}
\noindent\textbf{Extension to Fully-Supervised Learning.} 
The core principles of LaserMix, which involve partitioning and mixing LiDAR data to create diverse and challenging training samples, are not inherently limited to semi-supervised learning. These principles can be directly applied to fully-supervised settings, where the availability of labeled data allows for a more comprehensive exploration of the method’s potential. As shown in Table~\ref{tab:fully}, our approach consistently improves by approximately 0.4\% to 2.4\% mIoU scores across different backbones. These results confirm that the core principles of our mixing-based methods, originally designed for semi-supervised tasks, can also significantly enhance performance in fully-supervised learning scenarios.
\section{Conclusion}
\label{sec:conclusion}
In this study, we have extended semi-supervised LiDAR semantic segmentation by introducing LaserMix++, a framework that integrates LiDAR and camera data to exploit spatial and textural synergies. Building on our initial LaserMix technique, which intertwines laser beams from different scans, LaserMix++ enhances feature richness and model robustness, particularly under varied environmental conditions. Our empirical evaluations demonstrate that LaserMix++ significantly outperforms existing methods, achieving high accuracy with far fewer labels. This efficiency underscores its potential for reducing the dependency on extensive annotated data. Looking forward, we aim to refine spatial partitioning and incorporate advanced semi-supervised techniques to expand our framework's applications to other critical tasks such as 3D object detection and tracking.

\bibliographystyle{plainnat}
\bibliography{main}

\end{document}